\documentclass{article} 
\usepackage{times}
\usepackage{main}

\usepackage{amsmath,amsfonts,bm}

\def\ourname{\textsc{PairFlow}}
\def\ie{\emph{i.e.,}}

\def\Figref#1{Fig.~\ref{#1}}

\def\Secref#1{Sec.~\ref{#1}}

\def\Tabref#1{Tab.~\ref{#1}}
\def\eqref#1{eqn.~\ref{#1}}
\def\Eqref#1{Eqn.~\ref{#1}}

\def\Algref#1{Alg.~\ref{#1}}

\def\1{\bm{1}}

\DeclareMathAlphabet{\mathsfit}{\encodingdefault}{\sfdefault}{m}{sl}
\SetMathAlphabet{\mathsfit}{bold}{\encodingdefault}{\sfdefault}{bx}{n}

\usepackage[
linesnumbered,ruled,vlined]{algorithm2e}
\usepackage {algpseudocode}
\usepackage{algorithmicx}
\usepackage{algcompatible}
\usepackage[utf8]{inputenc} 
\usepackage[T1]{fontenc}    
\usepackage{hyperref}       
\usepackage{url}            
\usepackage{booktabs}       
\usepackage{amsfonts}       
\usepackage{nicefrac}       
\usepackage{microtype}      
\usepackage{xcolor}         
\usepackage[table]{xcolor}         
\usepackage{graphicx}
\usepackage{bbm}
\usepackage{tcolorbox}
\usepackage{amsthm}
\usepackage{float}
\usepackage{kotex}
\usepackage{subcaption}
\usepackage{caption}

\usepackage{tabularx}
\usepackage{multicol, multirow}
\usepackage{makecell}
\newcolumntype{Y}{>{\centering\arraybackslash}X}

\newcommand*{\methodname}[0]{\textsc{PairFlow}}

\title{PairFlow: Closed-Form Source-Target Coupling for Few-Step Generation in Discrete Flow Models}

\author{Mingue Park\textsuperscript{$\ast$} \hspace{0.5em} Jisung Hwang\textsuperscript{$\ast$} \hspace{0.5em} Seungwoo Yoo\textsuperscript{$\ast$} \hspace{0.5em} Kyeongmin Yeo \hspace{0.5em} Minhyuk Sung \\
KAIST\\
\texttt{\{kicikicik,4011hjs,dreamy1534,aaaaa,mhsung\}@kaist.ac.kr} \\
}

\iclrfinalcopy 
\begin{document}

\vspace*{-0.3cm}
\maketitle
\begingroup
\renewcommand\thefootnote{}\footnotetext{\textsuperscript{$\ast$}Equal contribution.}
\endgroup

\pagestyle{fancy}
\fancyhead{}            
\fancyhead[L]{Preprint}


\begin{abstract}
\vspace{-0.5\baselineskip}
We introduce~\ourname{}, a lightweight preprocessing step for training Discrete Flow Models (DFMs) to achieve few-step sampling without requiring a pretrained teacher. DFMs have recently emerged as a new class of generative models for discrete data, offering strong performance. However, they suffer from slow sampling due to their iterative nature. Existing acceleration methods largely depend on finetuning, which introduces substantial additional training overhead.~\ourname{} addresses this issue with a lightweight preprocessing step. Inspired by ReFlow and its extension to DFMs, we train DFMs from coupled samples of source and target distributions, without requiring any pretrained teacher.
At the core of our approach is a closed-form inversion for DFMs, which allows efficient construction of paired source–target samples. Despite its extremely low cost, taking only up to 1.7\% of the compute needed for full model training, \ourname{} matches or even surpasses the performance of two-stage training involving finetuning. Furthermore, models trained with our framework provide stronger base models for subsequent distillation, yielding further acceleration after finetuning. Experiments on molecular data as well as binary and RGB images demonstrate the broad applicability and effectiveness of our approach.
\end{abstract}

\section{Introduction}
\label{sec:intro}
\vspace{-0.75\baselineskip}

Discrete Flow Models (DFMs)~\citep{Campbell:2024MultiFlow,Gat:2024DFM} have recently emerged as a promising class of generative models, extending the idea of Flow Models (FMs)~\cite{} for continuous data to the discrete domain. By adapting flow-based principles to categorical structures, DFMs provide a principled and efficient way to capture complex discrete distributions through iterative sampling. They have shown success across a variety of applications, particularly in scientific domains such as molecule generation~\citep{ramakrishnan2014quantum,irwin2012zinc}, where DFMs offer a natural framework for modeling chemical structures and generating novel candidates.

Analogous to FMs in the continuous domain, a key challenge of DFMs is the long computation time for generation due to their iterative sampling nature. Recent work~\citep{Deschenaux:2025sdtt,Hayakawa:2025di4c,Sahoo:2025Duo,Yoo:2025ReDi} have sought to accelerate the generative process through distillation-based finetuning, which builds on ideas originally developed for continuous flow matching. Notably, ReFlow~\citep{Liu:2023RF} is a well-known technique for FMs that pairs samples from the source (prior) distribution and the target distribution by running the generative process of a pretrained model and using the resulting pairs for finetuning. Recently, this idea has also been extended to DFMs~\citep{Yoo:2025ReDi} to reduce conditional total correlation through finetuning, thereby enabling few-step generation.

Despite these promising results in acceleration, distillation-based methods incur substantial finetuning overhead, amounting to about 10–20\% of the time required to train the base model from scratch. In other words, the gain in generation speed comes at the expense of considerable additional training cost. To our knowledge, no prior work has addressed this training-time cost when pursuing inference-time acceleration. This raises a natural question: can we achieve speedups comparable to distillation-based approaches while requiring only a lightweight preprocessing phase that requires orders of magnitude less compute, on the order of tens of GPU minutes?

We propose~\ourname{}, a training framework for DFMs that enables few-step sampling by constructing paired source–target samples using closed-form velocities.~\textbf{While inspired by ReDi-style coupling-driven training, our approach eliminates the need for a pretrained teacher by using closed-form formulations and achieves acceleration without finetuning.} The algorithm for computing source–target pairs is fully parallelizable and requires at most 1.7\% of compute needed for full model training. Despite relying only on a lightweight preprocessing step,~\ourname{} attains performance comparable to, or even superior to state-of-the-art distillation-based techniques, which can require up to 143 times more computation.
Furthermore, models trained with our technique provide stronger bases for subsequent distillation, delivering additional performance gains while incurring only minimal preprocessing cost.

At the core of our framework is the simulation of probability paths connecting source (prior) and target (data) distributions in discrete spaces, made possible by closed-form expressions of velocities. While closed-form forward velocities have been studied for flow models in continuous domains~\citep{Karras:2022EDM,Bertrand:2025Closed}, they have, to the best of our knowledge, neither been explored for DFMs nor applied to identifying suitable source–target pairs in the context of distillation-based acceleration, as in ReDi~\citep{Yoo:2025ReDi}. In this work, we investigate this idea for the first time.
For DFMs, with a particular focus on uniform-state models~\citep{Sahoo:2025Duo,Schiff:2025UDLM} equipped with a self-correcting mechanism, we show that the closed-form forward velocity is determined by the Hamming distance, which measures the number of differing tokens between two sequences. Using this velocity, samples from the source (latent) distribution can be mapped to given target samples. However, because multiple source samples may map to the same target, covering all targets through coupling would require an impractically large number of source samples. To overcome this, we derive the corresponding backward velocity in closed form and leverage it to simulate backward probability paths that efficiently map data points to source points, making pair discovery computationally efficient.

In our experiments, we show that the proposed framework enables few-step sampling across diverse discrete domains, including molecular data~\citep{ramakrishnan2014quantum,irwin2012zinc} and 2D images, exemplified by MNIST-Binary~\citep{lecun2002gradient} and CIFAR-10~\citep{krizhevsky2009learning}. On the QM9~\citep{ramakrishnan2014quantum} and ZINC-250k~\citep{irwin2012zinc} datasets,~\textbf{~\ourname{} not only improves the base model but also performs comparably to, or even better than, distilled models that require up to $143\times$ more compute during finetuning, compared to our lightweight preprocessing algorithm.} Similar improvements are observed on MNIST-Binary, where models paired with~\ourname{} achieve performance comparable to those using DCD~\citep{Sahoo:2025Duo} and ReDi~\citep{Yoo:2025ReDi}, while being up to $35\times$ faster. Furthermore, after subsequent distillation, base models trained with pairs generated by our method consistently achieve higher performance, underscoring the importance of well-constructed source–target pairings.

\section{Related Work}
\label{sec:related}
\vspace{-0.5\baselineskip}

\subsection{Discrete Flow models}
\label{subsec:related_discrete_flow_models}
\vspace{-0.5\baselineskip}

The concept of Flow Matching~\citep{Lipman:2023FlowMatching} has recently been extended to the discrete flow-based models~\citep{Gat:2024DFM,Campbell:2024MultiFlow,Sahoo:2024MDLM,shi2024:simplemdm,Schiff:2025UDLM}, demonstrating its flexibility across high-dimensional and structured data~\citep{Bai:2025meissonic,Chang:2022maskgit,Weber:2024maskbit,Arriola:2025blockdiffusion,Nie:2025largeldm,Yu:2023magvit,Lee:2025genmol,Campbell:2024MultiFlow,Wang:2025drakes}.
Among these, uniform-state models~\citep{Sahoo:2025Duo,Schiff:2025UDLM} have been studied for their self-correcting properties, which enable recovery from errors introduced during parallel decoding. However, their performance degrades markedly in few-step settings, posing a key limitation for efficient generation under tight compute budgets.

\subsection{Accelerating Discrete Flow models}
\label{subsec:related_discrete_rectified_flow}
\vspace{-0.5\baselineskip}

Several approaches have been proposed to accelerate sampling with DFMs.~\citet{Park:2025jumpyoursteps} directly optimize sampling timesteps for improved parallelism while mitigating decoding errors.~\citet{Hayakawa:2025di4c} highlight the importance of capturing dimensional correlations for faster sampling and introduce mixture models to this end, at the cost of additional loss terms that complicate optimization. Another line of work adapts techniques from continuous domains, as in~\citet{Sahoo:2025Duo}, that propose a discrete analogue of Consistency Distillation (CD)~\citep{Song:2023consistency} by leveraging the duality between uniform-state and continuous Gaussian models. Most relevant to our approach is ReDi~\citep{Yoo:2025ReDi}, which draws inspiration from the concept of straight flows in ReFlow~\citep{Liu:2023RF} and iteratively optimizes pairs of data and noise samples.


\section{Preliminaries}
\label{sec:preliminary}
\vspace{-0.75\baselineskip}

In this section, we provide a brief overview of flow matching for generative modeling of discrete data (\Secref{subsec:flow_matching}), followed by a rectification technique~\citep{Yoo:2025ReDi} that enables faster generation (\Secref{subsec:rectified_flow}) by reducing total correlation errors.

\subsection{Discrete Flow matching}
\label{subsec:flow_matching}
\vspace{-0.5\baselineskip}

The goal of Discrete Flow Matching (DFM)~\citep{Campbell:2024MultiFlow,Gat:2024DFM} is to learn a probability path $p_t(\cdot)$ that connects a known, easy-to-sample source distribution $p(x)$ to an unknown target distribution $q(x)$, both defined over a discrete state space. Once $p_t(\cdot)$ is known, samples from $q$ can be generated by drawing $x_0 \sim p$ and transporting it along the path. 

Specifically, consider a sequence $x = (x^1, x^2, \ldots, x^N)$ of $N$ tokens, where each token takes values in a vocabulary $\mathcal{V} = \{1, 2, \dots, K\}$ of size $K$. A sequence $x$ then resides in the product space $\mathcal{V}^N$. We denote by $\Delta^K = \{ p \in \mathbb{R}^K \vert \sum_i p_i=1, p_i \geq 0\}$ the probability simplex of dimension $K-1$, on which distributions over $\mathcal{V}$ are defined.

Given a target probability path $p_t(x): \mathcal{V}^N \times [0,1] \to [0,1] $ with an associated velocity field $v_t(x): \mathcal{V}^N \times [0,1] \to \mathbb{R}^{N\times K}$, we introduce a network $p_{1 \vert t}^{\theta}(x): \mathcal{V}^N \times [0,1] \to (\Delta^K)^N$ to approximate $v_t(x)$.
Its parameters $\theta$ are optimized via the DFM objective~\citep{Gat:2024DFM}:
\begin{align}
    \mathcal{L}_{\text{DFM}}(\theta) = - \sum_{i \in \{1,\dots,N\}} \mathbb{E}_{t \sim \mathcal{U}[0,1], x_0 \sim p, x_1 \sim q, z \sim p_t(x|x_0, x_1)} \log p_{1 \vert t}^\theta (x_1^i \vert z),
    \label{eqn:dfm_loss}
\end{align}
where $p_{1 \vert t}^\theta(x_1^i \vert z)$ denotes the learned probability denoiser, which predicts the categorical distribution of the clean token $x_1^i$ given an intermediate sequence $z$.
Here, the conditional probability path $p_t(z \vert x_0, x_1)$ generates samples $z$ by interpolating between a data point $x_1~\sim q$ and a source sample $x_0~\sim p$. Assuming independence across tokens in sequence $x$, the conditional density factorizes as
\begin{align}
    p_t(z \vert x_0, x_1) = \prod_{i=1}^N p_t(z^i \vert x_0, x_1).
    \label{eqn:p_t_factorize}
\end{align}
As token-wise conditional paths $p_t(z^i \vert x_0, x_1)$,~\citet{Gat:2024DFM} employ the mixture path of form:
\begin{align}
    p_t(z^i|x_0, x_1) = (1-\kappa_t)\delta_{x_0}(z^i) + \kappa_t\delta_{x_1}(z^i),
\end{align}
where the~\emph{scheduler} $\kappa_t = \kappa(t)$ is a monotonically increasing function over $t \in [0,1]$ satisfying $\kappa_0 = 0$ and $\kappa_1 = 1$. For notational convenience, given $x, y \in \mathcal{V}^N$, we define the Dirac delta $\delta_y(x)$ as
\begin{align}
    \delta_y (x) = \prod_{i=1}^N \delta_{y^i} (x^i), \,\, \text{where } \delta_{y^i} (x^i) = \begin{cases} 1 & x^i=y^i \\ 0 & x^i \neq y^i \end{cases}.
\end{align}
We also use the shorthand $\delta_y(x^i) = \delta_{y^i}(x^i)$ .
After optimizing $\theta$, the learned model parameterizes an approximation of the marginal velocity field:
\begin{align}
    v_t^{\theta}(x^i, z) = \frac{\dot\kappa_t}{1 - \kappa_t} \left[p^{\theta}_{1|t}(x^i|z) - \delta_{z}(x^i) \right],
\end{align}
where $\dot{\kappa}_t=\frac{\partial \kappa_t}{\partial t}$.
This learned velocity field $v_t^{\theta}(x^i, z)$ then transports samples over the interval $[0, 1]$ to simulate trajectories along $p_t(\cdot)$ and thereby generate samples. Each update step is defined as:
\begin{align}
    x_{t+h}^i \sim \text{Cat} \left(x_{t+h}^i; \delta_{x_t^i}(\cdot) + h \cdot v^\theta_t(x_{t+h}^i, x_t) \right),
    \label{eqn:sampling_step}
\end{align}
where $h > 0$ is the step size.

\subsection{Straightening Probability Paths for Accelerated Sampling}
\label{subsec:rectified_flow}
\vspace{-0.5\baselineskip}

The concept of straight probability paths was originally introduced in the continuous domain to enable accelerated sampling. Prior work~\citep{Liu:2023RF} identified curved probability paths as a key challenge in few-step sampling: when velocity fields are evaluated only at coarse time steps, numerical integration deviates from the true trajectories. \citet{Liu:2023RF} addressed this issue through ``rectification,'' in which a student flow model is trained on source--target pairs generated by a teacher model, effectively yielding significantly straighter probability paths.

In the discrete setting, this challenge of \textit{path curvature} translates to capturing the \textit{statistical correlations} between tokens. Since DFMs approximate exponentially large joint transitions through fully factorized per-token updates, a mismatch inevitably arises between the true joint transition and its product-form approximation. This discrepancy becomes especially detrimental during few-step generation, where highly correlated tokens must be updated simultaneously. To address this, prior works have primarily relied on distillation-based approaches~\citep{Hayakawa:2025di4c,Sahoo:2025Duo,Deschenaux:2025sdtt}, aiming to better capture these correlations by explicitly transferring multi-step dependencies from a stronger teacher model.

\citet{Yoo:2025ReDi} formalized this factorization mismatch via conditional Total Correlation (TC), defined as:
\begin{align}
    \text{TC}_{\pi}(x_s|x_t) = \mathbb{E}_{x_t} \left[ D_{\mathrm{KL}} \left( p_{s|t}(x_s|x_t) \Vert \prod_{i=1}^{N} p_{s|t}(x_s^i|x_t) \right) \right],
    \label{eqn:redi_tc}
\end{align}
which serves as a metric for the factorization error. Crucially, \citet{Yoo:2025ReDi} interpret this factorization error as the discrete analog of path curvature: minimizing TC is equivalent to ``straightening'' the trajectory by decoupling token transitions. Analogous to ReFlow~\citep{Liu:2023RF}, which rectifies continuous paths, they demonstrate that reducing~\Eqref{eqn:redi_tc} requires iteratively refining the source--target coupling $\pi(x_0, x_1)$. To achieve this, they employ an iterative distillation process, alternating between generating improved pairs using the current model and optimizing $\mathcal{L}_{\text{DFM}}$. This procedure effectively finds a ``statistically straight'' coupling that enables efficient few-step generation.

\section{PairFlow}
\label{sec:background}
\vspace{-0.5\baselineskip}

For DFMs, ReDi~\citep{Yoo:2025ReDi} improves sample quality in few-step generation by rectifying source-target pairs. However, it relies on samples from a pretrained model followed by costly retraining or finetuning. We take this one step further and pose the following question: What if these pairs could be generated directly from the data, without relying on a pretrained model or sampling from it?

To address this question, we propose a principled approach for discovering well-aligned source–target pairs without relying on pretrained models, enabling models trained on such pairs to achieve strong performance with few-step sampling. Our method, termed~\ourname{}, leverages closed-form velocity fields that can be computed directly from the data samples, requiring only prior knowledge of the source distribution. We assume this distribution to be uniform, a choice extensively studied in recent work~\citep{Sahoo:2025Duo,Schiff:2025UDLM}, as models trained under this prior naturally acquire self-correcting properties.

In~\Secref{subsec:closed_form_forward_velocity}, we introduce the closed-form forward velocity for discrete flow matching~\citep{Gat:2024DFM}. In~\Secref{subsec:closed_form_backward_velocity}, we extend this to the closed-form backward velocity and propose an algorithm for discovering well-aligned source–target pairs during the preprocessing phase.


\subsection{Finding Pairs via Closed-Form Forward Velocity Fields}
\label{subsec:closed_form_forward_velocity}
\vspace{-0.5\baselineskip}

As discussed in~\Secref{subsec:flow_matching}, DFMs~\citep{Campbell:2024MultiFlow,Gat:2024DFM} aim to learn a marginal velocity field $v_t(\cdot)$ that induces a probability path $p_t(\cdot)$, transporting the source distribution $p_0(\cdot)$ to the target distribution $q(\cdot) = p_1(\cdot)$, which is unknown in practice. Instead, we only have access to a finite dataset of $M$ samples $\{d_m\}_{m=1}^M$. This empirical distribution $\tilde{q}(x)$ can be represented as a mixture of Dirac deltas centered at the observed samples:
\begin{equation}
    q(x) \approx \tilde{q}(x) = \frac{1}{M} \sum_{m=1}^M \delta_{d_m}(x).
    \label{eqn:empirical_target}
\end{equation}
For continuous domains,~\citet{Karras:2022EDM,Bertrand:2025Closed} have shown that the velocity field transporting $p_0$ to $q$ can be derived in closed form when both distributions admit tractable density expressions. To the best of our knowledge, this idea has not been explored in discrete domains; in the following, we derive the closed-form velocity field for discrete domains for the first time.

We base our framework on the assumption of a uniform prior distribution over the discrete state space $\mathcal{V}^N$, defined as $p_0(x) = \mathcal{U}^N$, where $\mathcal{U} = \text{Cat}\left(\cdot;\frac{\mathbf{1}}{K}\right)$ denotes the uniform distribution over the dictionary $\mathcal{V}$.
For the empirical target distribution $\tilde{q}(x)$ introduced in~\Eqref{eqn:empirical_target}, we show in App.~\ref{subsec:appx_proof_forward_velocity_discrete} that the closed-form denoiser $p_{1 \vert t}(x^i \vert z)$ and its associated velocity field $\hat{v}_t(x^i, z)$ are given by:

\begin{align}
    p_{1|t}(x^i|z) = \cfrac{\sum_{m=1}^M \delta_{d_{m}^i}(x^i) \gamma^{- h(d_{m}, z)}}{ \sum_{m=1}^M \gamma^{- h(d_{m}, z)}} \quad \Longrightarrow \quad 
    \hat{v}_t(x^i, z) = \frac{\dot\kappa_t}{1 - \kappa_t} \left[p_{1|t}(x^i|z) - \delta_{z}(x^i) \right]
    \label{eqn:closed_forward_velocity}
\end{align}

where $\gamma = (1 + (K-1)\kappa_t)/(1 - \kappa_t)$, $K$ denotes the vocabulary size, and $h(s, z) = N - \sum_{i=1}^N \delta_{s^i}(z^i)$ is the Hamming distance between sequences $s$ and $z$,~\ie{} the number of positions at which they differ. 
The token-wise denoiser $p_{1 \vert t}(x^i \vert z)$ above is a weighted mixture of Dirac deltas, where sequences closer to $z$ under the Hamming distance contribute more.
Intuitively, the forward velocity field $\hat{v}_t(x^i, z)$ pulls each token toward those from dataset sequences most similar to $z$.
The most direct way to construct source-target pairs using $\hat{v}_t(x^i, z)$ is to sample $x_0 \sim p_0(x)$ and evolve it along the velocity field until it reaches a dataset point $x_1$. In practice, however, the generated data points fail to fully cover $\tilde{q}(x)$, requiring an impractically large number of source samples to achieve sufficient coverage. Our empirical results, reported in App.~\ref{subsec:coverage_forward_velocity}, support this claim and motivate the exploration of a more efficient alternative, which we present in the following section.
{\scriptsize
\begin{figure}[t]
    \centering
    \begin{minipage}[t]{0.46\textwidth}
        \vspace{0pt}%
        \includegraphics[width=\linewidth]{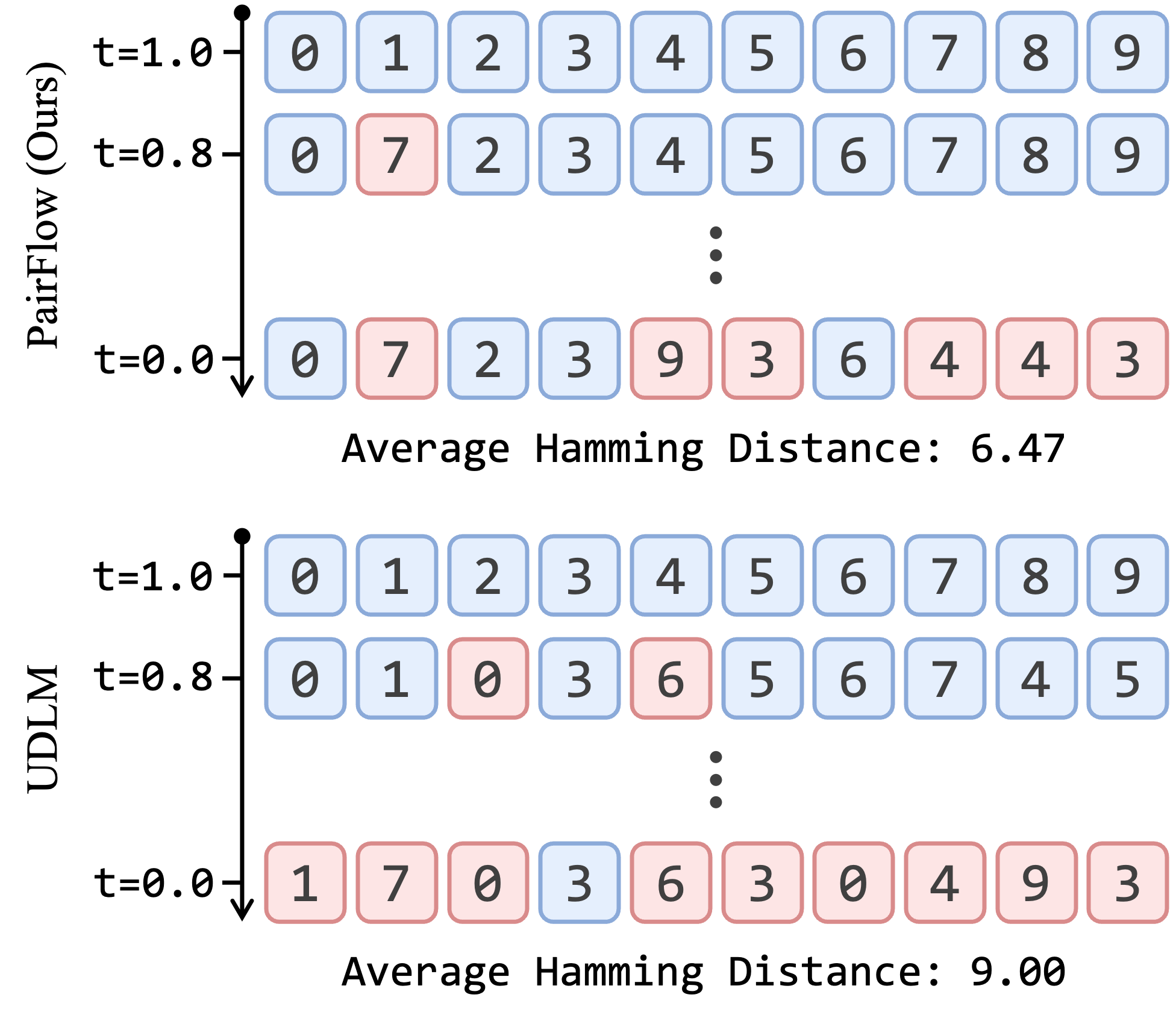}
        \vspace{-20pt}
        \caption{Illustrations of data inversion in~\ourname{} and the standard corruption process in UDLM.~\ourname{} achieves a lower average Hamming distance (6.47 vs.\ 9.0), promoting straighter paths during training.}
        \label{fig:pairflow_method}
    \end{minipage}%
    \hfill
    \begin{minipage}[t]{0.46\textwidth}
        \vspace{0pt}%
        \begin{algorithm}[H]
        \caption{\ourname{}}
        \label{alg:coupling_backward}
        \textbf{Input:} Dataset $\{d_m\}_{m=1}^M$, Steps $T$ \\
        \textbf{Output:} Pairs $\pi=\{(x_{0,m}, x_{1,m})\}_{m=1}^M$ 
        
        \medskip
        
        Initialize $\pi \gets \emptyset$
        
        \For{$m \gets 1$ \textbf{to} $M$}{
            $x_{1,m} \gets d_m$ \\
            $x \gets x_{1,m}$
        
            \For{$t \gets 1$ \textbf{to} $T$}{
                Compute $p_{0 \vert t}(\cdot \vert x)$ and $\check{v}_t(\cdot, x)$ from~\Eqref{eqn:closed_backward_velocity} \\
                Sample $z \sim \text{Cat}\!\left(z;\, \delta_{x}(\cdot) - h \cdot \check{v}_t(\cdot, x)\right)$ \\
                $x \gets z$
            }
        
            $x_{0,m} \gets x$ \\
            $\pi \gets \pi \cup \{(x_{0,m}, x_{1,m})\}$
        }
        \Return $\pi$
        \end{algorithm}
    \end{minipage}
\end{figure}
}


\subsection{Finding Pairs via Closed-Form Backward Velocity Fields}
\label{subsec:closed_form_backward_velocity}
\vspace{-0.5\baselineskip}

We address this issue by \emph{backtracing} trajectories along $p_t(\cdot)$, starting from $\tilde{q}(x)$ and progressing toward the source distribution $p_0(\cdot)$. Unlike the forward construction in~\Secref{subsec:closed_form_forward_velocity}, this guarantees that all data points in $\tilde{q}(x)$ are included in the resulting pairs by design.
As illustrated at the top of~\Figref{fig:pairflow_method},~\ourname{} inverts data samples toward the source distribution, assumed to be uniform. 
Unlike the standard corruption process used by UDLM~\citep{Schiff:2025UDLM} shown at the bottom of~\Figref{fig:pairflow_method}, the source samples obtained by~\ourname{} remain closer to the original data in terms of Hamming distance. This helps the model learn to recover data with fewer token transitions during training, effectively approximating the straight probability paths explored in ReFlow~\citep{Liu:2023RF} and ReDi~\citep{Yoo:2025ReDi}.

The remaining challenge is to derive the closed-form backward velocity that governs this process. This can be obtained by following a construction analogous to~\Secref{subsec:closed_form_forward_velocity}. Specifically, we first derive the closed-form noise predictor $p_{0 \vert t}(x^i \vert z)$:
\begin{align}
    p_{0|t}(x^i|z) = \delta_{z}(x^i) - \frac{\kappa_t(K \delta_{x^i}(z^i) - 1)}{1 + (K-1)\kappa_t}  \frac{\sum_{m=1}^M  \delta_{d_m^i}(z^i)\gamma^{-h(d_m, z)}}{\sum_{m=1}^M \gamma^{-h(d_m, z)}},
    \label{eqn:closed_form_noise_predictor}
\end{align}
with a detailed derivation provided in App.~\ref{subsec:appx_proof_backward_velocity_discrete}. Substituting this into the definition of the backward velocity field from~\citet{Gat:2024DFM}
\begin{align}
    \label{eqn:backward_velocity}
    \check{v}_t(x^i, z) = \frac{\dot\kappa_t}{\kappa_t} \left[ \delta_z(x^i) - p_{0 \vert t} (x^i \vert z) \right],
\end{align}
we obtain the desired closed-form expression
\begin{align}
    \check{v}_t(x^i, z) = \frac{\dot{\kappa}_t(K \delta_{x^i}(z^i)-1)}{1 + (K-1)\kappa_t}  \frac{\sum_{m=1}^M \delta_{d_m^i}(z^i)\gamma^{-h(d_m, z)}}{\sum_{m=1}^M \gamma^{-h(d_m, z)}}.
    \label{eqn:closed_backward_velocity}
\end{align}

The second term in~\Eqref{eqn:closed_form_noise_predictor} computes the conditional likelihood of the $i$-th token taking value $x^i \in \mathcal{V}$ given the current sequence $z$, marginalized over all dataset $\{d_m\}_{m=1}^M$. The contribution of each data sample $d_m$ is determined by its proximity to $z$ under the Hamming distance $h(d_m, z)$, assigning higher weight to tokens with greater local consensus. Consequently, updating with $\check{v}_t(x^i, z)$ pushes the sample away from the data distribution and toward the source distribution $p_0(x)$.
Using $\check{v}_t(x^i, z)$, we construct pairs $\{(x_{0,m}, x_{1,m})\}_{m=1}^{M}$ by initializing from data points $\{d_{m}\}_{m=1}^M$ (equivalently, $\left\{x_{1,m} \right\}_{m=1}^M$) and iteratively applying the backward update rule in~\Eqref{eqn:sampling_step} for a fixed number of iterations $T$. The overall procedure is summarized in~\Algref{alg:coupling_backward}.
\vspace{-0.5\baselineskip}
\section{Experimental Results}
\label{sec:results}
\vspace{-0.5\baselineskip}
We validate the effectiveness of the proposed method and the source–target pairs it discovers across several discrete generative modeling benchmarks involving molecular data and images. We first summarize the experimental setup in~\Secref{subsec:experiment_setting}. In~\Secref{subsec:result_molecular} and~\Secref{subsec:result_image}, we compare our method against baselines in molecular and image generation, respectively. In~\Secref{subsec:result_distilation}, we further demonstrate that models trained with pairs discovered by our method not only achieve improved performance directly, but also benefit subsequent distillation phases, as the resulting base model provides a stronger initialization for existing distillation techniques.

\vspace{-0.5\baselineskip}
\subsection{Experiment Setup}
\label{subsec:experiment_setting}
\begin{table}[!t]
\small
\centering
\caption{Dataset and training statistics. $N$ denotes the number of tokens per sample, $K$ the dictionary size, $\lvert X_1 \rvert$ the dataset size, and $T_{*}$ the runtime of each method (in minutes, measured in wall-clock time with RTX A6000). The numbers in parentheses are the proportion of time relative to $T_{\text{Base}}$.} 
\label{tab:computation_cost_table} 
\vspace{-0.5\baselineskip}
\begin{tabularx}{\textwidth}{
    l|p{0.4cm}p{0.4cm}p{1.0cm}|p{2.0cm}p{1.7cm}p{1.7cm}p{2.2cm}
}
\toprule
Dataset & $N$ & $K$ & $|X_1|$ & $T_{\text{Base}}$ & $T_{\text{DCD}}$ & $T_{\text{ReDi}}$ & $T_{\methodname{}}$ \\
\midrule
MNIST-Binary & 768 & 2 & 60,000 & 80 (100.0\%) & 40 (50.0\%) & 49 (61.0\%) & 1.4 (1.7\%) \\
CIFAR-10 & 3,072 & 256 & 100,000 & 6720 (100.0\%) & 360 (5.3\%) & 468 (6.9\%)  & 20 (0.3\%) \\
QM9 & 32 & 40 & 127,190 & 450 (100.0\%) & 115 (24.8\%) & 100 (22.2\%) & 0.8 (0.2\%) \\
ZINC-250k & 72 & 74 & 224,568 & 1,110 (100.0\%) & 211 (19.0\%) & 194 (17.4\%) & 13 (1.2\%) \\
\bottomrule
\end{tabularx}
\vspace{-0.5\baselineskip}
\end{table}

\vspace{-0.5\baselineskip}
\paragraph{Baselines.} 
Across multiple benchmarks, we compare our approach against state-of-the-art discrete flow models, including MDLM~\citep{Sahoo:2024MDLM} and UDLM~\citep{Schiff:2025UDLM}. Since our method is based on a uniform source distribution, we adopt UDLM~\citep{Schiff:2025UDLM}, the leading uniform-state model, as our base  and denote UDLM trained with pairs generated by~\Algref{alg:coupling_backward} as~\ourname{} throughout the remainder of this section. In addition, we compare against these models augmented with distillation-based techniques that require additional finetuning, Discrete Consistency Distillation (DCD)~\citep{Sahoo:2025Duo} and ReDi~\citep{Yoo:2025ReDi}, denoted throughout this section by the suffixes ``+ DCD'' and ``+ ReDi''. 
The detailed training setups of these models, such as hyperparameters, are provided in App.~\ref{sec:appendix_results_detail}.
Additionally, we report the performance of the same base model trained on pairs formed by each data point and a source sample randomly drawn from the uniform distribution with our detailed experimental results in App.~\ref{sec:appendix_results_full}.

\vspace{-0.5\baselineskip}
\paragraph{Benchmarks.}
We evaluate our method across a diverse set of discrete generative modeling benchmarks, covering both molecule and image generation tasks. For molecule generation, we experiment with the QM9~\citep{ramakrishnan2014quantum} and ZINC-250k~\citep{irwin2012zinc} datasets. For image generation, we use the MNIST dataset~\citep{lecun2002gradient} with binarized pixel values (denoted MNIST-Binary) and the CIFAR-10 dataset~\citep{krizhevsky2009learning}, where pixel intensities are scaled to 8-bit integers, and horizontal flip augmentation is applied. Dataset statistics, including sample size, vocabulary size, and overall dataset size, are summarized in~\Tabref{tab:computation_cost_table}.
\vspace{-0.5\baselineskip}
\paragraph{Evaluation Setup.}
For molecular generation, we follow~\citet{Schiff:2025UDLM} and evaluate the validity, uniqueness, and novelty of generated molecules. Specifically, we sample 1,024 SMILES strings~\citep{weininger1988smiles}, convert them into molecular graphs, and compute these metrics. All results are averaged over 10 trials, with further details provided in App.~\ref{sec:appendix_results_full}. For image generation, we report Fréchet Inception Distance (FID)~\citep{Heusel:2017FID} and Inception Score (IS)~\citep{Salimans:2016IS}. FID is computed with 1,000 images for MNIST-Binary, and both FID and IS are computed with 5,000 generated images for CIFAR-10. The training dataset is used as the reference for FID computation.
Across all experiments, we vary the number of sampling steps to evaluate performance in both low- and high-NFE settings. In particular, we generate samples using $1-64$ steps for molecular benchmarks (QM9 and ZINC-250k) and MNIST-Binary benchmark, and $8-1024$ steps for CIFAR-10~\citep{krizhevsky2009learning}, as models yielded excessively high FIDs under extremely low-step settings.


\subsection{Molecule Generation}
\label{subsec:result_molecular}
\begin{figure}[t!]
    \centering
    \begin{minipage}[t]{0.31\textwidth}
        \centering
        \includegraphics[width=\textwidth]{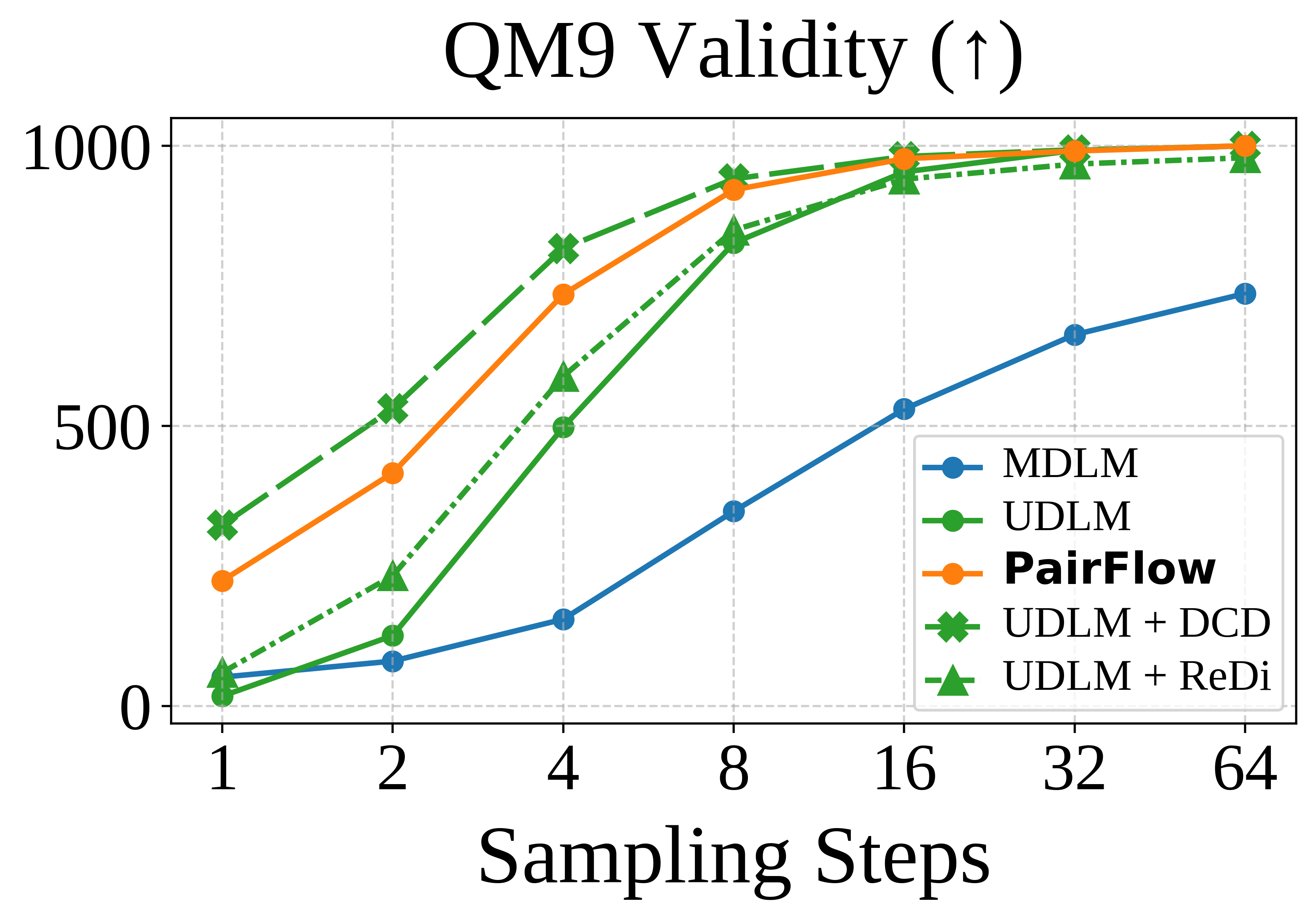}
    \end{minipage}
    \hfill
    \begin{minipage}[t]{0.31\textwidth}
        \centering
        \includegraphics[width=\textwidth]{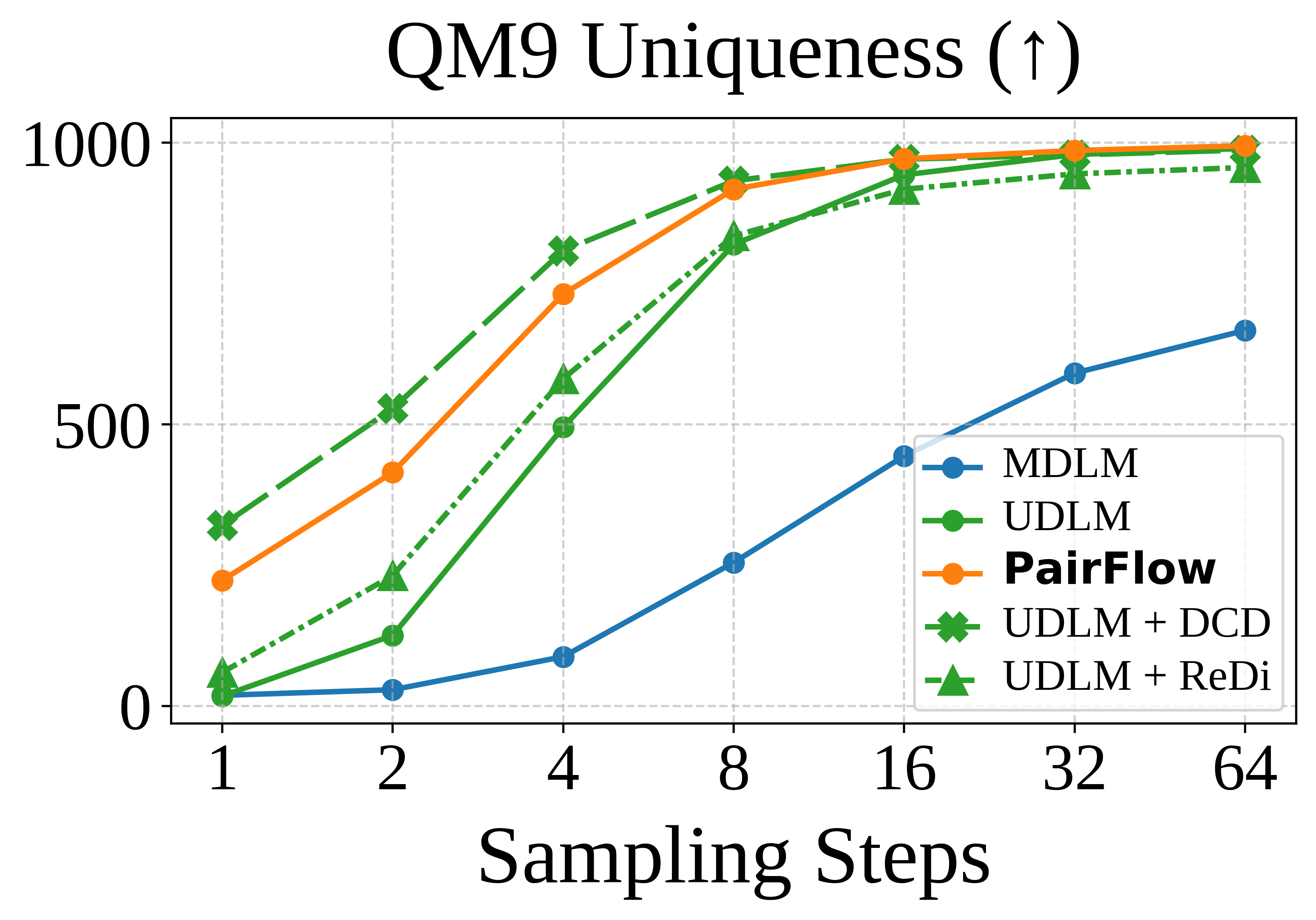}
    \end{minipage}
    \hfill
    \begin{minipage}[t]{0.31\textwidth}
        \centering
        \includegraphics[width=\textwidth]{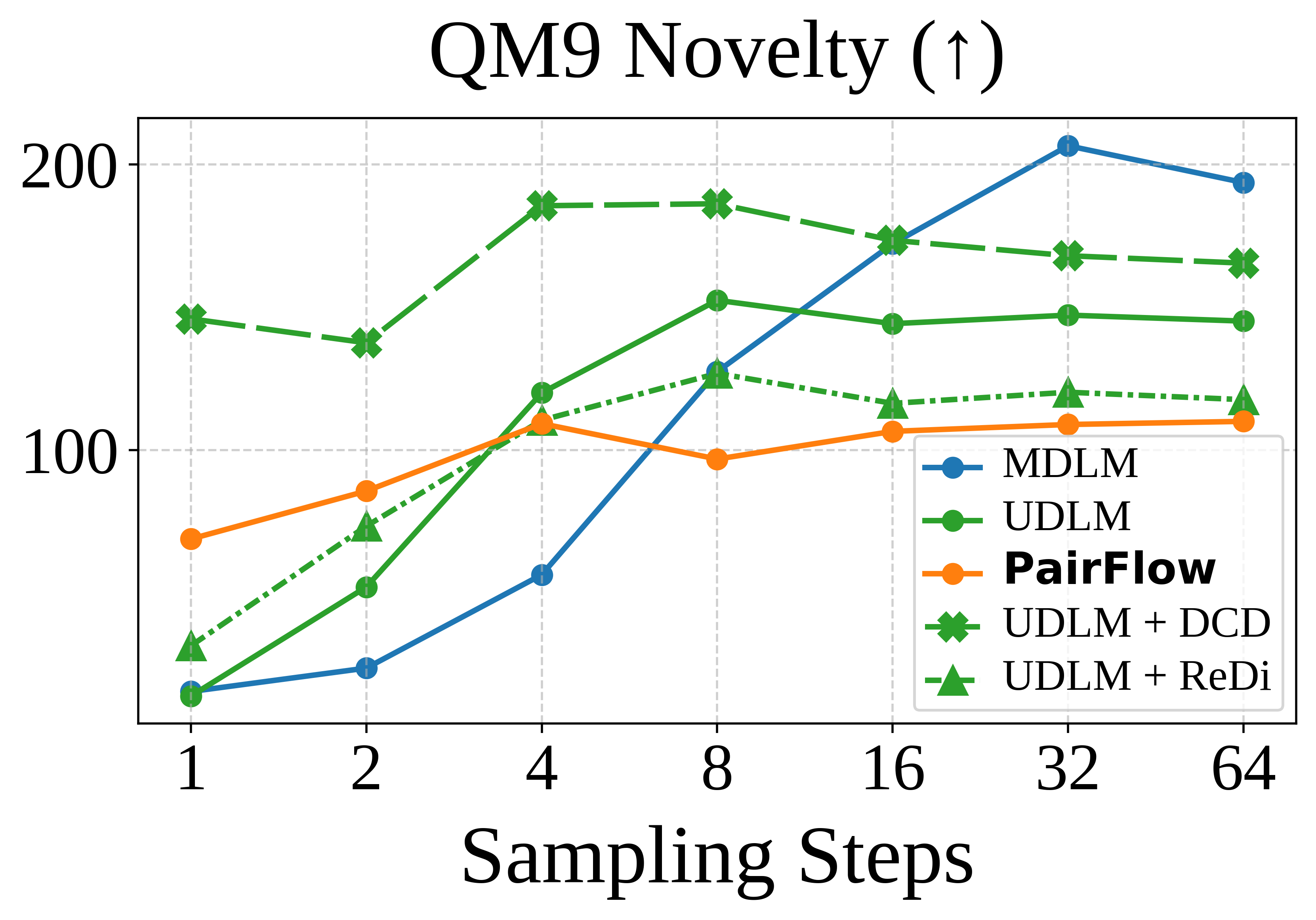}
    \end{minipage}
    \vspace{-\baselineskip}
    \caption{Step-wise performance analysis on the QM9 dataset~\citep{ramakrishnan2014quantum}. Each plot reports the number of valid (left), unique (middle), and novel (right) SMILES strings~\citep{weininger1988smiles} out of 1,024 generated samples. Best viewed when zoomed in.}
    \label{fig:qm9_discrete_plots}
\end{figure}

\begin{figure}[t!]
    \centering
    \begin{minipage}[t]{0.31\textwidth}
        \centering
        \includegraphics[width=\textwidth]{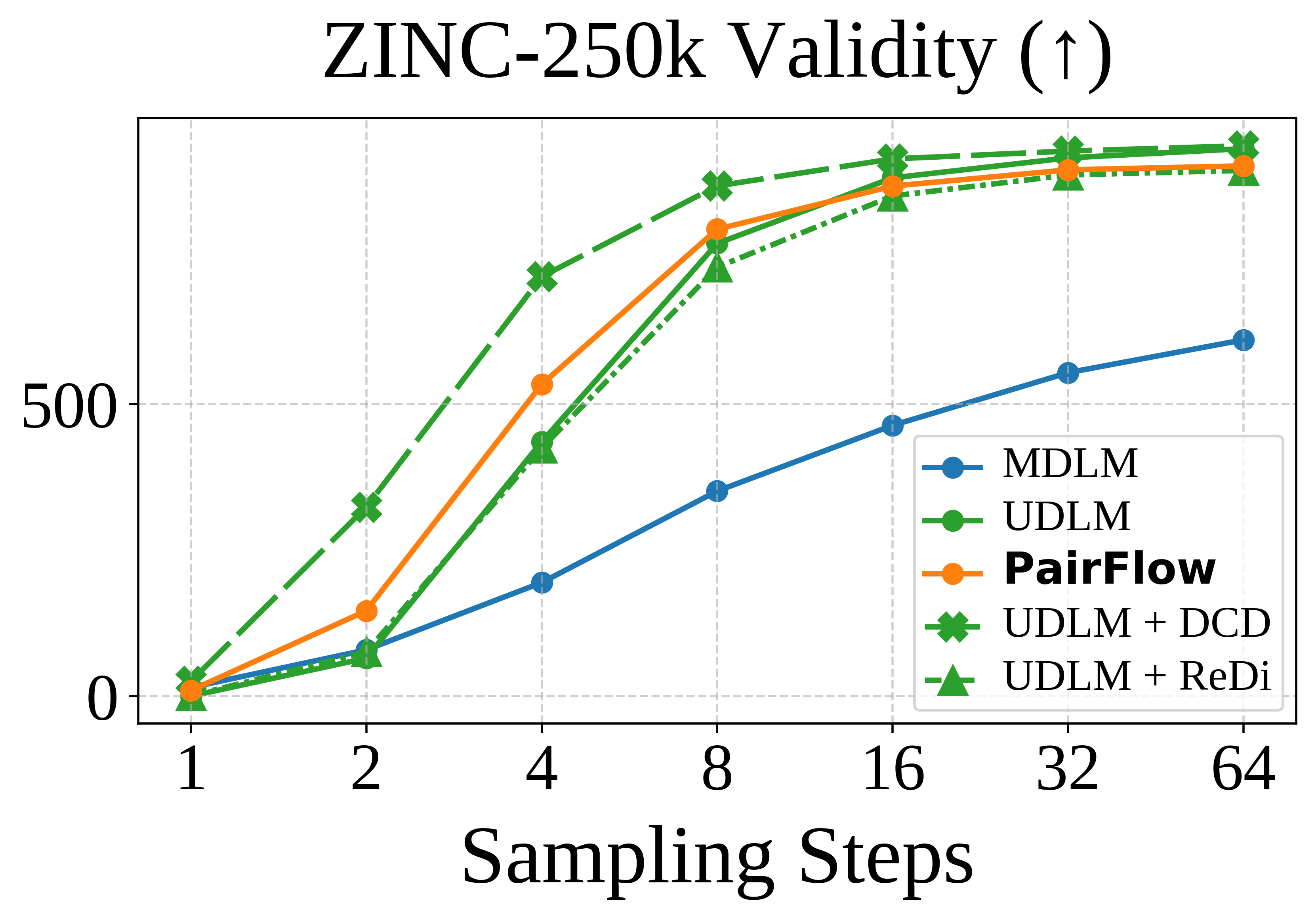}
    \end{minipage}
    \hfill
    \begin{minipage}[t]{0.31\textwidth}
        \centering
        \includegraphics[width=\textwidth]{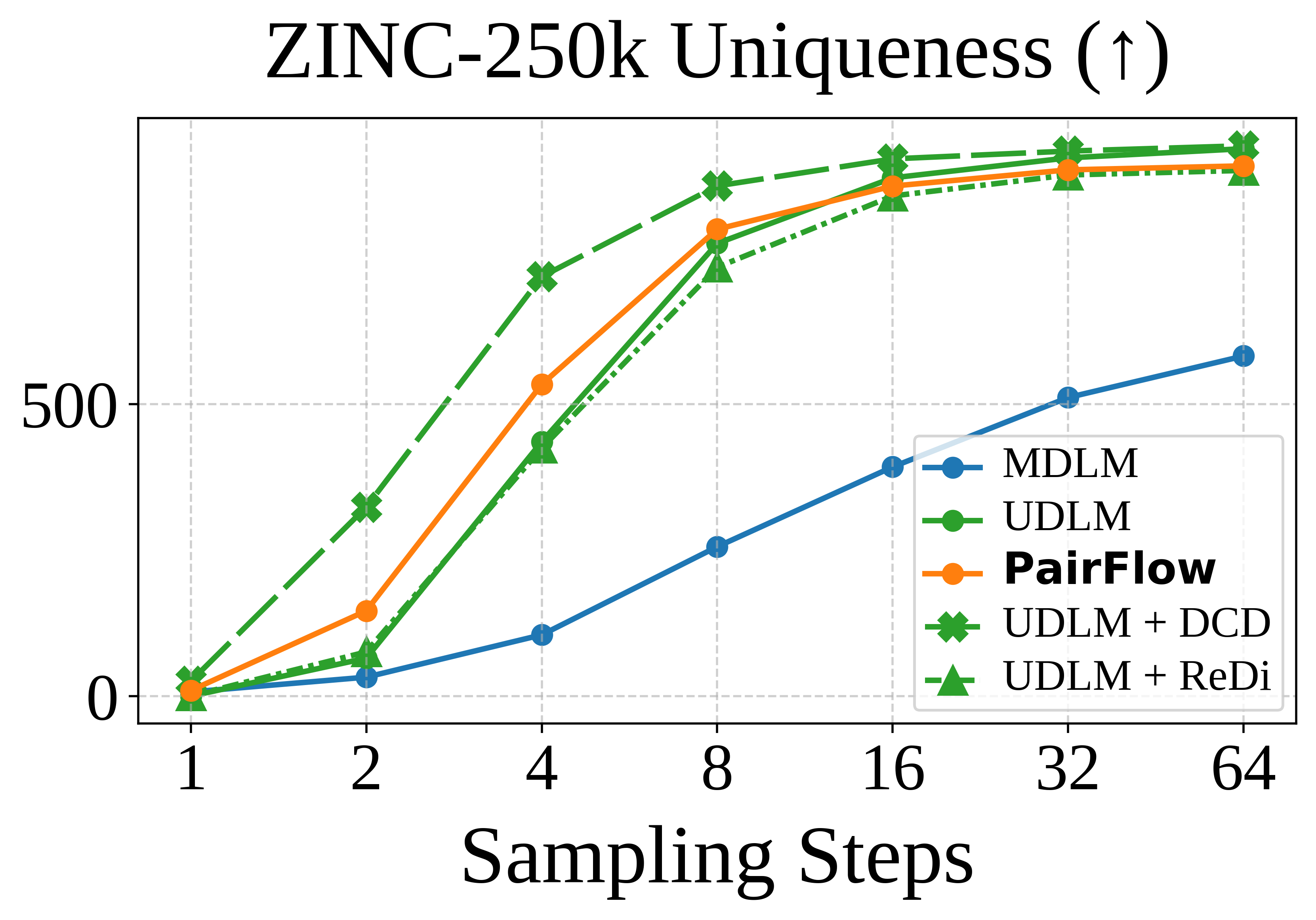}
    \end{minipage}
    \hfill
    \begin{minipage}[t]{0.31\textwidth}
        \centering
        \includegraphics[width=\textwidth]{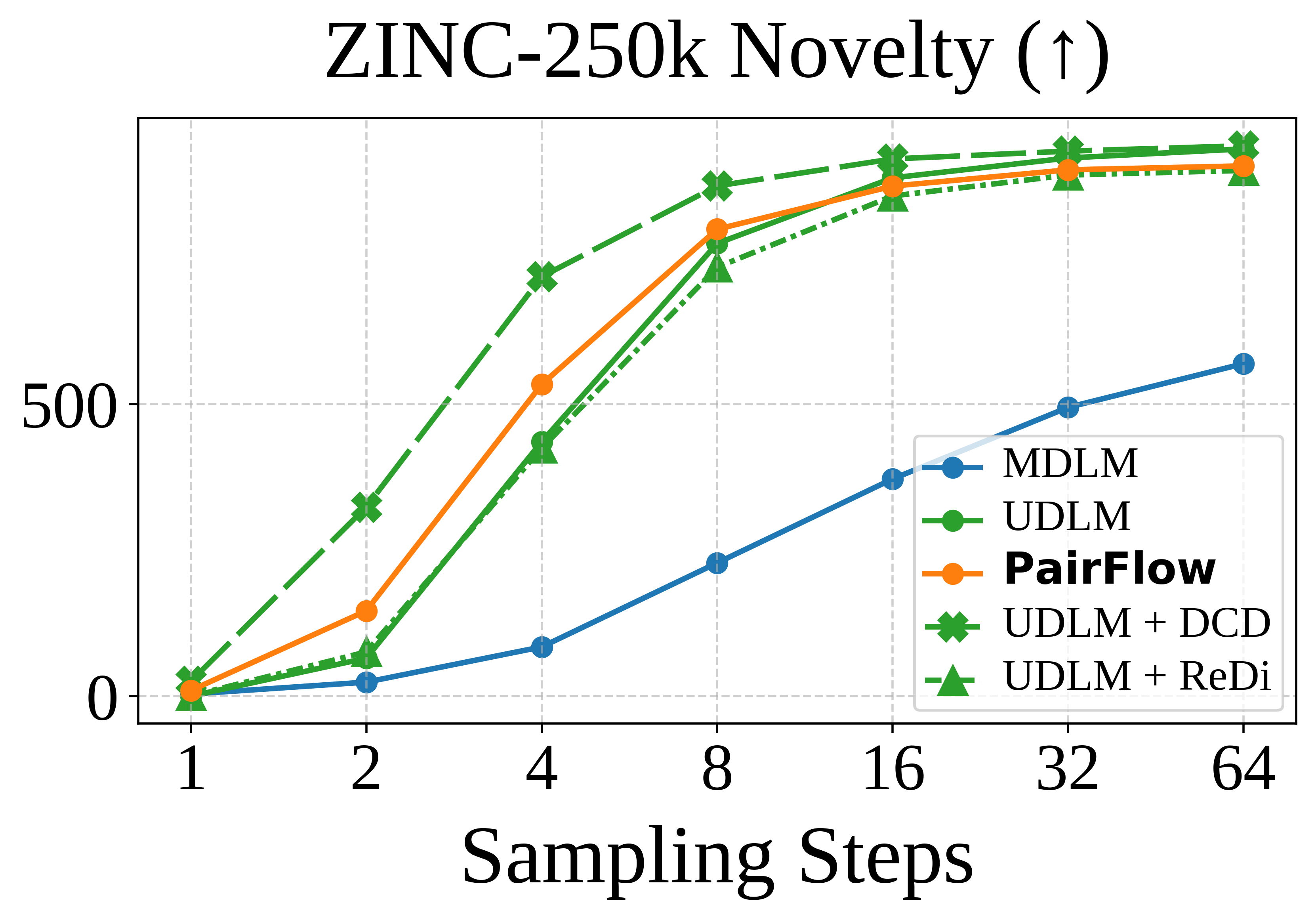}
    \end{minipage}
    \caption{Step-wise performance analysis on the ZINC-250k dataset~\citep{irwin2012zinc}. Each plot reports the number of valid (left), unique (middle), and novel (right) SMILES strings~\citep{weininger1988smiles} out of 1,024 generated samples. Best viewed when zoomed in.}
    \vspace{-\baselineskip}
    \label{fig:zinc250k_discrete_plots}
\end{figure}

\vspace{-0.5\baselineskip}
We begin by benchmarking unconditional molecule generation, where models are tasked with generating SMILES strings~\citep{weininger1988smiles} that represent molecules. As illustrated in~\Figref{fig:qm9_discrete_plots} and~\Figref{fig:zinc250k_discrete_plots}, which summarize validity (left), uniqueness (middle), and novelty (right),~\ourname{} consistently improves upon its base model UDLM~\citep{Schiff:2025UDLM}, yielding substantial gains in few-step settings.
It facilitates 1-step generation on QM9~\citep{ramakrishnan2014quantum}, a challenging setting that requires capturing all token-wise dependencies simultaneously. In this case, validity increases from $17.5$ to $223.4$, corresponding to a $12.8 \times$ improvement.
Similar trends are observed in the 2-step and 4-step settings, with validity improving by $231\%$ and $47.6\%$, respectively. As shown in~\Figref{fig:qm9_discrete_plots} (left), this improvement is particularly significant: the 2-step and 4-step validities of~\ourname{} are comparable to the 4-step and 8-step validities achieved by UDLM~\citep{Schiff:2025UDLM}.
Comparable improvements are also seen on the ZINC-250k~\citep{irwin2012zinc} dataset.

Remarkably,~\textbf{\ourname{} introduces minimal overhead—less than $2\%$ of the training cost as shown in~\Tabref{tab:computation_cost_table}—and requires no pretrained models, yet achieves performance comparable to, and in some cases surpassing, models distilled from the same base using DCD~\citep{Sahoo:2025Duo} and ReDi~\citep{Yoo:2025ReDi}}, both of which rely on pretrained models and finetuning. On both QM9~\citep{ramakrishnan2014quantum} and ZINC-250k~\citep{irwin2012zinc},~\ourname{} consistently outperforms $\text{UDLM + ReDi}$ across all few-step settings, achieving substantially higher 2-step validities on QM9 ($232.4$ vs.\ $416.0$) and ZINC-250k ($75.9$ vs.\ $146.3$). At the same time,~\ourname{} matches the performance of $\text{UDLM + DCD}$, with comparable 2-step validities on QM9 ($416.0$ vs.\ $530.8$). This is particularly notable given that the additional preprocessing cost of~\ourname{} amounts to only $0.69\%$ on QM9~\citep{ramakrishnan2014quantum} and $6.16\%$ on ZINC-250k~\citep{irwin2012zinc}, relative to the full cost of DCD~\citep{Sahoo:2025Duo}. Detailed numerical results with standard deviations are reported in App.~\ref{sec:appendix_results_full}.  


\vspace{-0.25\baselineskip}
\subsection{Image Generation}
\label{subsec:result_image}
\vspace{-0.5\baselineskip}
\begin{figure}[t!]
    \centering
    \begin{minipage}[t]{0.31\textwidth} 
        \centering
        \includegraphics[width=\textwidth]{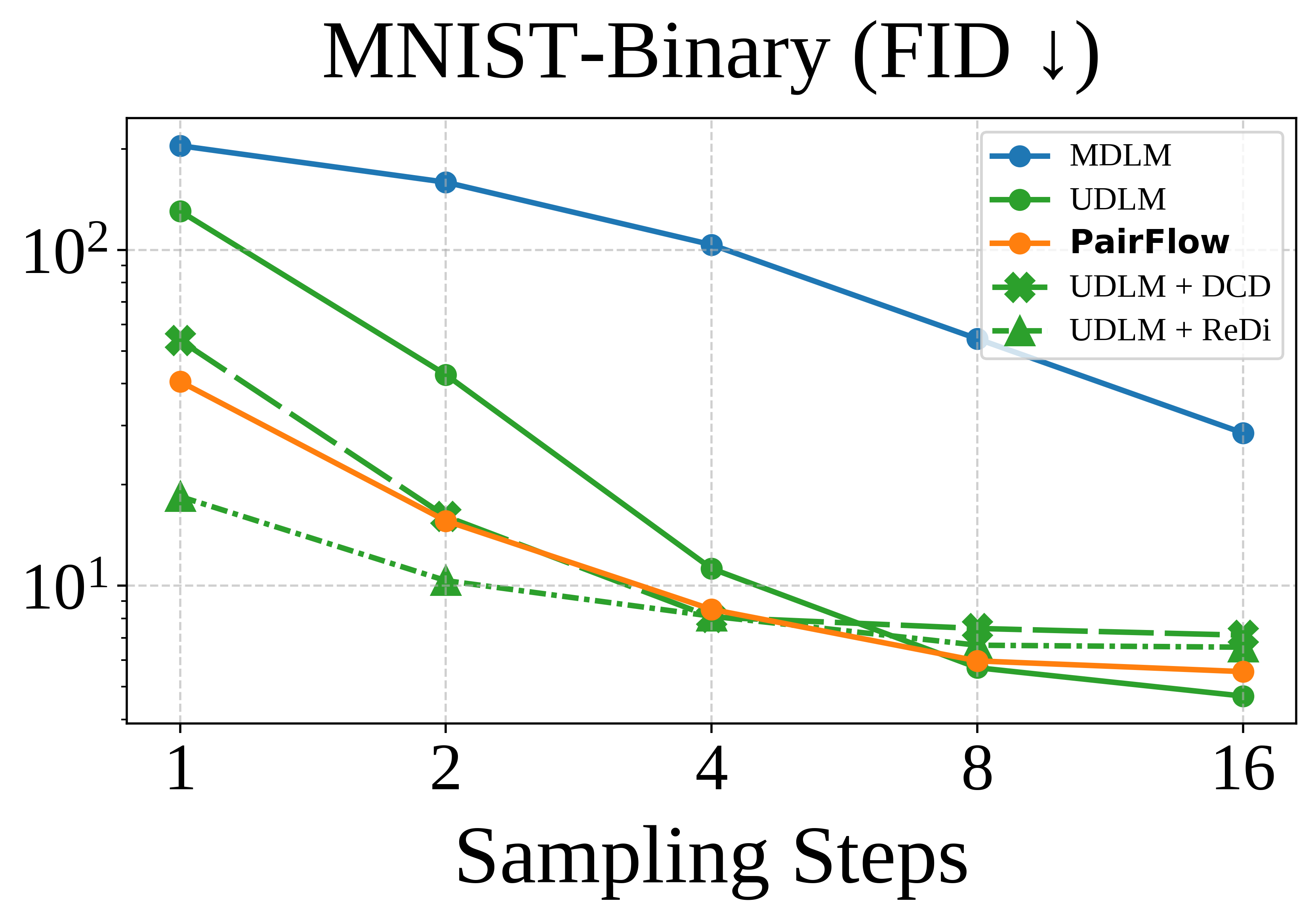}
    \end{minipage}
    \hfill 
    \vrule width 0.4pt
    \hfill
    \begin{minipage}[t]{0.31\textwidth} 
        \centering
        \includegraphics[width=\textwidth]{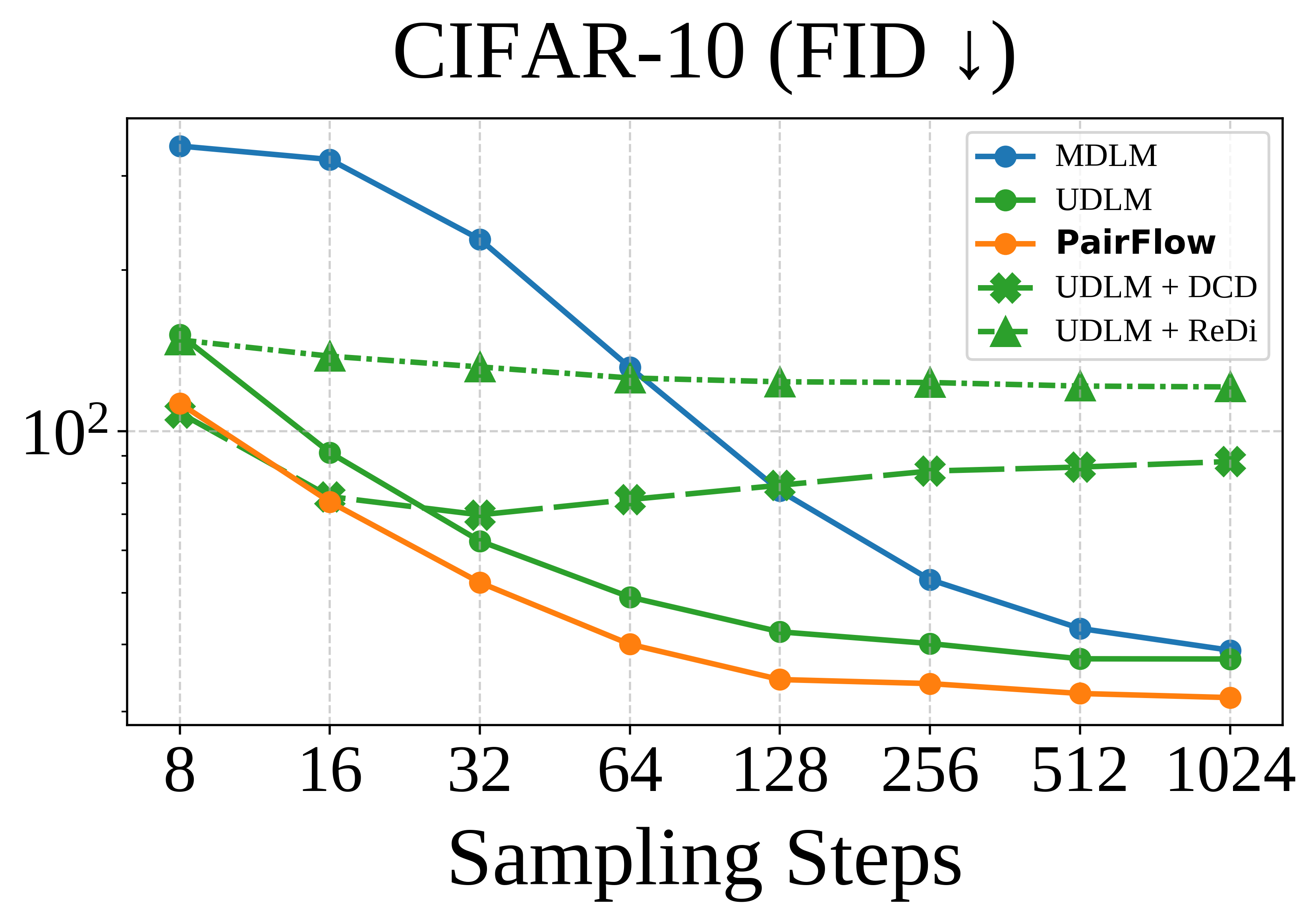}
    \end{minipage}
    \hfill 
    \begin{minipage}[t]{0.31\textwidth} 
        \centering
        \includegraphics[width=\textwidth]{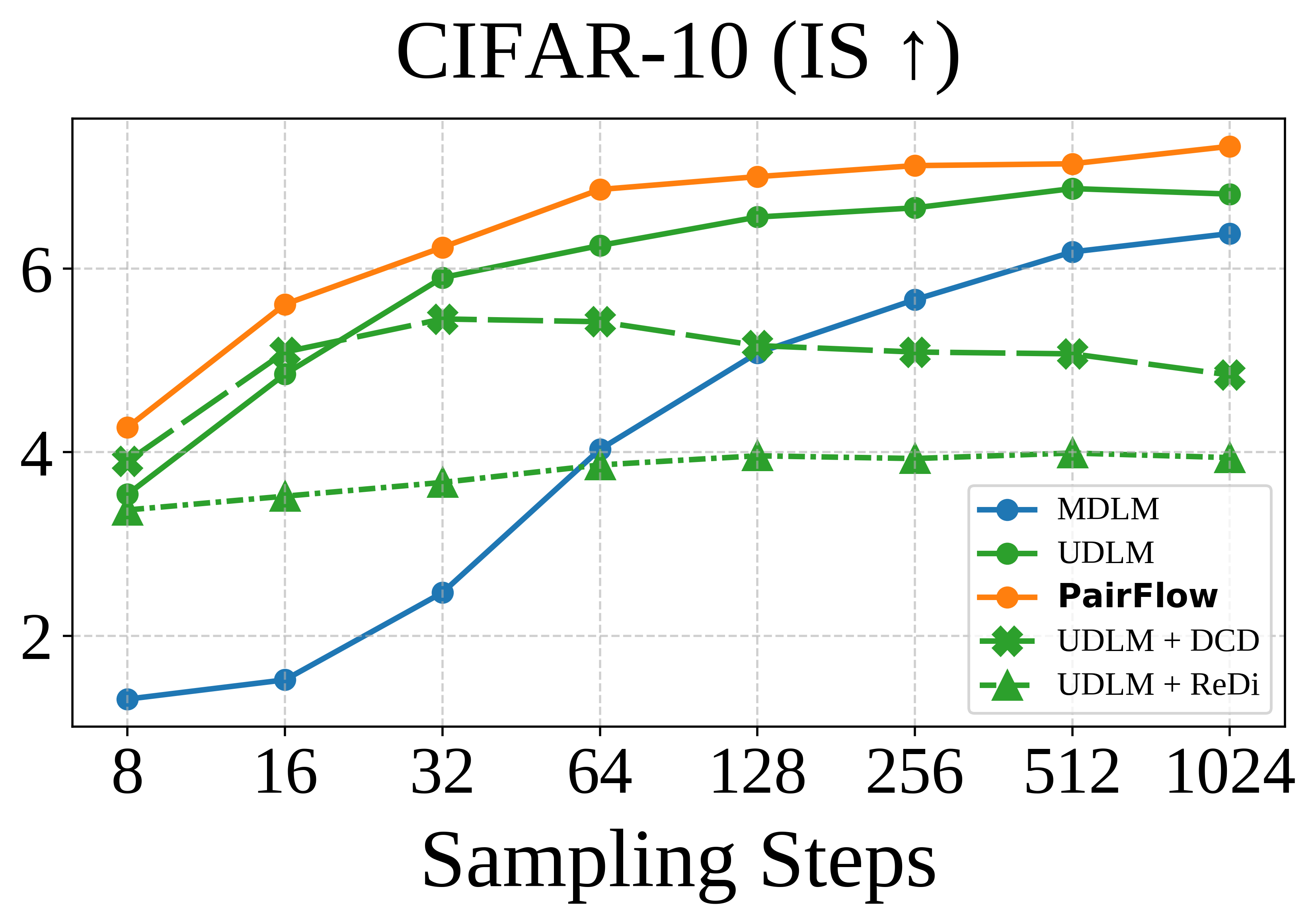}
    \end{minipage} 
    \caption{Step-wise performance analysis on discretized image datasets. From left to right: FID on MNIST-Binary~\citep{lecun2002gradient}, FID on CIFAR-10~\citep{krizhevsky2009learning}, and Inception Scores (IS) on CIFAR-10. Best viewed when zoomed in.}
    \label{fig:image_discrete_plots}
\end{figure}
\begin{figure}[t!]
\centering
\setlength{\tabcolsep}{0.4pt}
\renewcommand{\arraystretch}{0.3}
\begin{minipage}{0.41\textwidth}
\centering
\begin{tabular}{c @{\hskip 2.0pt} cccc}
\raisebox{0.35\height}{\rotatebox{90}{\scriptsize MDLM}} &
\includegraphics[width=0.22\linewidth]{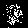} &
\includegraphics[width=0.22\linewidth]{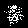} &
\includegraphics[width=0.22\linewidth]{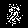} &
\includegraphics[width=0.22\linewidth]{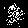} \\
\raisebox{0.35\height}{\rotatebox{90}{\scriptsize UDLM}} &
\includegraphics[width=0.22\linewidth]{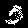} &
\includegraphics[width=0.22\linewidth]{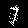} &
\includegraphics[width=0.22\linewidth]{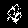} &
\includegraphics[width=0.22\linewidth]{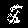} \\
\raisebox{0.01\height}{\rotatebox{90}{\scriptsize \textbf{\ourname{}}}} &
\includegraphics[width=0.22\linewidth]{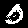} &
\includegraphics[width=0.22\linewidth]{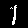} &
\includegraphics[width=0.22\linewidth]{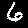} &
\includegraphics[width=0.22\linewidth]{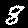} \\
\raisebox{0.35\height}{\rotatebox{90}{\scriptsize \makecell{UDLM \\ + DCD}}} &
\includegraphics[width=0.22\linewidth]{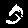} &
\includegraphics[width=0.22\linewidth]{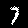} &
\includegraphics[width=0.22\linewidth]{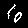} &
\includegraphics[width=0.22\linewidth]{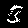} \\
\raisebox{0.01\height}{\rotatebox{90}{\scriptsize \makecell{\textbf{\ourname{}} \\ + DCD}}} &
\includegraphics[width=0.22\linewidth]{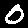} &
\includegraphics[width=0.22\linewidth]{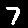} &
\includegraphics[width=0.22\linewidth]{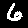} &
\includegraphics[width=0.22\linewidth]{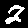} \\
\raisebox{0.35\height}{\rotatebox{90}{\scriptsize \makecell{UDLM \\ + ReDi}}} &
\includegraphics[width=0.22\linewidth]{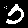} &
\includegraphics[width=0.22\linewidth]{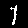} &
\includegraphics[width=0.22\linewidth]{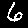} &
\includegraphics[width=0.22\linewidth]{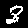} \\
\raisebox{0.01\height}{\rotatebox{90}{\scriptsize \makecell{\textbf{\ourname{}} \\ + ReDi}}} &
\includegraphics[width=0.22\linewidth]{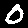} &
\includegraphics[width=0.22\linewidth]{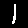} &
\includegraphics[width=0.22\linewidth]{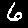} &
\includegraphics[width=0.22\linewidth]{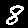} \\
\end{tabular}
\end{minipage}
\hfill
\begin{minipage}{0.575\textwidth}
\centering
\begin{tabular}{!{\vrule width 1.2pt} @{\hskip 2.0pt} c @{\hskip 2.0pt} c @{\hskip 2.0pt} ccccc}
\multirow{3}{*}{\raisebox{-1.0\height}{\rotatebox{90}{\textbf{64 steps}}}} &
\raisebox{0.4\height}{\rotatebox{90}{\scriptsize MDLM}} &
\includegraphics[width=0.18\linewidth]{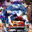} &
\includegraphics[width=0.18\linewidth]{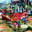} &
\includegraphics[width=0.18\linewidth]{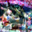} &
\includegraphics[width=0.18\linewidth]{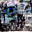} &
\includegraphics[width=0.18\linewidth]{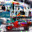} \\
& \raisebox{0.5\height}{\rotatebox{90}{\scriptsize UDLM}} &
\includegraphics[width=0.18\linewidth]{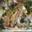} &
\includegraphics[width=0.18\linewidth]{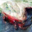} &
\includegraphics[width=0.18\linewidth]{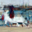} &
\includegraphics[width=0.18\linewidth]{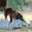} &
\includegraphics[width=0.18\linewidth]{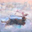} \\
& \raisebox{0.1\height}{\rotatebox{90}{\scriptsize \textbf{\ourname{}}}} &
\includegraphics[width=0.18\linewidth]{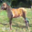} &
\includegraphics[width=0.18\linewidth]{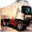} &
\includegraphics[width=0.18\linewidth]{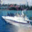} &
\includegraphics[width=0.18\linewidth]{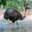} &
\includegraphics[width=0.18\linewidth]{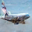} \\ [-0.1em]
\specialrule{1.2pt}{2.5pt}{2.3pt}
\multirow{3}{*}{\raisebox{-0.9\height}{\rotatebox{90}{\textbf{256 steps}}}} &
\raisebox{0.4\height}{\rotatebox{90}{\scriptsize MDLM}} &
\includegraphics[width=0.18\linewidth]{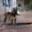} &
\includegraphics[width=0.18\linewidth]{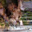} &
\includegraphics[width=0.18\linewidth]{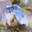} &
\includegraphics[width=0.18\linewidth]{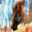} &
\includegraphics[width=0.18\linewidth]{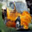} \\
& \raisebox{0.5\height}{\rotatebox{90}{\scriptsize UDLM}} &
\includegraphics[width=0.18\linewidth]{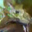} &
\includegraphics[width=0.18\linewidth]{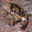} &
\includegraphics[width=0.18\linewidth]{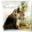} &
\includegraphics[width=0.18\linewidth]{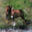} &
\includegraphics[width=0.18\linewidth]{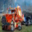} \\
& \raisebox{0.1\height}{\rotatebox{90}{\scriptsize \textbf{\ourname{}}}} &
\includegraphics[width=0.18\linewidth]{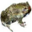} &
\includegraphics[width=0.18\linewidth]{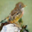} &
\includegraphics[width=0.18\linewidth]{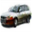} &
\includegraphics[width=0.18\linewidth]{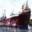} &
\includegraphics[width=0.18\linewidth]{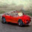} \\
 
\end{tabular}

\end{minipage}
\vspace{-0.7em}
\caption{Qualitative results of 1-step generation on MNIST-Binary ($28 \times 28$; left) and 64-step (top right) and 256-step (bottom right) generation on CIFAR-10 ($32 \times 32$).}
\vspace{-\baselineskip}
\label{fig:image_qualitative}
\end{figure}

We further extend our experiments to image domains where each pixel has discretized intensities. As in~\Secref{subsec:result_molecular}, we evaluate model performance across multiple sampling steps and summarize the results in~\Figref{fig:image_discrete_plots}. Qualitative samples for MNIST-Binary and CIFAR-10 are shown in~\Figref{fig:image_qualitative}. Both qualitative and quantitative results show that~\ourname{} improves the performance of UDLM~\citep{Schiff:2025UDLM} and, in few-step settings, achieves performance comparable to DCD~\citep{Sahoo:2025Duo}.

On MNIST-Binary (\Figref{fig:image_discrete_plots}, left),~\ourname{} achieves an FID of $40.59$ in the 1-step setting, equivalently a $68.9\%$ improvement over UDLM~\citep{Schiff:2025UDLM}. Consistent gains are observed across other few-step settings as well: at 2 steps, FID is reduced by $63.3\%$ ($15.61$ vs.\ $42.54$), and at 4 steps by $24.4\%$ ($8.51$ vs.\ $11.25$). On CIFAR-10~\citep{krizhevsky2009learning}, where FID~\citep{Heusel:2017FID} and IS~\citep{Salimans:2016IS} are reported in~\Figref{fig:image_discrete_plots} (middle and right),~\ourname{} likewise outperforms the base UDLM, validating the effectiveness of the discovered source–target pairs.

As in molecular generation (\Secref{subsec:result_molecular}),~\textbf{\ourname{} performs comparable to distillation-based acceleration methods.} On MNIST-Binary, it achieves a lower FID than $\text{UDLM+DCD}$ in the 1-step setting ($40.59$ vs.\ $53.84$) and comparable performance at 2 steps ($15.61$ vs.\ $16.09$). Likewise,~\ourname{} performs competitively with $\text{UDLM+ReDi}$ at 2 steps ($15.61$ vs.\ $10.36$), while requiring substantially less compute than both. As summarized in~\Tabref{tab:computation_cost_table}, DCD~\citep{Sahoo:2025Duo} requires 40 minutes ($T_{\text{DCD}}$) and ReDi~\citep{Yoo:2025ReDi} takes 49 minutes, whereas the preprocessing phase of~\ourname{} completes in just $1.4$ minutes ($T_{\text{\ourname{}}}$), yielding $28.6\times$ and $35\times$ speedups, respectively.
On the CIFAR-10 benchmark, both DCD~\citep{Sahoo:2025Duo} and ReDi~\citep{Yoo:2025ReDi} degrade model performance, as indicated by the higher FID in~\Figref{fig:image_discrete_plots} (middle) and lower IS in~\Figref{fig:image_discrete_plots} (right). The results in~\Tabref{tab:cifar10_results} suggest that, overall, acceleration methods do not work well on CIFAR-10. We hypothesize that this issue arises from the low performance of the teacher model, which negatively affects the student model when applying acceleration methods. Detailed results are reported in App.~\ref{sec:appendix_results_full}.  

\vspace{-0.25\baselineskip}
\subsection{Distilling Models Trained with Aligned Pairs}
\label{subsec:result_distilation}
\vspace{-0.5\baselineskip}
\begin{figure}[t!]
    \centering
    \begin{minipage}[t]{0.31\textwidth} 
        \centering
        \includegraphics[width=\textwidth]{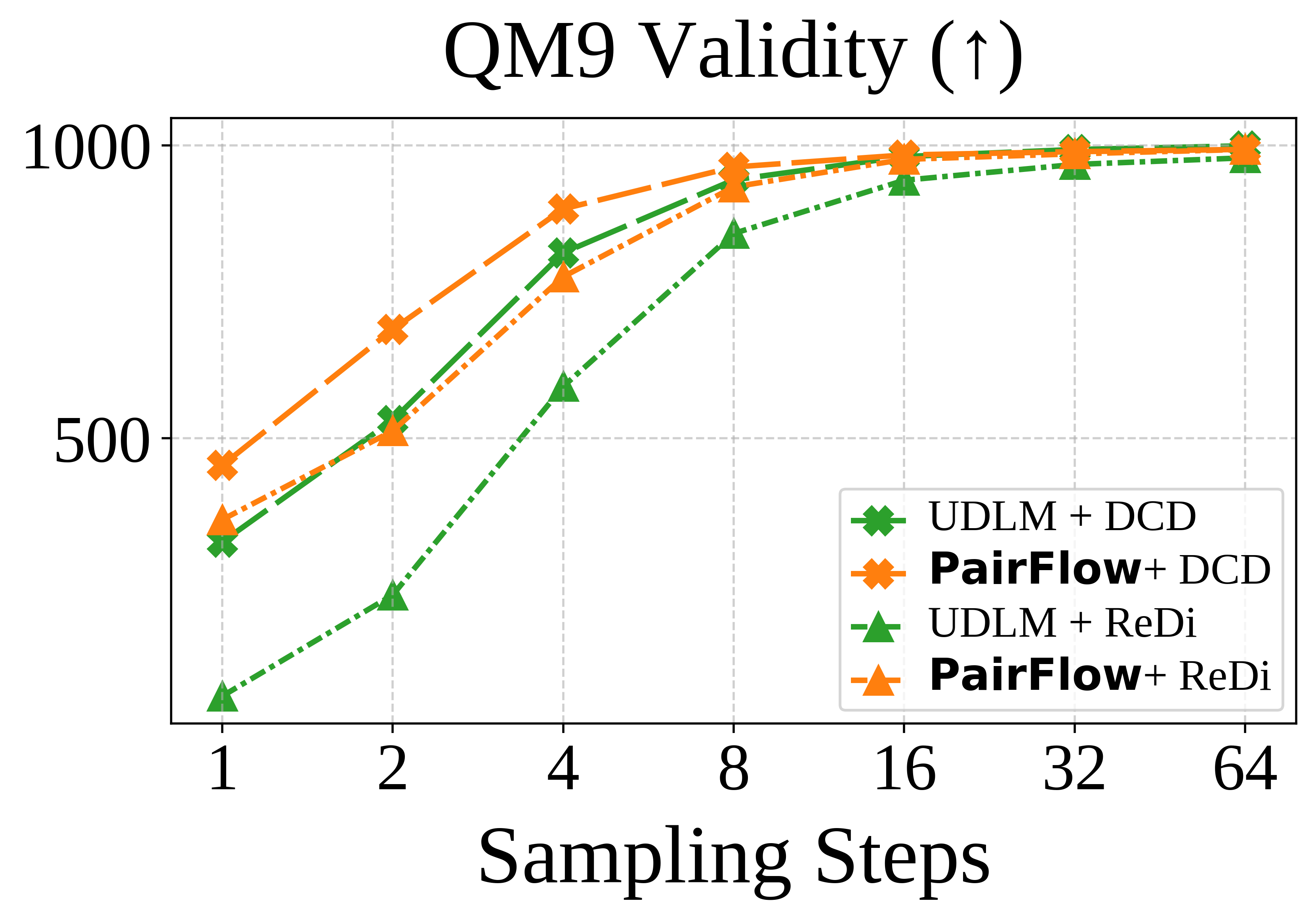}
    \end{minipage}
    \hfill 
    \begin{minipage}[t]{0.31\textwidth} 
        \centering
        \includegraphics[width=\textwidth]{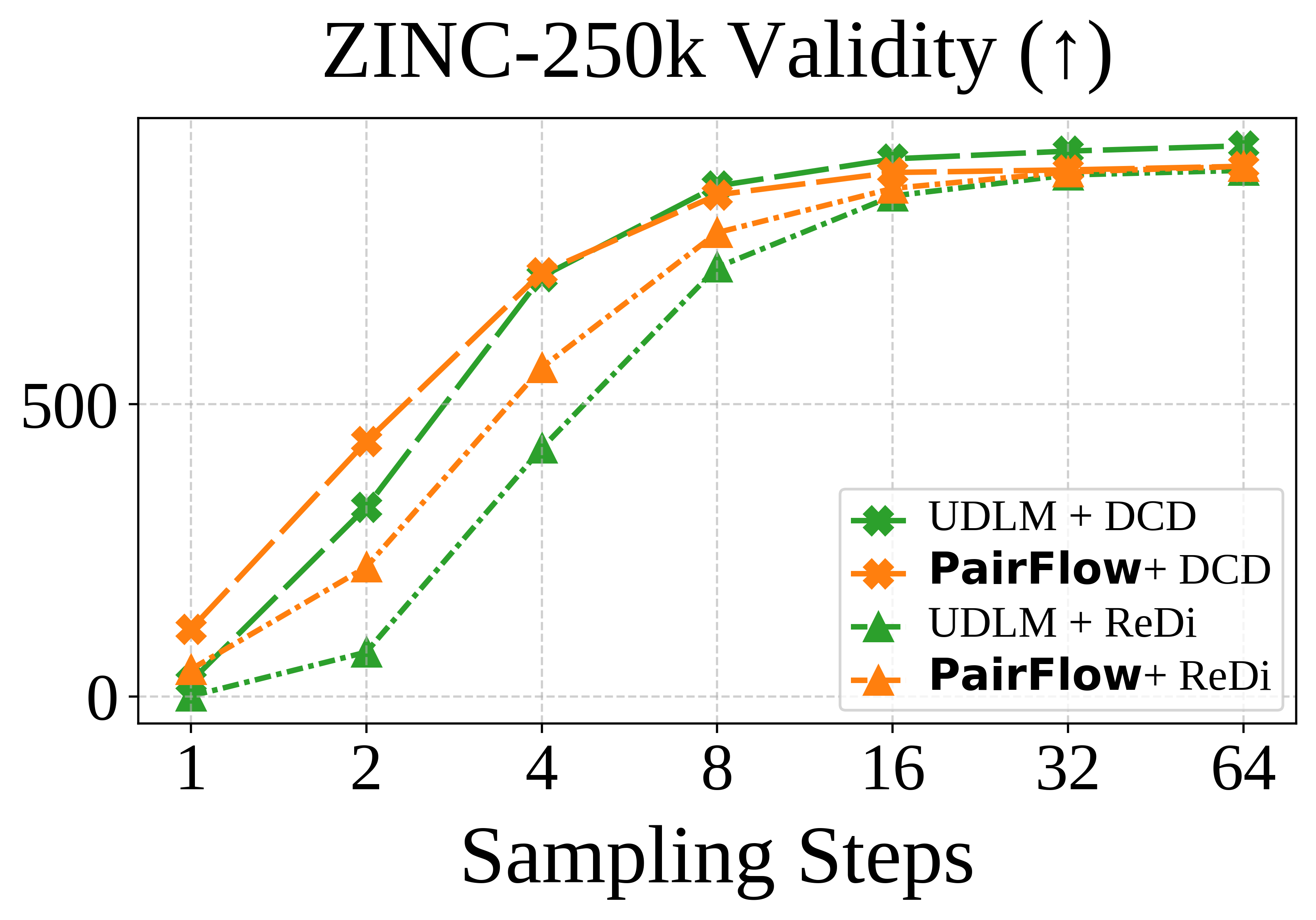}
    \end{minipage}
    \hfill 
    \begin{minipage}[t]{0.31\textwidth} 
        \centering
        \includegraphics[width=\textwidth]{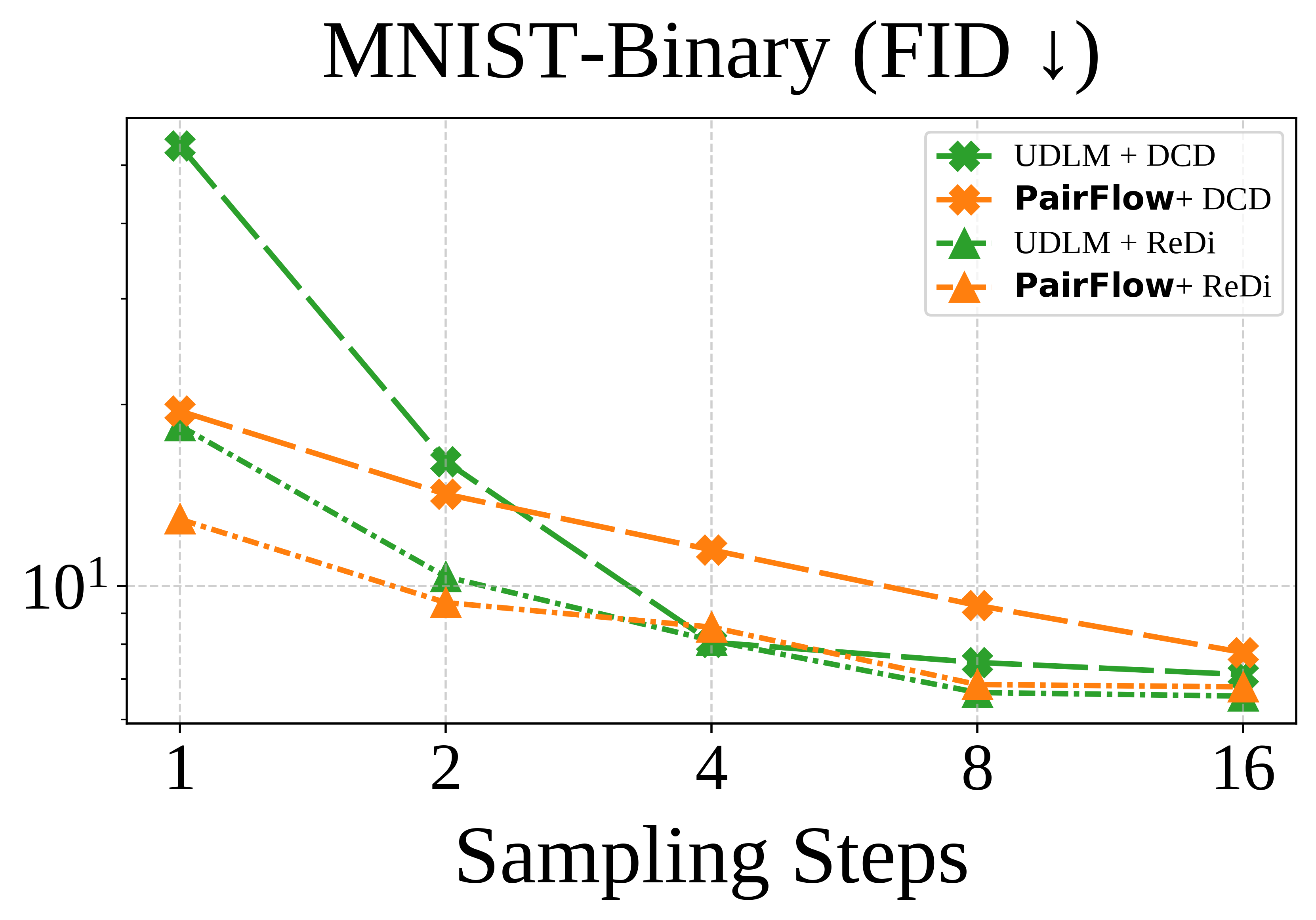}
    \end{minipage}
    \caption{Step-wise performance analysis of distilled models on molecular and image datasets. From left to right: number of valid molecules on QM9, number of valid molecules on ZINC-250k, and FID on MNIST-Binary.}
    \vspace{-\baselineskip} 
    \label{fig:image_distillation_plots}
\end{figure}

While~\ourname{} alone achieves performance comparable to, or even exceeding, distillation-based techniques~\citep{Yoo:2025ReDi,Sahoo:2025Duo}, as shown in~\Secref{subsec:result_molecular} and~\Secref{subsec:result_image}, we further emphasize that it also serves as a strong initialization for subsequent distillation, yielding even greater performance gains when combined with existing methods. Crucially, this incurs negligible additional cost relative to the overall time required for distillation.

We validate this by distilling~\ourname{}, trained on QM9~\citep{ramakrishnan2014quantum}, ZINC-250k~\citep{irwin2012zinc}, and MNIST-Binary, using DCD~\citep{Sahoo:2025Duo} and ReDi~\citep{Yoo:2025ReDi}, and comparing their performance against distilled models whose teachers were the base UDLM~\citep{Schiff:2025UDLM}.
As shown in~\Figref{fig:image_distillation_plots},~\textbf{student models distilled from~\ourname{}, denoted~\ourname{}+DCD and~\ourname{}+ReDi, push the frontier of performance previously achieved by distillation-based techniques.} For example, on the QM9 dataset~\citep{ramakrishnan2014quantum},~\ourname{}+DCD substantially improves validity over UDLM+DCD ($453.8$ vs.\ $323$ for 1 step, $685.8$ vs.\ $530.8$ for 2 steps). A similar trend is observed for~\ourname{}+ReDi on ZINC-250k~\citep{irwin2012zinc}, yielding higher scores in both 1-step ($46.3$ vs.\ $0.7$) and 2-step ($221.5$ vs.\ $75.9$) generation. Importantly, as summarized in~\Tabref{tab:computation_cost_table}, these gains are achieved at only minimal additional preprocessing cost: $3.15\%$ of the average runtime of distillation on MNIST-Binary, $0.77\%$ on QM9, and $6.42\%$ on ZINC-250k.

\vspace{-0.75\baselineskip}
\section{Conclusion}
\label{sec:conclusion}
\vspace{-0.75\baselineskip}
We have presented~\ourname{}, a novel approach to accelerating the generative process of Discrete Flow Models (DFMs) through a lightweight preprocessing step performed prior to training. Our preprocessing, which couples source and target samples, requires only up to 1.7\% of the base model training cost, making it at least 20× more efficient than finetuning, while still achieving comparable or even superior performance. The key enabler is the closed-form inversion, which eliminates the need for a pretrained teacher model.

\bibliography{main}
\bibliographystyle{main}


\clearpage
\appendix

\section{Proof for Closed-form Velocity}
\label{sec:appendix_proof}

In this section, we present the detailed derivations of the closed-form forward velocity~\Eqref{eqn:closed_forward_velocity} and backward velocity~\Eqref{eqn:closed_backward_velocity} introduced in~\Secref{sec:background}. \Secref{subsec:appx_proof_forward_velocity_discrete} provides the proof of the closed-form forward velocity, while \Secref{subsec:appx_proof_backward_velocity_discrete} presents the proof of the closed-form backward velocity. Both derivations are based on the assumption of a uniform source distribution.

\subsection{Proof of Closed-form Forward Velocity in Discrete Flow Models}
\label{subsec:appx_proof_forward_velocity_discrete}

Let $x, z \in \mathcal{V}^N$ be sequences of tokens $x^i, z^i \in \mathcal{V}$ for $i \in {1, \dots, N}$, where each token takes values from the discrete vocabulary $\mathcal{V} = \{1, \dots, K\}$.
We begin with the expression of the forward velocity given in~\Eqref{eqn:closed_forward_velocity}:
\begin{align}
    \hat{v}_t(x^i, z) = \frac{\dot{\kappa}_t}{1-\kappa_t}\left[p_{1|t}(x^i|z) - \delta_{z}(x^i)\right].
\end{align}

We first derive the closed-form expression for the probability denoiser $p_{1|t}(x^i  |  z)$:

Using Bayes' rule,
\begin{align}
p_{1|t}(x^i|z)
&= \sum_{x_0,x_1}\delta_{x_1^i}(x^i)\,p_t(x_0,x_1 |  z) \\
&= \sum_{x_0,x_1}\delta_{x_1^i}(x^i)\,\frac{p_t(x_0,x_1,z)}{p_t(z)} \\
&= \frac{\sum_{x_0,x_1}\delta_{x_1^i}(x^i)\,p_t(x_0,x_1,z)}{\sum_{x_0,x_1}p_t(x_0,x_1,z)}.
\label{eq:denoiser_ratio_app}
\end{align}
We factor the joint as
\begin{align}
p_t(x_0,x_1,z)\;&\propto\; p_0(x_0)\,p_1(x_1)\,p_t(z |  x_0,x_1) \\
p_t(z |  x_0,x_1)\;&=\;\prod_{j=1}^N\!\big[\kappa_t\,\delta_{x_1^j}(z^j)+(1-\kappa_t)\,\delta_{x_0^j}(z^j)\big]
\label{eq:kernel_app}
\end{align}

and use the empirical target
\begin{equation}
p_1(x_1)\;=\;\frac{1}{M}\sum_{m=1}^M \delta_{d_m}(x_1)
\quad\text{(\Eqref{eqn:empirical_target})}.
\label{eq:empirical_p1_app}
\end{equation}
Since $p_0(x_0)$ is constant, it cancels between the numerator and denominator of
\Eqref{eq:denoiser_ratio_app}, yielding
\begin{align}
p_{1 |  t}(x^i |  z)
&= \frac{\sum_{x_0,x_1}\delta_{x_1^i}(x^i)\,p_1(x_1)\,p_t(z| x_0,x_1)}
{\sum_{x_0,x_1}p_1(x_1)\,p_t(z |  x_0,x_1)} \nonumber\\
&= \frac{\sum_{m=1}^M \sum_{x_0}\delta_{d_m^i}(x^i)\,\prod_{j=1}^N\!\big[\kappa_t\,\delta_{d_m^j}(z^j)+(1-\kappa_t)\,\delta_{x_0^j}(z^j)\big]}
{\sum_{m=1}^M \sum_{x_0}\prod_{j=1}^N\!\big[\kappa_t\,\delta_{d_m^j}(z^j)+(1-\kappa_t)\,\delta_{x_0^j}(z^j)\big]}.
\label{eq:denoiser_expanded_app}
\end{align}

According to this expression, the probability denoiser $p_{1 \vert t}(x^i \vert z)$ can be interpreted as a weighted sum over all data points $x_1$, given by the term
\begin{align}
    \sum_{x_0} \prod_j \left[ \kappa_t \delta_{x_1^j}(z^j) + (1-\kappa_t) \delta_{x_0^j}(z^j) \right]. 
\end{align}

Let $h(s, z)$ denote the Hamming distance between two sequences $s$ and $z$, defined as
\begin{align}
    h(s, z) = N - \sum_j \delta_{s^j}(z^j),
\end{align}

and let $h_{+}(s, z)$ represent the similarity between the sequences, defined as
\begin{align}
    h_{+}(s, z) = \sum_j \delta_{s^j}(z^j) = N - h(s, z).
\end{align}

The weight is computed only when $d_m^i$ coincides with the target token $x^i$ (\ie{} $\delta_{d_m^i}(x^i)=1$). In this case, the term can be expressed as:
\begin{align}
    \label{eqn:closed_form_forward_pf}
    &\sum_{x_0} \prod_j \left[\kappa_t\delta_{d_m^j}(z^j) + (1-\kappa_t)\delta_{x_0^j}(z^j)\right] \\
    &= \sum_{k=0}^{h_{+}(d_m, z)} \binom{h_{+}(d_m, z)}{k} \left(1-\kappa_t\right)^{N - h_{+}(d_m, z)} \left((K-1)\kappa_t\right)^{h_{+}(d_m, z) - k}.
\end{align}

To understand this transition, we first note that $d_m$ and $z$ are fixed in this scope, while $x_0$ is independent across each dimension and follows a uniform distribution. This implies that we only need to consider $x_0$. For an arbitrary dimension $j$, the cases can be divided into four possibilities, and the corresponding values of $\left[\kappa_t \delta_{d_m^j}(z^j) + (1-\kappa_t)\delta_{x_0^j}(z^j)\right]$ are as follows:
\begin{description}
    \item[Case 1.] $d_m^j = z^j$, $x_0^j = z^j$, $\left[\kappa_t\delta_{d_m^j}(z^j) + (1-\kappa_t)\delta_{x_0^j}(z^j)\right] = 1.$
    \item[Case 2.] $d_m^j = z^j$, $x_0^j \neq z^j$, $\left[\kappa_t\delta_{d_m^j}(z^j) + (1-\kappa_t)\delta_{x_0^j}(z^j)\right] = \kappa_t.$
    \item[Case 3.] $d_m^j \neq z^j$, $x_0^j = z^j$, $\left[\kappa_t\delta_{d_m^j}(z^j) + (1-\kappa_t)\delta_{x_0^j}(z^j)\right] = 1-\kappa_t.$
    \item[Case 4.] $d_m^j \neq z^j$, $x_0^j \neq z^j$, $\left[\kappa_t\delta_{d_m^j}(z^j) + (1-\kappa_t)\delta_{x_0^j}(z^j)\right] = 0.$
\end{description}

Note that \textbf{Case 4} makes the term inside the product $\left[\kappa_t \delta_{d_m^j}(z^j) + (1-\kappa_t)\delta_{x_0^j}(z^j)\right]$ equal to zero. Thus, we only need to consider $x_0$ for which no dimension falls into \textbf{Case 4}. Among the $|K|^N$ possible choices of $x_0$, only $|K|^{h_{+}(d_m, z)}$ satisfy $x_0^j = z^j$ for all dimensions where $d_m^j \neq z^j$. We then classify the remaining cases according to the Hamming distance between $x_0$ and $d_m$. Note that the maximum value of $h_{+}(x_0, d_m)$ is $h_{+}(d_m, z)$. Let $k$ denote an integer in the range $0$ to $h_{+}(d_m, z)$. Then, the number of $x_0$ satisfying $h_{+}(x_0, d_m) = k$ is $\binom{h_{+}(d_m, z)}{k} (K-1)^{h_{+}(d_m, z)-k}$, and in this case the product term becomes
\begin{align}
    \label{eqn:forward_product_conversion}
    \prod_j \left[\kappa_t\delta_{d_m^j}(z^j) + (1-\kappa_t)\delta_{x_0^j}(z^j)\right] = (\kappa_t)^{h_{+}(d_m, z)-k}(1-\kappa_t)^{N-h_{+}(d_m, z)}.
\end{align}

We can then arrive at the equation above by summing over all possible $k$. Resuming the proof, the term can be further simplified as follows:
\begin{align}
    & \sum_{x_0} \prod_j \left[\kappa_t\delta_{d_m^j}(z^j) + (1-\kappa_t)\delta_{x_0^j}(z^j)\right] \label{eqn:closed_form_hard_part_start} \\
    &= \sum_{k=0}^{h_{+}(d_m, z)} \binom{h_{+}(d_m, z)}{k} \left(1-\kappa_t\right)^{N - h_{+}(d_m, z)} \left((K-1)\kappa_t\right)^{h_{+}(d_m, z) - k} \\
    &= \left(1-\kappa_t\right)^{N - h_{+}(d_m, z)} \sum_{k=0}^{h_{+}(d_m, z)} \binom{h_{+}(d_m, z)}{k} \left((K-1)\kappa_t\right)^{k} \\
    &= \left(1-\kappa_t\right)^{N} \left(\frac{(K-1)\kappa_t + 1}{1-\kappa_t}\right)^{h_{+}(d_m, z)} \\
    &= \left(1-\kappa_t\right)^{N} \left(1 + \frac{\kappa_t}{1-\kappa_t}K\right)^{h_{+}(d_m, z)}. \label{eqn:closed_form_hard_part_end}
\end{align}

We define $\gamma := 1 + \frac{\kappa_t}{1-\kappa_t}K$, and by substituting this simplified expression into the noise predictor above, we finally obtain~\Eqref{eqn:closed_form_noise_predictor}.
\begin{align}
    p_{1|t}(x^i|z) &= \frac{\sum_{m=1}^M \delta_{d_m^i}(x^i) \sum_{x_0} \prod_j \left[\kappa_t\delta_{d_m^j}(z^j) + (1-\kappa_t)\delta_{x_0^j}(z^j)\right]}{\sum_{m=1}^M \sum_{x_0} \prod_j \left[\kappa_t\delta_{d_m^j}(z^j) + (1-\kappa_t)\delta_{x_0^j}(z^j)\right]} \\
    &= \frac{\sum_{m=1}^M \delta_{d_m^i}(x^i) \left(1-\kappa_t\right)^{N} \left(1 + \frac{\kappa_t}{1-\kappa_t}K\right)^{h_{+}(d_m, z)}}{\sum_{m=1}^M \left(1-\kappa_t\right)^{N} \left(1 + \frac{\kappa_t}{1-\kappa_t}K\right)^{h_{+}(d_m, z)}} \\
    &= \frac{\sum_{m=1}^M \delta_{d_m^i}(x^i) \left(1 + \frac{\kappa_t}{1-\kappa_t}K\right)^{h_{+}(d_m, z)}}{\sum_{m=1}^M \left(1 + \frac{\kappa_t}{1-\kappa_t}K\right)^{h_{+}(d_m, z)}} \\
    &= \frac{\sum_{m=1}^M \delta_{d_m^i}(x^i) \gamma^{-h(d_m, z)}}{\sum_{m=1}^M \gamma^{-h(d_m, z)}}.
\end{align}

\subsection{Proof of Closed-form Backward Velocity in Discrete Flow Models}
\label{subsec:appx_proof_backward_velocity_discrete}

Similarly to the proof of the closed-form forward velocity in~\Secref{subsec:appx_proof_forward_velocity_discrete}, we start from the backward velocity in~\Eqref{eqn:backward_velocity}:
\begin{align}
    \check{v_t}(x^i, z) = \frac{\dot{\kappa}_t}{\kappa_t}\left[\delta_{z^i}(x^i) - p_{0|t}(x^i|z)\right].
\end{align}

We derive the closed-form noise predictor as follows:
\begin{align}
    p_{0|t}(x^i|z) &= \sum_{x_0, x_1} \delta_{x_0^i}(x^i) p_t(x_0, x_1 | z) \\
    &= \sum_{x_0, x_1} \delta_{x_0^i}(x^i) \frac{p_t(x_0, x_1, z)}{p_t(z)} \\
    &= \frac{\sum_{x_0, x_1} \delta_{x_0^i}(x^i) p_t(x_0, x_1, z)}{p_t(z)} \\
    &= \frac{\sum_{x_0, x_1} \delta_{x_0^i}(x^i) p_t(x_0, x_1, z)}{\sum_{x_0, x_1}p_t(x_0, x_1, z)},
\end{align}
The last expression is further expanded to:
\begin{align}
    p_{0|t}(x^i|z) &= \frac{\sum_{x_0, x_1} \delta_{x_0^i}(x^i) p_t(x_0, x_1, z)}{\sum_{x_0, x_1}p_t(x_0, x_1, z)} \\
    &= \frac{\sum_{x_0,x_1}\delta_{x_0^i}(x^i)\,p_1(x_1)\,p_t(z| x_0,x_1)}
    {\sum_{x_0,x_1}p_1(x_1)\,p_t(z |  x_0,x_1)} \\
    &= \frac{\sum_{m=1}^M \sum_{x_0}\delta_{x_0^i}(x^i)\,\prod_{j=1}^N\!\big[\kappa_t\,\delta_{d_m^j}(z^j)+(1-\kappa_t)\,\delta_{x_0^j}(z^j)\big]}
    {\sum_{m=1}^M \sum_{x_0}\prod_{j=1}^N\!\big[\kappa_t\,\delta_{d_m^j}(z^j)+(1-\kappa_t)\,\delta_{x_0^j}(z^j)\big]}.
\end{align}

For the denominator, we use the same formula as in~\Eqref{eqn:closed_form_hard_part_end}:
\begin{align}
    \sum_{m=1}^M \sum_{x_0} \prod_{j=1}^N \left[\kappa_t\delta_{d_m^j}(z^j) + (1-\kappa_t)\delta_{x_0^j}(z^j)\right] = \sum_{m=1}^M \left(1-\kappa_t\right)^{N} \left(1 + \frac{\kappa_t}{1-\kappa_t}K\right)^{h_{+}(d_m, z)}.
\end{align}

Next, for the numerator, we can rewrite it as:
\begin{align}
&\sum_{m=1}^M \sum_{x_0}\delta_{x_0^i}(x^i)\,\prod_{j=1}^N\!\big[\kappa_t\,\delta_{d_m^j}(z^j)+(1-\kappa_t)\,\delta_{x_0^j}(z^j)\big] \\
&=\sum_{m=1}^M \sum_{\substack{x_0 \text{ with} \\ x_0^i = x^i}} \prod_j \left[\kappa_t \delta_{d_m^j}(z^j) + (1-\kappa_t)\delta_{x_0^j}(z^j)\right].
\end{align}
The $i$-th index should be considered separately as $x_0^i$ is set to be equal to $x^i$. Separating the $j=i$ term from the product yields
\begin{align}
&\sum_{m=1}^M \sum_{\substack{x_0 \text{ with} \\ x_0^i = x^i}} \left[\kappa_t \delta_{d_m^i}(z^i) + (1-\kappa_t)\delta_{x_0^i}(z^i)\right] \prod_{j \neq i} \left[\kappa_t \delta_{d_m^j}(z^j) + (1-\kappa_t)\delta_{x_0^j}(z^j)\right] \\
& =\sum_{m=1}^M \sum_{\substack{x_0 \text{ with} \\ x_0^i = x^i}} \left[\kappa_t \delta_{d_m^i}(z^i) + (1-\kappa_t)\delta_{x^i}(z^i)\right] \prod_{j \neq i} \left[\kappa_t \delta_{d_m^j}(z^j) + (1-\kappa_t)\delta_{x_0^j}(z^j)\right] \\
& =\sum_{m=1}^M \left[\kappa_t \delta_{d_m^i}(z^i) + (1-\kappa_t)\delta_{x^i}(z^i)\right] \sum_{\substack{x_0 \text{ with} \\ x_0^i = x^i}} \prod_{j \neq i} \left[\kappa_t \delta_{d_m^j}(z^j) + (1-\kappa_t)\delta_{x_0^j}(z^j)\right].
\end{align}
Since the $i$-th coordinate of $x_0$ is fixed to $x^{i}$, the summation over $x_0$ with $x_0^i = x^i$ no longer depends on this index. Consequently, when we consider the summation only over the remaining coordinates $j \neq i$, the resulting expression takes exactly the same form as the computation presented in ~\Secref{subsec:appx_proof_forward_velocity_discrete} (\Eqref{eqn:closed_form_hard_part_start}-\Eqref{eqn:closed_form_hard_part_end}). The only differences are that (i) the effective dimensionality of the product is reduced from $N$ to $N-1$, and (ii) the matching count term must exclude the $i$-th coordinate, yielding $h_{+}(d_m, z)$ to $h_{+}(d_m, z) - \delta_{d_m^{i}}(z^{i})$.
Reflecting these adjustments, we obtain
\begin{align}
    & \sum_{\substack{x_0 \text{ with} \\ x_0^i = x^i}} \prod_{j \neq i} \left[\kappa_t \delta_{d_m^j}(z^j) + (1-\kappa_t)\delta_{x_0^j}(z^j)\right] = \left(1-\kappa_t\right)^{N-1} \left(1 + \frac{\kappa_t}{1-\kappa_t}K\right)^{h_{+}(d_m, z) - \delta_{d_m^i}(z^i)},
\end{align}
and
\begin{align}
    &\sum_{m=1}^M \left[\kappa_t \delta_{d_m^i}(z^i) + (1-\kappa_t)\delta_{x^i}(z^i)\right] \sum_{\substack{x_0 \text{ with} \\ x_0^i = x^i}} \prod_{j \neq i} \left[\kappa_t \delta_{d_m^j}(z^j) + (1-\kappa_t)\delta_{x_0^j}(z^j)\right] \\
    &= \sum_{m=1}^M \left[\kappa_t \delta_{d_m^i}(z^i) + (1-\kappa_t)\delta_{x^i}(z^i)\right] \left(1-\kappa_t\right)^{N-1} \left(1 + \frac{\kappa_t}{1-\kappa_t}K\right)^{h_{+}(d_m, z) - \delta_{d_m^i}(z^i)} \\
    &= \sum_{m=1}^M \left[\frac{\kappa_t}{1 - \kappa_t} \delta_{d_m^i}(z^i) + \delta_{x^i}(z^i)\right] \left(1-\kappa_t\right)^{N} \left(1 + \frac{\kappa_t}{1-\kappa_t}K\right)^{h_{+}(d_m, z) - \delta_{d_m^i}(z^i)}.
\end{align}
Using this nominator, we can denote the closed-form noise predictor:
\begin{align}
    p_{0|t}(x^i|z) &= \frac{\sum_{m=1}^M \left[\frac{\kappa_t}{1 - \kappa_t} \delta_{d_m^i}(z^i) + \delta_{x^i}(z^i)\right] \left(1-\kappa_t\right)^{N} \left(1 + \frac{\kappa_t}{1-\kappa_t}K\right)^{h_{+}(d_m, z) - \delta_{d_m^i}(z^i)}}{\sum_{m=1}^M \left(1-\kappa_t\right)^{N} \left(1 + \frac{\kappa_t}{1-\kappa_t}K\right)^{h_{+}(d_m, z)}} \\
    &= \sum_{m=1}^M \left[\frac{\kappa_t}{1 - \kappa_t} \delta_{d_m^i}(z^i) + \delta_{x^i}(z^i)\right] \frac{\left(1 + \frac{\kappa_t}{1-\kappa_t}K\right)^{h_{+}(d_m, z) - \delta_{d_m^i}(z^i)}}{\sum_{m'=1}^M \left(1 + \frac{\kappa_t}{1-\kappa_t}K\right)^{h_{+}(d_{m'}, z)}} \\
    &= \sum_{m=1}^M \left[\frac{\kappa_t}{1 - \kappa_t} \delta_{d_m^i}(z^i) + \delta_{x^i}(z^i)\right] \left[1 - \frac{K \kappa_t \ \delta_{d_m^i}(z^i)}{1 + (K-1)\kappa_t}\right] \frac{\gamma^{h_{+}(d_m, z)}}{\sum_{m'=1}^M \gamma^{h_{+}(d_{m'}, z)}} \\
    &= \delta_{x^i}(z^i) - \frac{\kappa_t \left(K \delta_{x^i}(z^i) - 1\right)}{1 + (K-1)\kappa_t} \sum_{m=1}^M \delta_{d_m^i}(z^i) \frac{\gamma^{h_{+}(d_m, z)}}{\sum_{m'=1}^M \gamma^{h_{+}(d_{m'}, z)}},
\end{align}
where $\gamma := 1 + \frac{\kappa_t}{1-\kappa_t}K$.

We can then express the closed-form backward velocity as follows:
\begin{align}
    \check{v_t}(x^i, z) &= \frac{\dot{\kappa}_t}{\kappa_t}\left[\delta_{z^i}(x^i) - p_{0|t}(x^i|z)\right] \\
    &= \frac{\dot{\kappa}_t \left(K \delta_{x^i}(z^i) - 1\right)}{1 + (K-1) \kappa_t} \sum_{m=1}^M \delta_{d_m^i}(z^i) \frac{\gamma^{h_{+}(d_m, z)}}{\sum_{m'=1}^M \gamma^{h_{+}(d_{m'}, z)}} \\
    &= \frac{\dot{\kappa}_t(K \delta_{x^i}(z^i)-1)}{1 + (K-1) \kappa_t}\sum_{m=1}^M \delta_{d_m^i}(z^i) \frac{\gamma^{-h(d_m, z)}}{\sum_{m'=1}^M \gamma^{-h(d_{m'}, z)}}.
\end{align}

Since $\kappa_t = 1$ for $t=1$, then $\gamma \rightarrow \infty$. So this equation is formally is not defined at $t=1$. Nevertheless, as $\lim_{t \rightarrow 1}$, the weighted sum over power of $\gamma$ is dominated by the maximum term, which converges to $1$. Hence, the expression can be rigorously interpreted as $\lim_{t \rightarrow 1}\check{v_t}(x^i, z)$, and in practice, this limiting value is used for sampling at $t=1$.

\section{Experiment details}
\label{sec:appendix_results_detail}

In~\Tabref{tab:appendix_experiment_details}, we summarize the hyperparameters used in the experiments presented in~\Secref{sec:results}, covering both training and finetuning configurations for each dataset. All reported samples were generated using the greedy-tail denoiser described in~\citep{Sahoo:2025Duo}. We employed an implementation of the closed-form backward velocity that is optimized at the CUDA level.

For the CIFAR-10 dataset, we follow the same setting as baseline~\citep{Schiff:2025UDLM}.~\Tabref{tab:appendix_cifar_fid_check} reports the FID and IS of baseline and \methodname{} measured with 1,000 steps and 50K samples, which are consistent with the results reported in Tab. 6 of~\citep{Schiff:2025UDLM} and \methodname{} outperforms it.

\begin{table}[ht!]
{\small
\centering
\caption{Summary of the training settings used in~\Secref{sec:results}. Specifically, ``Sampling Steps'' under~\ourname{} and ReDi~\citep{Yoo:2025ReDi} indicate the number of steps taken to generate pairs, ``Teacher Update Period'' under DCD~\citep{Sahoo:2025Duo} denotes the number of fine-tuning iterations between updates, when the teacher model is replaced by the current student model. ``\# Pairs'' under ReDi~\citep{Yoo:2025ReDi} denotes the number of pairs for the fine-tuning.
}
\label{tab:appendix_experiment_details}
\begin{tabularx}{\textwidth}{l|YYYY}
\toprule
 & MNIST-Binary & QM9 & CIFAR10 & ZINC-250k \\
\midrule
Training Iterations     & 10K  & 50K  & 300K  & 200K \\
Data Dimension    & $28 \times 28$ & 32 & $32\times 32 \times 3$ & 72 \\
Batch Size      & 256 & 1024 & 512 & 256 \\
Network Architecture    & U-Net & Transformer & U-Net & Transformer \\
Parameter Count & 25.8M & 92.4M & 35.8M & 92.4M \\
\toprule
 & \multicolumn{4}{c}{\methodname{}} \\
\midrule
Sampling Steps  & 20 & 20 & 20 & 64 \\
\toprule
 & \multicolumn{4}{c}{DCD~\citep{Sahoo:2025Duo}} \\
\midrule
Training Iterations  & 5K & 10K & 50K & 30K \\
Teacher Update Period  & 1K & 2K  & 10K & 5K  \\
\toprule
 & \multicolumn{4}{c}{ReDi~\citep{Yoo:2025ReDi}} \\
\midrule
Training Iterations  & 5K & 10K & 50K & 30K \\
\# Pairs  & 10K & 20K & 10K & 20K \\
Sampling Steps & 256 & 256 & 1024 & 256 \\
\bottomrule
\end{tabularx}
}
\end{table}

\begin{table}[t!]
\small
\centering
\begin{minipage}[t]{0.48\textwidth}
\centering
\caption{FID~\citep{Heusel:2017FID} and IS~\citep{Salimans:2016IS} of UDLM~\citep{Schiff:2025UDLM} and \methodname{} on the CIFAR-10 dataset~\citep{krizhevsky2009learning}.}
\label{tab:appendix_cifar_fid_check}
\begin{tabular}{l|cc}
\toprule
            & FID     & IS     \\
\midrule
UDLM        & 33.65 & 6.96 \\
\methodname{} & \textbf{28.07} & \textbf{7.37} \\
\bottomrule
\end{tabular}
\end{minipage}
\hfill
\begin{minipage}[t]{0.48\textwidth}
\centering
\caption{Total Correlation measure with pairs sampled from UDLM~\citep{Schiff:2025UDLM} and \methodname{} trained on QM9~\citep{ramakrishnan2014quantum}.}
\label{tab:appendix_tc_measure}
\begin{tabular}{l|c}
\toprule
            & Total Correlation \\
\midrule
UDLM & 31.87 \\
\methodname{} & \textbf{30.72} \\
\bottomrule
\end{tabular}
\end{minipage}
\end{table}

\section{Additional Experiments}
\label{sec:appendix_additional_experiment}

\subsection{Coverage of training dataset by sampling with forward velocity}
\label{subsec:coverage_forward_velocity}

As discussed in~\Secref{subsec:closed_form_forward_velocity}, constructing pairs using the closed-form forward velocity with a training dataset of size $\lvert X_1 \rvert$ incurs significantly higher cost to achieve full coverage compared to using the backward velocity. Let $k$ denote the number of source samples drawn from the source (prior) distribution, assumed to be uniform in our work. The probability that a given element in the training set is selected is $\left(1 - \tfrac{1}{\lvert X_1 \rvert}\right)^k$. Accordingly, we denote by $\bar{k}$ the number of unique samples among the $k$ draws, whose expectation is: $\sum_{i=1}^{\lvert X_1 \rvert} \Big[ 1 - (1 - \frac{1}{\lvert X_1 \rvert})^k \Big] = \lvert X_1 \rvert \Big[  1 - (1 - \frac{1}{\lvert X_1 \rvert})^k \Big]$.
In addition, we define the coverage as the ratio between the number of unique elements obtained through this sampling procedure and the training set size: $\text{COV} = \bar{k}/\vert X_1 \vert$.

\begin{table}[!t]
\small
\centering
\caption{Summary of training set sizes $\vert X_1 \vert$ for each dataset, the number of unique samples $\bar{k}$ obtained by simulating paths using the closed-form forward velocity in~\Eqref{eqn:closed_forward_velocity}, and the corresponding coverage values: empirical ($\text{COV}$) and theoretically predicted ($\text{COV}_{\text{Pred}}$).
}
\label{tab:appendix_forward_velocity_coverage}
\begin{tabular}{l|cccc}
\toprule
    & QM9 & ZINC-250k & MNIST-Binary & CIFAR-10      \\
\midrule
$|X_1| $ & 127,190 & 224,568 & 60,000 & 100,000 \\
$\bar{k}$ & 77,104 &	140,779 &	37,711 &	63,117 \\
$\text{COV}$ & 60.62\% & 62.68\% & 62.85\% & 63.11\% \\
\midrule
$\text{COV}_{\text{Pred}}$ & \multicolumn{4}{c}{63.21\%} \\
\bottomrule
\end{tabular}
\end{table}

To validate our claim in~\Secref{subsec:closed_form_forward_velocity}, we sample $k = \lvert X_1 \rvert$ data points $x_1$ by transporting source samples $x_0$, independently drawn from the uniform distribution, along the velocity field defined in~\Eqref{eqn:closed_forward_velocity}. Using these samples, we evaluate the coverage following the definition above. The dataset sizes, number of unique samples among the generated samples, and the empirical and theoretical coverages are summarized in~\Tabref{tab:appendix_forward_velocity_coverage}.
These findings indicate that, even when sampling the same number of points as the training set size, only about $63\%$ of the training distribution can be recovered in practice. Achieving full coverage would therefore require a substantially larger number of samples, introducing significantly higher computational cost. Motivated by this finding, we instead propose tracing backward from data samples, using a closed-form velocity field that we derive for this purpose (\Secref{subsec:closed_form_backward_velocity}).


\subsection{Total Correlation analysis of closed-form velocity}
\label{subsec:total_correlation_measure}

As in~\Secref{subsec:rectified_flow}, \citet{Yoo:2025ReDi} demonstrated that iteratively refining the joint distribution of source-target pairs in discrete flow models improves few-step performance by reducing the total correlation of the model. In this section, we measure the total correlation following their methodology. Specifically, we perform sampling with neural networks, including UDLM~\citep{Schiff:2025UDLM} and \methodname{} trained on QM9~\citep{ramakrishnan2014quantum}, starting from identical initial states $x_0$ but with varying random seeds. We randomly select $20{,}000$ initial states $x_0$, and for each $x_0$, we generate 10 samples with a step size of 256. As shown in~\Tabref{tab:appendix_tc_measure}, \methodname{} achieves a lower total correlation, consistent with the improved performance observed in few-step sampling, as discussed above.


\section{Detailed Experimental Results}
\label{sec:appendix_results_full}

In this section, we provide the detailed experimental results corresponding to those summarized in~\Secref{sec:results}. For the molecular datasets (QM9~\citep{ramakrishnan2014quantum} and ZINC-250k~\citep{irwin2012zinc}), we generate 1,024 samples across varying timesteps and evaluate validity, uniqueness, and novelty. Reported values are averaged over 10 trials, with standard deviations also included. For the image domain, we report FID on MNIST-Binary~\citep{lecun2002gradient}, and both FID~\citep{Heusel:2017FID} and IS~\citep{Salimans:2016IS} on CIFAR-10~\citep{krizhevsky2009learning}. 
Detailed experimental settings are provided in~\Secref{subsec:experiment_setting}.

Results on QM9 are presented in Tables~\ref{tab:qm9_results_valid},~\ref{tab:qm9_results_unique}, and~\ref{tab:qm9_results_novel}, reporting validity, uniqueness, and novelty, respectively. Corresponding results on ZINC-250k are shown in Tables~\ref{tab:zinc_results_valid},~\ref{tab:zinc_results_unique}, and~\ref{tab:zinc_results_novel}. Finally, results for the image datasets are summarized in~\Tabref{tab:binary_mnist_results} (FID on MNIST-Binary) and~\Tabref{tab:cifar10_results} (FID and IS on CIFAR-10).
The FID measured on MNIST-Binary~\citep{lecun2002gradient}, FID and IS measured on CIFAR-10~\citep{krizhevsky2009learning}, are summarized in~\Tabref{tab:binary_mnist_results} and~\Tabref{tab:cifar10_results}, respectively.

\clearpage

\begin{table}[H]
\small
\centering
\caption{Validity scores ($\uparrow$) on QM9~\citep{ramakrishnan2014quantum} for various methods across different steps. Best values per column are highlighted in bold.}
\label{tab:qm9_results_valid}
\setlength{\tabcolsep}{3.2pt}
\begin{tabularx}{\textwidth}{l|YYYYYYY}
\toprule
Method & 1 & 2 & 4 & 8 & 16 & 32 & 64 \\
\midrule
MDLM & 51.4{\scriptsize$\pm$5.7} & 80.0{\scriptsize$\pm$11.1} & 154.8{\scriptsize$\pm$10.0} & 347.5{\scriptsize$\pm$11.5} & 530.0{\scriptsize$\pm$11.6} & 662.9{\scriptsize$\pm$16.5} & 736.4{\scriptsize$\pm$16.4} \\
UDLM & 17.5{\scriptsize$\pm$3.2} & 125.5{\scriptsize$\pm$11.8} & 497.6{\scriptsize$\pm$8.3} & 826.6{\scriptsize$\pm$10.3} & 953.5{\scriptsize$\pm$6.1} & \textbf{991.9}{\scriptsize$\pm$4.2} & 1000.1{\scriptsize$\pm$3.5} \\
Random & 47.1{\scriptsize$\pm$5.7} & 194.6{\scriptsize$\pm$8.2} & 554.2{\scriptsize$\pm$13.3} & 858.3{\scriptsize$\pm$17.5} & 962.0{\scriptsize$\pm$6.7} & 989.9{\scriptsize$\pm$7.6} & 998.6{\scriptsize$\pm$5.3} \\
\methodname{} & \textbf{223.4}{\scriptsize$\pm$12.7} & \textbf{416.0}{\scriptsize$\pm$12.4} & \textbf{734.9}{\scriptsize$\pm$7.2} & \textbf{921.5}{\scriptsize$\pm$11.0} & \textbf{977.1}{\scriptsize$\pm$3.9} & 990.9{\scriptsize$\pm$5.9} & \textbf{1000.2}{\scriptsize$\pm$4.5} \\
\midrule
UDLM + DCD & 323.0{\scriptsize$\pm$19.5} & 530.8{\scriptsize$\pm$20.0} & 816.6{\scriptsize$\pm$14.4} & 941.1{\scriptsize$\pm$8.5} & 981.0{\scriptsize$\pm$4.8} & \textbf{993.0}{\scriptsize$\pm$3.6} & \textbf{999.4}{\scriptsize$\pm$4.7} \\
\methodname{} + DCD & \textbf{453.8}{\scriptsize$\pm$16.4} & \textbf{685.8}{\scriptsize$\pm$16.7} & \textbf{891.6}{\scriptsize$\pm$11.9} & \textbf{963.1}{\scriptsize$\pm$7.8} & \textbf{983.7}{\scriptsize$\pm$8.5} & 989.3{\scriptsize$\pm$3.5} & 993.2{\scriptsize$\pm$5.7} \\
\midrule
UDLM + Redi & 59.7{\scriptsize$\pm$8.8} & 232.4{\scriptsize$\pm$9.2} & 588.4{\scriptsize$\pm$15.8} & 849.6{\scriptsize$\pm$14.2} & 940.5{\scriptsize$\pm$8.5} & 967.5{\scriptsize$\pm$5.2} & 978.8{\scriptsize$\pm$5.2} \\
\methodname{} + Redi & \textbf{361.0}{\scriptsize$\pm$115.2} & \textbf{512.6}{\scriptsize$\pm$44.2} & \textbf{775.7}{\scriptsize$\pm$10.0} & \textbf{929.1}{\scriptsize$\pm$11.6} & \textbf{976.2}{\scriptsize$\pm$4.5} & \textbf{985.6}{\scriptsize$\pm$6.7} & \textbf{993.1}{\scriptsize$\pm$7.1} \\
\bottomrule
\end{tabularx}
\end{table}

\begin{table}[H]
\small
\centering
\caption{Uniqueness scores ($\uparrow$) on QM9~\citep{ramakrishnan2014quantum} for various methods across different steps. Best values per column are highlighted in bold.}
\setlength{\tabcolsep}{3.2pt}
\label{tab:qm9_results_unique}
\begin{tabularx}{\textwidth}{l|YYYYYYY}
\toprule
Method & 1 & 2 & 4 & 8 & 16 & 32 & 64 \\
\midrule
MDLM & 18.9{\scriptsize$\pm$3.0} & 28.8{\scriptsize$\pm$3.3} & 87.4{\scriptsize$\pm$8.4} & 254.7{\scriptsize$\pm$10.3} & 443.8{\scriptsize$\pm$16.4} & 591.0{\scriptsize$\pm$18.6} & 666.9{\scriptsize$\pm$17.5} \\
UDLM & 17.5{\scriptsize$\pm$3.2} & 125.4{\scriptsize$\pm$11.7} & 495.0{\scriptsize$\pm$8.2} & 819.5{\scriptsize$\pm$11.4} & 943.0{\scriptsize$\pm$5.7} & 979.1{\scriptsize$\pm$5.0} & 990.0{\scriptsize$\pm$4.7} \\
Random & 47.1{\scriptsize$\pm$5.7} & 194.5{\scriptsize$\pm$8.1} & 551.2{\scriptsize$\pm$12.5} & 853.1{\scriptsize$\pm$16.9} & 953.5{\scriptsize$\pm$7.8} & 981.1{\scriptsize$\pm$6.7} & 989.6{\scriptsize$\pm$5.7} \\
\methodname{} & \textbf{223.0}{\scriptsize$\pm$12.3} & \textbf{414.7}{\scriptsize$\pm$12.0} & \textbf{731.4}{\scriptsize$\pm$6.9} & \textbf{917.4}{\scriptsize$\pm$11.8} & \textbf{971.6}{\scriptsize$\pm$4.3} & \textbf{986.2}{\scriptsize$\pm$5.3} & \textbf{994.8}{\scriptsize$\pm$5.2} \\
\midrule
UDLM + DCD & 320.9{\scriptsize$\pm$18.7} & 528.0{\scriptsize$\pm$19.7} & 808.5{\scriptsize$\pm$12.7} & 932.3{\scriptsize$\pm$8.1} & 970.9{\scriptsize$\pm$5.8} & 978.3{\scriptsize$\pm$4.9} & 987.2{\scriptsize$\pm$4.3} \\
\methodname{} + DCD & \textbf{451.8}{\scriptsize$\pm$15.7} & \textbf{681.6}{\scriptsize$\pm$16.5} & \textbf{886.5}{\scriptsize$\pm$12.6} & \textbf{957.8}{\scriptsize$\pm$7.6} & \textbf{978.7}{\scriptsize$\pm$9.1} & \textbf{983.4}{\scriptsize$\pm$3.9} & \textbf{989.0}{\scriptsize$\pm$5.3} \\
\midrule
UDLM + Redi & 59.7{\scriptsize$\pm$8.8} & 231.6{\scriptsize$\pm$9.5} & 581.0{\scriptsize$\pm$15.1} & 834.7{\scriptsize$\pm$11.4} & 917.5{\scriptsize$\pm$9.6} & 944.8{\scriptsize$\pm$6.2} & 956.1{\scriptsize$\pm$5.2} \\
\methodname{} + Redi & \textbf{359.5}{\scriptsize$\pm$113.1} & \textbf{507.7}{\scriptsize$\pm$43.0} & \textbf{765.3}{\scriptsize$\pm$8.8} & \textbf{913.5}{\scriptsize$\pm$10.2} & \textbf{959.7}{\scriptsize$\pm$5.8} & \textbf{968.8}{\scriptsize$\pm$8.4} & \textbf{973.0}{\scriptsize$\pm$9.6} \\
\bottomrule
\end{tabularx}
\end{table}

\begin{table}[H]
\small
\centering
\caption{Novelty scores ($\uparrow$) on QM9~\citep{ramakrishnan2014quantum} for various methods across different steps. Best values per column are highlighted in bold.}
\setlength{\tabcolsep}{3.2pt}
\label{tab:qm9_results_novel}
\begin{tabularx}{\textwidth}{l|YYYYYYY}
\toprule
Method & 1 & 2 & 4 & 8 & 16 & 32 & 64 \\
\midrule
MDLM & 15.4{\scriptsize$\pm$3.1} & 23.6{\scriptsize$\pm$2.6} & 56.3{\scriptsize$\pm$5.2} & 127.4{\scriptsize$\pm$9.4} & \textbf{172.2}{\scriptsize$\pm$12.4} & \textbf{206.5}{\scriptsize$\pm$7.4} & \textbf{193.6}{\scriptsize$\pm$10.4} \\
UDLM & 13.8{\scriptsize$\pm$2.9} & 52.0{\scriptsize$\pm$8.2} & 120.0{\scriptsize$\pm$3.8} & 152.4{\scriptsize$\pm$9.1} & 144.2{\scriptsize$\pm$12.4} & 147.2{\scriptsize$\pm$9.7} & 145.1{\scriptsize$\pm$9.0} \\
Random & 26.9{\scriptsize$\pm$5.0} & 66.5{\scriptsize$\pm$7.1} & \textbf{128.8}{\scriptsize$\pm$13.0} & \textbf{156.8}{\scriptsize$\pm$10.6} & 145.5{\scriptsize$\pm$9.1} & 146.5{\scriptsize$\pm$7.0} & 150.1{\scriptsize$\pm$8.7} \\
\methodname{} & \textbf{68.8}{\scriptsize$\pm$7.8} & \textbf{85.6}{\scriptsize$\pm$10.0} & 109.2{\scriptsize$\pm$7.8} & 96.8{\scriptsize$\pm$9.9} & 106.5{\scriptsize$\pm$12.5} & 108.9{\scriptsize$\pm$9.4} & 110.0{\scriptsize$\pm$9.9} \\
\midrule
UDLM + DCD & \textbf{145.9}{\scriptsize$\pm$7.9} & \textbf{137.5}{\scriptsize$\pm$9.8} & \textbf{185.5}{\scriptsize$\pm$10.9} & \textbf{186.2}{\scriptsize$\pm$13.7} & \textbf{173.4}{\scriptsize$\pm$15.5} & \textbf{168.0}{\scriptsize$\pm$10.4} & \textbf{165.4}{\scriptsize$\pm$11.6} \\
\methodname{} + DCD & 110.3{\scriptsize$\pm$6.3} & 136.1{\scriptsize$\pm$12.0} & 146.2{\scriptsize$\pm$12.9} & 139.2{\scriptsize$\pm$9.6} & 131.8{\scriptsize$\pm$11.7} & 140.1{\scriptsize$\pm$9.6} & 133.2{\scriptsize$\pm$7.9} \\
\midrule
UDLM + Redi & 31.4{\scriptsize$\pm$8.0} & 73.3{\scriptsize$\pm$7.1} & \textbf{110.4}{\scriptsize$\pm$12.2} & \textbf{126.8}{\scriptsize$\pm$8.9} & \textbf{116.3}{\scriptsize$\pm$8.4} & \textbf{120.2}{\scriptsize$\pm$9.0} & \textbf{117.6}{\scriptsize$\pm$7.1} \\
\methodname{} + Redi & \textbf{84.2}{\scriptsize$\pm$11.3} & \textbf{92.0}{\scriptsize$\pm$6.9} & 101.1{\scriptsize$\pm$9.6} & 98.6{\scriptsize$\pm$8.4} & 98.8{\scriptsize$\pm$13.9} & 95.8{\scriptsize$\pm$8.3} & 98.9{\scriptsize$\pm$7.5} \\
\bottomrule
\end{tabularx}
\end{table}

\begin{table}[H]
\small
\centering
\caption{Validity scores ($\uparrow$) on ZINC-250k~\citep{irwin2012zinc} for various methods across different steps. Best values per column are highlighted in bold.}
\setlength{\tabcolsep}{3.2pt}
\label{tab:zinc_results_valid}
\begin{tabularx}{\textwidth}{l|YYYYYYY}
\toprule
Method & 1 & 2 & 4 & 8 & 16 & 32 & 64 \\
\midrule
MDLM & \textbf{15.0}{\scriptsize$\pm$4.0} & 79.4{\scriptsize$\pm$4.7} & 194.6{\scriptsize$\pm$15.1} & 351.3{\scriptsize$\pm$20.6} & 463.7{\scriptsize$\pm$17.7} & 553.9{\scriptsize$\pm$10.7} & 610.1{\scriptsize$\pm$17.6} \\
UDLM & 0.3{\scriptsize$\pm$0.5} & 65.2{\scriptsize$\pm$8.2} & 435.7{\scriptsize$\pm$14.4} & 775.1{\scriptsize$\pm$19.5} & \textbf{887.3}{\scriptsize$\pm$12.7} & \textbf{921.5}{\scriptsize$\pm$8.5} & \textbf{937.3}{\scriptsize$\pm$3.9} \\
Random & 0.6{\scriptsize$\pm$0.9} & 68.3{\scriptsize$\pm$10.7} & 351.2{\scriptsize$\pm$15.8} & 569.4{\scriptsize$\pm$16.6} & 611.0{\scriptsize$\pm$16.3} & 602.4{\scriptsize$\pm$13.3} & 571.0{\scriptsize$\pm$13.2} \\
\methodname{} & 9.9{\scriptsize$\pm$2.3} & \textbf{146.3}{\scriptsize$\pm$10.4} & \textbf{533.9}{\scriptsize$\pm$13.9} & \textbf{799.4}{\scriptsize$\pm$9.2} & 873.2{\scriptsize$\pm$14.1} & 901.0{\scriptsize$\pm$14.2} & 907.8{\scriptsize$\pm$7.7} \\
\midrule
UDLM + DCD & 25.7{\scriptsize$\pm$4.7} & 323.9{\scriptsize$\pm$12.5} & 718.2{\scriptsize$\pm$13.5} & \textbf{873.5}{\scriptsize$\pm$8.4} & \textbf{919.8}{\scriptsize$\pm$10.0} & \textbf{933.1}{\scriptsize$\pm$5.9} & \textbf{942.2}{\scriptsize$\pm$4.3} \\
\methodname{} + DCD & \textbf{114.9}{\scriptsize$\pm$14.3} & \textbf{436.3}{\scriptsize$\pm$16.5} & \textbf{725.1}{\scriptsize$\pm$11.5} & 858.2{\scriptsize$\pm$10.0} & 896.5{\scriptsize$\pm$8.2} & 900.9{\scriptsize$\pm$9.5} & 907.1{\scriptsize$\pm$13.7} \\
\midrule
UDLM + Redi & 0.7{\scriptsize$\pm$0.8} & 75.9{\scriptsize$\pm$7.9} & 424.8{\scriptsize$\pm$16.6} & 734.4{\scriptsize$\pm$8.6} & 856.3{\scriptsize$\pm$11.3} & 892.2{\scriptsize$\pm$10.4} & 900.1{\scriptsize$\pm$10.8} \\
\methodname{} + Redi & \textbf{46.3}{\scriptsize$\pm$6.3} & \textbf{221.5}{\scriptsize$\pm$11.0} & \textbf{562.8}{\scriptsize$\pm$12.7} & \textbf{793.6}{\scriptsize$\pm$8.4} & \textbf{869.3}{\scriptsize$\pm$14.2} & \textbf{897.4}{\scriptsize$\pm$9.1} & \textbf{907.0}{\scriptsize$\pm$10.5} \\
\bottomrule
\end{tabularx}
\end{table}

\begin{table}[H]
\small
\centering
\caption{Uniqueness scores ($\uparrow$) on ZINC-250k~\citep{irwin2012zinc} for various methods across different steps. Best values per column are highlighted in bold.}
\setlength{\tabcolsep}{3.2pt}
\label{tab:zinc_results_unique}
\begin{tabularx}{\textwidth}{l|YYYYYYY}
\toprule
Method & 1 & 2 & 4 & 8 & 16 & 32 & 64 \\
\midrule
MDLM & 7.4{\scriptsize$\pm$2.7} & 32.7{\scriptsize$\pm$2.4} & 105.1{\scriptsize$\pm$8.3} & 256.0{\scriptsize$\pm$18.3} & 392.6{\scriptsize$\pm$16.6} & 511.3{\scriptsize$\pm$14.0} & 582.4{\scriptsize$\pm$17.3} \\
UDLM & 0.3{\scriptsize$\pm$0.5} & 65.2{\scriptsize$\pm$8.2} & 435.7{\scriptsize$\pm$14.4} & 775.1{\scriptsize$\pm$19.5} & \textbf{887.3}{\scriptsize$\pm$12.7} & \textbf{921.5}{\scriptsize$\pm$8.5} & \textbf{937.2}{\scriptsize$\pm$3.8} \\
Random & 0.6{\scriptsize$\pm$0.9} & 68.3{\scriptsize$\pm$10.7} & 351.2{\scriptsize$\pm$15.8} & 569.4{\scriptsize$\pm$16.6} & 611.0{\scriptsize$\pm$16.3} & 602.4{\scriptsize$\pm$13.3} & 571.0{\scriptsize$\pm$13.2} \\
\methodname{} & \textbf{9.9}{\scriptsize$\pm$2.3} & \textbf{146.3}{\scriptsize$\pm$10.4} & \textbf{533.9}{\scriptsize$\pm$13.9} & \textbf{799.4}{\scriptsize$\pm$9.2} & 873.2{\scriptsize$\pm$14.1} & 901.0{\scriptsize$\pm$14.2} & 907.8{\scriptsize$\pm$7.7} \\
\midrule
UDLM + DCD & 25.7{\scriptsize$\pm$4.7} & 323.9{\scriptsize$\pm$12.5} & 718.2{\scriptsize$\pm$13.5} & \textbf{873.5}{\scriptsize$\pm$8.4} & \textbf{919.8}{\scriptsize$\pm$10.0} & \textbf{933.1}{\scriptsize$\pm$5.9} & \textbf{942.2}{\scriptsize$\pm$4.3} \\
\methodname{} + DCD & \textbf{114.9}{\scriptsize$\pm$14.3} & \textbf{436.3}{\scriptsize$\pm$16.5} & \textbf{725.1}{\scriptsize$\pm$11.5} & 858.2{\scriptsize$\pm$10.0} & 896.5{\scriptsize$\pm$8.2} & 900.9{\scriptsize$\pm$9.5} & 907.1{\scriptsize$\pm$13.7} \\
\midrule
UDLM + Redi & 0.7{\scriptsize$\pm$0.8} & 75.9{\scriptsize$\pm$7.9} & 424.8{\scriptsize$\pm$16.6} & 734.3{\scriptsize$\pm$8.7} & 856.3{\scriptsize$\pm$11.3} & 892.2{\scriptsize$\pm$10.4} & 900.0{\scriptsize$\pm$10.9} \\
\methodname{} + Redi & \textbf{46.3}{\scriptsize$\pm$6.3} & \textbf{221.5}{\scriptsize$\pm$11.0} & \textbf{562.8}{\scriptsize$\pm$12.7} & \textbf{793.6}{\scriptsize$\pm$8.4} & \textbf{869.3}{\scriptsize$\pm$14.2} & \textbf{897.4}{\scriptsize$\pm$9.1} & \textbf{907.0}{\scriptsize$\pm$10.5} \\
\bottomrule
\end{tabularx}
\end{table}

\begin{table}[H]
\small
\centering
\caption{Novelty scores ($\uparrow$) on ZINC-250k~\citep{irwin2012zinc} for various methods across different steps. Best values per column are highlighted in bold.}
\setlength{\tabcolsep}{3.2pt}
\label{tab:zinc_results_novel}
\begin{tabularx}{\textwidth}{l|YYYYYYY}
\toprule
Method & 1 & 2 & 4 & 8 & 16 & 32 & 64 \\
\midrule
MDLM & 3.8{\scriptsize$\pm$2.7} & 24.2{\scriptsize$\pm$2.6} & 84.5{\scriptsize$\pm$7.4} & 228.3{\scriptsize$\pm$17.1} & 372.0{\scriptsize$\pm$15.2} & 494.5{\scriptsize$\pm$15.2} & 569.3{\scriptsize$\pm$17.1} \\
UDLM & 0.3{\scriptsize$\pm$0.5} & 65.2{\scriptsize$\pm$8.2} & 435.7{\scriptsize$\pm$14.4} & 775.1{\scriptsize$\pm$19.5} & \textbf{887.3}{\scriptsize$\pm$12.7} & \textbf{921.3}{\scriptsize$\pm$8.8} & \textbf{936.9}{\scriptsize$\pm$4.1} \\
Random & 0.6{\scriptsize$\pm$0.9} & 68.3{\scriptsize$\pm$10.7} & 351.2{\scriptsize$\pm$15.8} & 569.4{\scriptsize$\pm$16.6} & 611.0{\scriptsize$\pm$16.3} & 602.4{\scriptsize$\pm$13.3} & 571.0{\scriptsize$\pm$13.2} \\
\methodname{} & \textbf{9.9}{\scriptsize$\pm$2.3} & \textbf{146.3}{\scriptsize$\pm$10.4} & \textbf{533.9}{\scriptsize$\pm$13.9} & \textbf{799.4}{\scriptsize$\pm$9.2} & 873.2{\scriptsize$\pm$14.1} & 901.0{\scriptsize$\pm$14.2} & 907.8{\scriptsize$\pm$7.7} \\
\midrule
UDLM + DCD & 25.7{\scriptsize$\pm$4.7} & 323.9{\scriptsize$\pm$12.5} & 718.2{\scriptsize$\pm$13.5} & \textbf{873.5}{\scriptsize$\pm$8.4} & \textbf{919.7}{\scriptsize$\pm$9.9} & \textbf{933.0}{\scriptsize$\pm$5.8} & \textbf{942.2}{\scriptsize$\pm$4.3} \\
\methodname{} + DCD & \textbf{114.9}{\scriptsize$\pm$14.3} & \textbf{436.3}{\scriptsize$\pm$16.5} & \textbf{725.1}{\scriptsize$\pm$11.5} & 858.2{\scriptsize$\pm$10.0} & 896.4{\scriptsize$\pm$8.3} & 900.9{\scriptsize$\pm$9.5} & 907.1{\scriptsize$\pm$13.7} \\
\midrule
UDLM + Redi & 0.7{\scriptsize$\pm$0.8} & 75.9{\scriptsize$\pm$7.9} & 424.8{\scriptsize$\pm$16.6} & 734.3{\scriptsize$\pm$8.7} & 856.3{\scriptsize$\pm$11.3} & 892.1{\scriptsize$\pm$10.5} & 900.0{\scriptsize$\pm$10.9} \\
\methodname{} + Redi & \textbf{46.3}{\scriptsize$\pm$6.3} & \textbf{221.5}{\scriptsize$\pm$11.0} & \textbf{562.8}{\scriptsize$\pm$12.7} & \textbf{793.5}{\scriptsize$\pm$8.3} & \textbf{869.3}{\scriptsize$\pm$14.2} & \textbf{897.4}{\scriptsize$\pm$9.1} & \textbf{907.0}{\scriptsize$\pm$10.5} \\
\bottomrule
\end{tabularx}
\end{table}
\begin{table}[H]
\small
\centering
\caption{FID ($\downarrow$) on MNIST-Binary~\citep{lecun2002gradient} for various methods across different steps. Best values per column are bolded.}
\label{tab:binary_mnist_results}
\begin{tabularx}{\textwidth}{l|YYYYYYY}
\toprule
Method & 1 & 2 & 4 & 8 & 16 & 32 & 64 \\
\midrule
MDLM & 204.64 & 159.26 & 103.74 & 54.41 & 28.51 & 12.31 & 7.01 \\
UDLM & 130.57 & 42.54 & 11.25 & 5.70 & \textbf{4.69} & \textbf{4.77} & \textbf{5.01} \\
Random & 128.57 & 36.59 & 9.41 & \textbf{5.60} & 5.00 & 5.10 & 5.19 \\
\methodname{} & \textbf{40.58} & \textbf{15.61} & \textbf{8.50} & 5.97 & 5.55 & 5.24 & 5.17 \\
\midrule
UDLM + DCD & 53.84 & 16.09 & \textbf{8.06} & \textbf{7.46} & \textbf{7.12} & \textbf{6.52}& \textbf{6.65} \\
\methodname{} + DCD & \textbf{19.51} & \textbf{14.20} & 11.47 & 9.28 & 7.75 & 7.82 & 8.42 \\
\midrule
UDLM + ReDi & 18.44 & 10.35 & \textbf{8.11} & \textbf{6.65}& \textbf{6.56} & \textbf{6.55} & \textbf{6.49} \\
\methodname{} + ReDi & \textbf{12.90} & \textbf{9.38} & 8.54 & 6.85 & 6.79 & 6.96 & 6.94 \\
\bottomrule
\end{tabularx}
\end{table}
\begin{table}[t!]
\small
\centering
\caption{FID ($\downarrow$) and IS ($\uparrow$) on CIFAR-10~\citep{krizhevsky2009learning} for various methods across different timesteps. Best values per column are bolded.}
\label{tab:cifar10_results}
\setlength{\tabcolsep}{3.3pt}
\begin{tabularx}{\textwidth}{l|YYYYYYYYYYY}
\toprule
 & 1 & 2 & 4 & 8 & 16 & 32 & 64 & 128 & 256 & 512 & 1024 \\
\midrule
Method & \multicolumn{11}{c}{FID ($\downarrow$)} \\
\midrule
MDLM & 407.31 & 359.97 & 340.43 & 340.98 & 321.79 & 228.34 & 131.63 & 77.44 & 52.89 & 42.88 & 39.04 \\
UDLM & 340.47 & 321.98 & 255.95 & 151.60 & 91.31 & 62.43 & 49.06 & 42.26 & 40.18 & 37.63 & 37.60 \\
Random & 328.83 & 314.64 & 220.16 & 127.66 & 82.76 & 58.42 & 45.18 & 39.50 & 36.78 & 35.16 & 35.07 \\
\methodname{} & \textbf{269.87} & \textbf{260.29} & \textbf{192.62} & \textbf{112.78} & \textbf{73.87} & \textbf{52.22} & \textbf{40.06} & \textbf{34.42} & \textbf{33.83} & \textbf{32.42} & \textbf{31.85} \\
\midrule
UDLM + DCD & 318.33 & 282.38 & 204.15 & 108.30 & \textbf{75.56} & \textbf{69.87} & \textbf{74.62} & \textbf{79.35} & \textbf{84.34} & \textbf{85.79} & \textbf{87.87} \\
\ourname{} + DCD & \textbf{223.86} & \textbf{190.15} & \textbf{139.46} & \textbf{95.91} & 82.11 & 80.70 & 86.59 & 100.16 & 114.25 & 123.41 & 129.45 \\
\midrule
UDLM + ReDi & \textbf{251.02} & \textbf{212.20} & \textbf{171.24} & 148.18 & 138.18 & 131.94 & 125.81 & 123.77 & 123.39 & 121.52 & 121.06 \\
\ourname{} + ReDi & 275.09 & 250.87 & 184.45 & \textbf{119.04} & \textbf{89.46} & \textbf{74.46} & \textbf{66.22} & \textbf{61.83} & \textbf{61.35} & \textbf{60.10} & \textbf{59.53} \\
\midrule
Method & \multicolumn{11}{c}{IS ($\uparrow$)} \\
\midrule
MDLM & 1.21 & 1.22 & 1.24 & 1.31 & 1.52 & 2.47 & 4.03 & 5.08 & 5.66 & 6.18 & 6.38 \\
UDLM & 1.32 & 1.48 & 2.11 & 3.54 & 4.85 & 5.90 & 6.25 & 6.56 & 6.66 & 6.87 & 6.81 \\
Random & 1.37 & 1.52 & 2.47 & 4.01 & 5.24 & 6.08 & 6.56 & 6.75 & 6.87 & 7.07 & 7.05 \\
\methodname{} & \textbf{1.72} & \textbf{1.80} & \textbf{2.70} & \textbf{4.27} & \textbf{5.61} & \textbf{6.23} & \textbf{6.86} & \textbf{7.00} & \textbf{7.12} & \textbf{7.14} & \textbf{7.33} \\
\midrule
UDLM + DCD & 1.49 & 1.60 & 2.23 & 3.90 & \textbf{5.09} & \textbf{5.45} & \textbf{5.42} & \textbf{5.16} & \textbf{5.09} & \textbf{5.07} & \textbf{4.84} \\
\ourname{} + DCD & \textbf{2.21} & \textbf{2.38} & \textbf{3.23} & \textbf{4.38} & 4.99 & 5.13 & 4.97 & 4.48 & 4.26 & 3.82 & 3.81 \\
\midrule
UDLM + ReDi & \textbf{1.85} & \textbf{2.35} & \textbf{2.96} & 3.37 & 3.52 & 3.67 & 3.86 & 3.96 & 3.93 & 3.99 & 3.94 \\
\ourname{} + ReDi & 1.80 & 1.95 & 2.91 & \textbf{4.18} & \textbf{5.02} & \textbf{5.65} & \textbf{5.88} & \textbf{6.03} & \textbf{6.06} & \textbf{6.09} & \textbf{6.30} \\
\bottomrule
\end{tabularx}
\end{table}
\section{Experiment on Continuous Flow Matching}
\label{sec:continuous_flow_matching}


\begin{table}[!t]
\small
\centering
\caption{FID of~\methodname{} on MNIST \citep{lecun2002gradient} with continuous values, measured using FID over 50K samples across various timesteps. Best values are bolded.}
\begin{tabularx}{\textwidth}{
l|YYYYYYY
}
\toprule
Method           & 1 & 2 & 4 & 8 & 16 & 32 & 64 \\
\midrule
CondOT                  & 398.43 & 91.17  & 27.34  & 10.80  & 5.81    & 3.99    & 3.16    \\
\methodname{}    & \textbf{74.89}  & \textbf{14.40}  & \textbf{6.89}   & \textbf{3.78}   &\textbf{2.70}    & \textbf{2.42}    & \textbf{2.37}    \\
\midrule
CondOT+RF               & 32.70  & 9.12   & 5.46   & 4.24   & 3.93    & 3.56    & 3.24    \\
\methodname{}+RF & \textbf{28.61}  & \textbf{4.15}   & \textbf{3.01}   & \textbf{2.87}   & \textbf{2.89}    & \textbf{2.91}    & \textbf{2.94}    \\
\bottomrule
\end{tabularx}
\label{tab:continuous_mnist}
\end{table}

Alongside our main experiments in the discrete setting, we also demonstrate the potential of our method to extend to continuous domains, as illustrated by the toy experiment presented below. Here, we denote by~\ourname{} a continuous flow model trained on source–target pairs constructed using the continuous variant of the algorithm described in~\Secref{subsec:closed_form_backward_velocity}.

\subsection{Closed-form Velocity in Continuous Flow Matching}
\label{subsec:closed_velocity_flow}

\paragraph{Setup.}
Let $X_0\sim p_0$ (source), $X_1\sim q$ (target) be independent random variables in $\mathbb{R}^N$ and consider the linear probability path
\begin{equation}
\label{eq:linear_path_cont}
X_t = (1-t)X_0 + t X_1, \qquad t\in[0,1].
\end{equation}
For flow matching with the linear path~\Eqref{eq:linear_path_cont}, the optimal velocity field equals the conditional drift:
\begin{equation}
\label{eq:fm_velocity_def}
v_t(x) = \mathbb{E}\left[X_1 - X_0 \middle| X_t = x \right].
\end{equation}
We derive a closed form of~\Eqref{eq:fm_velocity_def} that is directly computable from $p_0$ and $q$.

\paragraph{Derivation.}
By Bayes’ rule with a Dirac constraint for the linear relation~\Eqref{eq:linear_path_cont},
\begin{equation}
\label{eq:posterior_joint}
p(x_0,x_1 \mid X_t=x) \quad \propto  \quad p_0(x_0)\ q(x_1)\ \delta\left(x - (1-t)x_0 - t x_1\right).
\end{equation}
Integrating out $x_0$ using $\delta(Ay-b)=|{\det A}|^{-1}\ \delta\left(y-A^{-1}b\right)$ with $A=(1-t)I$ gives
\begin{align}
\label{eq:posterior_x1}
p(x_1 \mid X_t=x)\quad \propto \quad q(x_1)\ (1-t)^{-D}\ p_0\left( \frac{x - t x_1}{1-t} \right).
\end{align}
Hence,
\begin{align}
\label{eq:vt_ratio_integrals}
v_t(x)
&= \frac{\displaystyle \iint (x_1-x_0)\  p(x_0,x_1 \mid X_t=x)\ \mathrm{d}x_0 \mathrm{d}x_1}{\displaystyle \iint p(x_0,x_1 \mid X_t=x)\ \mathrm{d}x_0\ \mathrm{d}x_1} \\
&= \frac{\displaystyle \iint (x_1-x_0)\  p_0(x_0)\  q(x_1)\  \delta\left(x-(1-t)x_0-tx_1\right) \mathrm{d}x_0 \mathrm{d}x_1}{\displaystyle \iint p_0(x_0)\ q(x_1)\ \delta\left(x-(1-t)x_0-tx_1\right)\ \mathrm{d}x_0\ \mathrm{d}x_1} \\
&= \frac{\displaystyle \int q(x_1)\ p_0\left(\tfrac{x-tx_1}{1-t}\right)\ \Big(x_1 - \tfrac{x-tx_1}{1-t}\Big)\ \mathrm{d}x_1}{\displaystyle \int q(x_1)\ p_0\left(\tfrac{x-tx_1}{1-t}\right), \mathrm{d}x_1}.
\end{align}
Observing $x_1 - \tfrac{x-tx_1}{1-t} = \tfrac{x_1 - x}{1-t}$, we obtain the compact form
\begin{equation}
\label{eq:vt_closed_form_general}
v_t(x) = \frac{1}{1-t}\ \frac{\displaystyle \int q(x_1)\ p_0\left(\tfrac{x-tx_1}{1-t}\right)\ (x_1 - x)\ \mathrm{d}x_1}{\displaystyle \int q(x_1)\ p_0\left(\tfrac{x-tx_1}{1-t}\right)\ \mathrm{d}x_1}.
\end{equation}

When we have a dataset with samples $\{d_m\}_{m=1}^M$, the target distribution $q$ is approximated by the empirical measure $q(x_1) \approx \tfrac{1}{M}\sum_{m=1}^M \delta_{d_m}(x_1)$, then \Eqref{eq:vt_closed_form_general} reduces as follow:

\begin{equation}
\label{eq:vt_empirical_cont}
v_t(x) = \frac{1}{1-t}\ \frac{\sum_{m=1}^M p_0\left(\tfrac{x-td_m}{1-t}\right)\ (d_m - x)}{\sum_{m=1}^M p_0\left(\tfrac{x-td_m}{1-t}\right)}.
\end{equation}

When $p_0$ is standard Gaussian, $p_0(y)=(2\pi)^{-D/2}\exp\left(-\tfrac{1}{2}\|y\|_2^2\right)$, the normalizing constants cancel in~\Eqref{eq:vt_empirical_cont}, yielding the closed form velocity:
\begin{equation}
\label{eq:vt_continuous}
v_t(x) = \frac{1}{1-t}\ \frac{\sum_{m=1}^M \exp\left(-\tfrac{1}{2}\left\|\tfrac{x-td_m}{1-t}\right\|_2^2\right)\ (d_m - x)}{\sum_{m=1}^M \exp\left(-\tfrac{1}{2}\left\|\tfrac{x-td_m}{1-t}\right\|_2^2\right)}.
\end{equation}

This formulation has already been introduced in previous works~\citep{Karras:2022EDM,Bertrand:2025Closed}; however, to the best of our knowledge, no prior work has extended this idea to designing couplings for accelerating flow models using the re-flow technique~\citep{Liu:2023RF}. In the continuous domain, the backward velocity can be obtained directly by flipping the sign of the forward velocity. In contrast, in the discrete domain, the corresponding expression does not converge as $\lim_{t \to 1}$, and thus the backward velocity cannot be employed for sampling starting from data points. Therefore, in this section, we perform experiments using the forward velocity. 


\subsection{Continuous flow matching on MNIST}
\label{subsec:continuous_mnist}

We train rectified flow models on MNIST~\citep{lecun2002gradient} using two pairing strategies: (i) independent pairing (baseline) and (ii) closed-form pairing as described in \Secref{subsec:closed_velocity_flow}. We adopt CondOT~\citep{Lipman:2023FlowMatching} as our base flow model, which is originally trained with a independent pairing. We denote the variant of CondOT trained on pairs generated by the closed-form forward velocity as \methodname{}. To enable a few-step sampling, we subsequently apply rectification distillation (ReFlow~\citep{Liu:2023RF}) to each pretrained model, denoted by the suffix ``+RF''.

We use an NCSN++-style U-Net backbone~\citep{song2021scorebased} with a base width of 64 and 3 downsampling stages (doubling channels at each stage), optimized using Adam~\citep{Kingma:2014Adam} with a learning rate of $2\times 10^{-4}$. The pretraining takes 500 epochs. The distillation stage requires 200 epochs with a learning rate of $2\times 10^{-5}$.

\Tabref{tab:continuous_mnist} summarizes performance at various sampling steps. Without distillation, closed-form pairing (\ourname{}) yields significantly better FID in the few-step settings and maintains the performance in the many-step settings, relative to the baseline. With distillation (ReFlow~\citep{Liu:2023RF}, our method still shows better performance: \ourname{}+RF achieves a lower FID in every sampling budget than ReFlow applied to the baseline. These results show that closed-form pairing benefits both undistilled and distilled flow models, with especially large gains when the sampling steps are small.

\subsection{Continuous rectified flows on dimension-varying synthetic data}
\label{subsec:continuous_synthetic}

To assess scalability, we construct an N-fold product of the standard two-moons distribution, yielding a dataset in $\mathbb{R}^{2N}$. We consider dimensions $d\in\{2,4,8,16,32,64,128,256\}$ (i.e., $d=2N$) and train rectified flow models with and without closed-form pairing under a common training setup. The architecture is a simple transformer-based encoder with depth $8$, where the hidden size increases with dimension as $32, 64, 128, 192, 256, 384, 512, 768$, respectively.

For the synthetic experiments we report the Chamfer distance (log scale) between $50{,}000$ training datapoints and $5{,}000$ generated samples. Since the dataset is an $N$-fold product of 2D two-moons, Chamfer distance is computed using only the first two coordinates to keep the metric scale consistent across $d$ and measure fidelity to the base 2D geometry.

\begin{figure}[t!]
    \centering
    \begin{minipage}[t]{0.46\textwidth} 
        \centering
        \includegraphics[width=\textwidth]{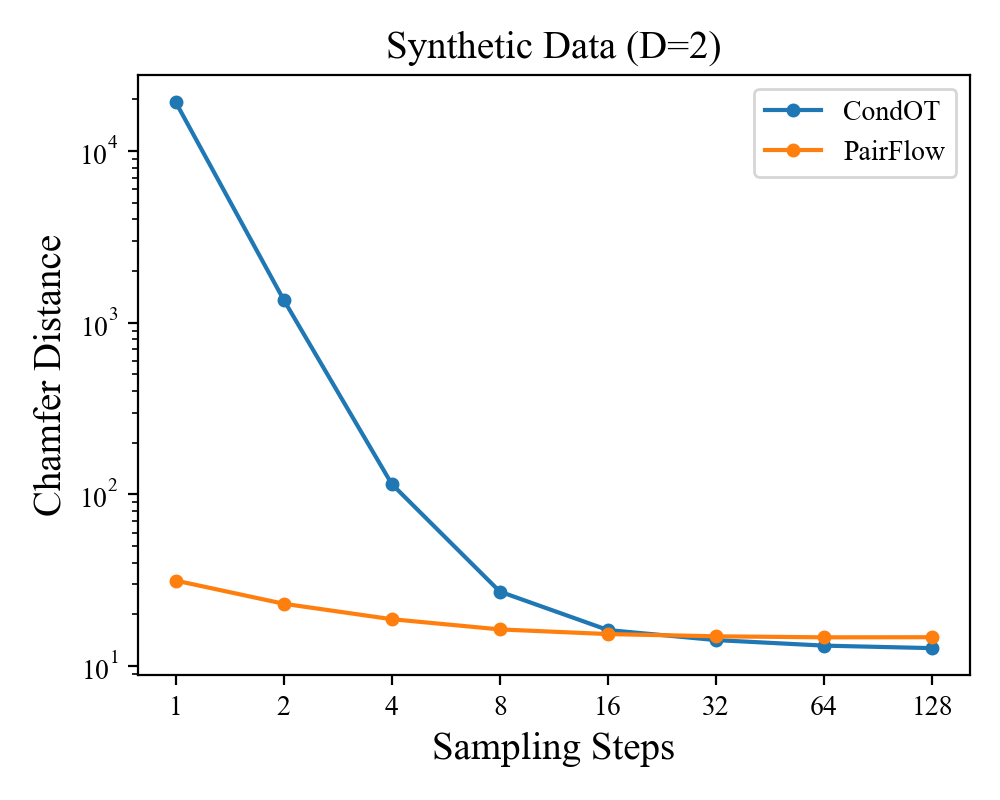}
    \end{minipage}
    \hfill 
    \begin{minipage}[t]{0.46\textwidth} 
        \centering
        \includegraphics[width=\textwidth]{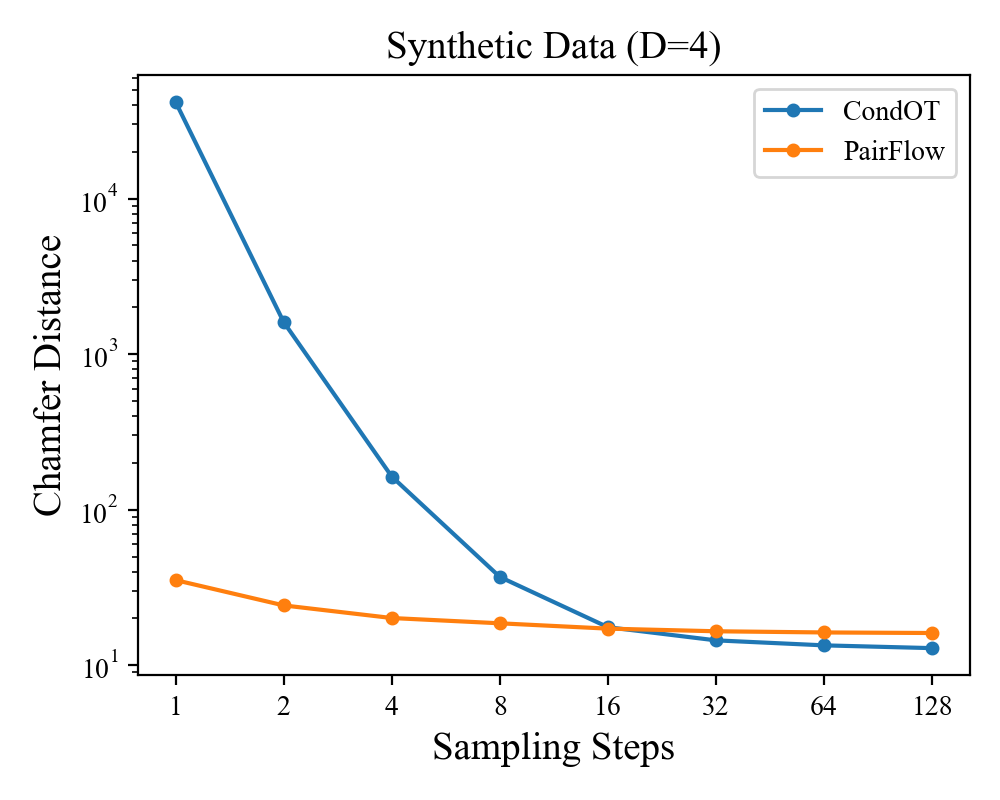}
    \end{minipage}
    \\
    \centering
    \begin{minipage}[t]{0.46\textwidth} 
        \centering
        \includegraphics[width=\textwidth]{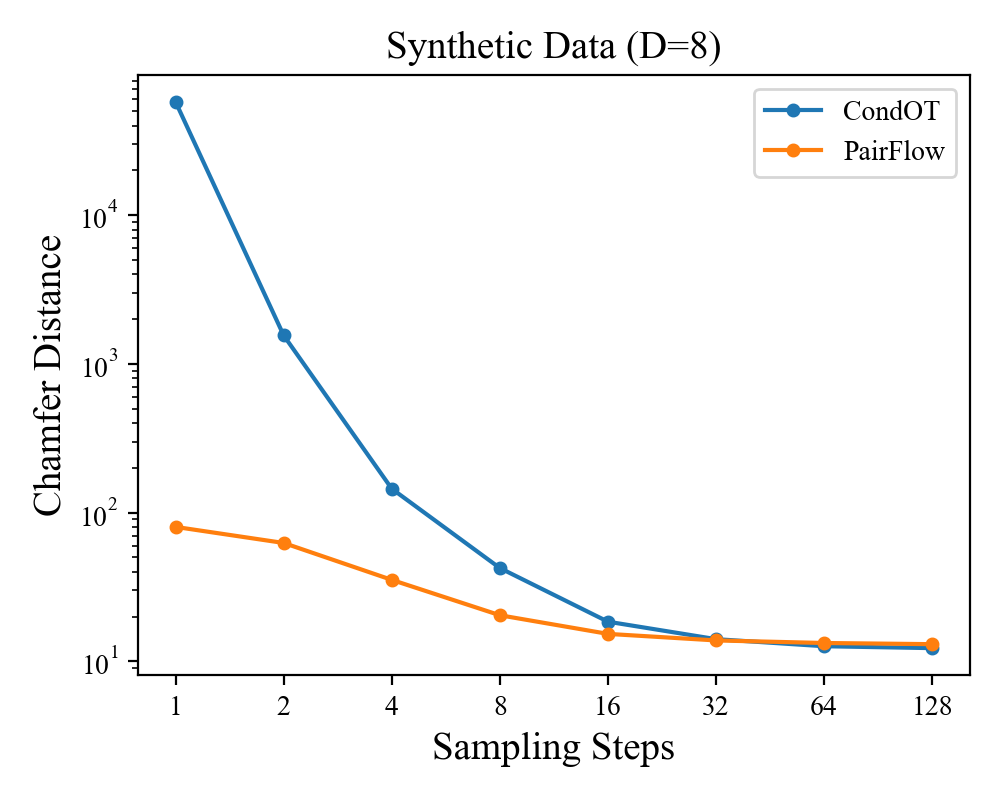}
    \end{minipage}
    \hfill 
    \begin{minipage}[t]{0.46\textwidth} 
        \centering
        \includegraphics[width=\textwidth]{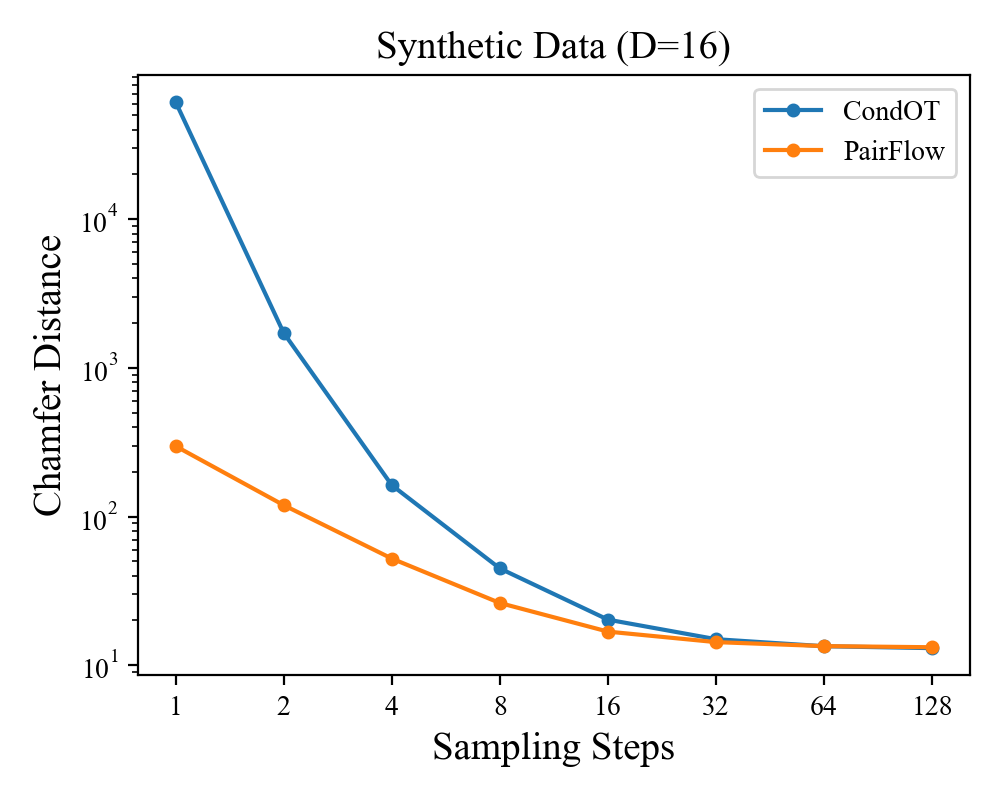}
    \end{minipage}
    \hfill 
    \\
    \centering
    \begin{minipage}[t]{0.46\textwidth} 
        \centering
        \includegraphics[width=\textwidth]{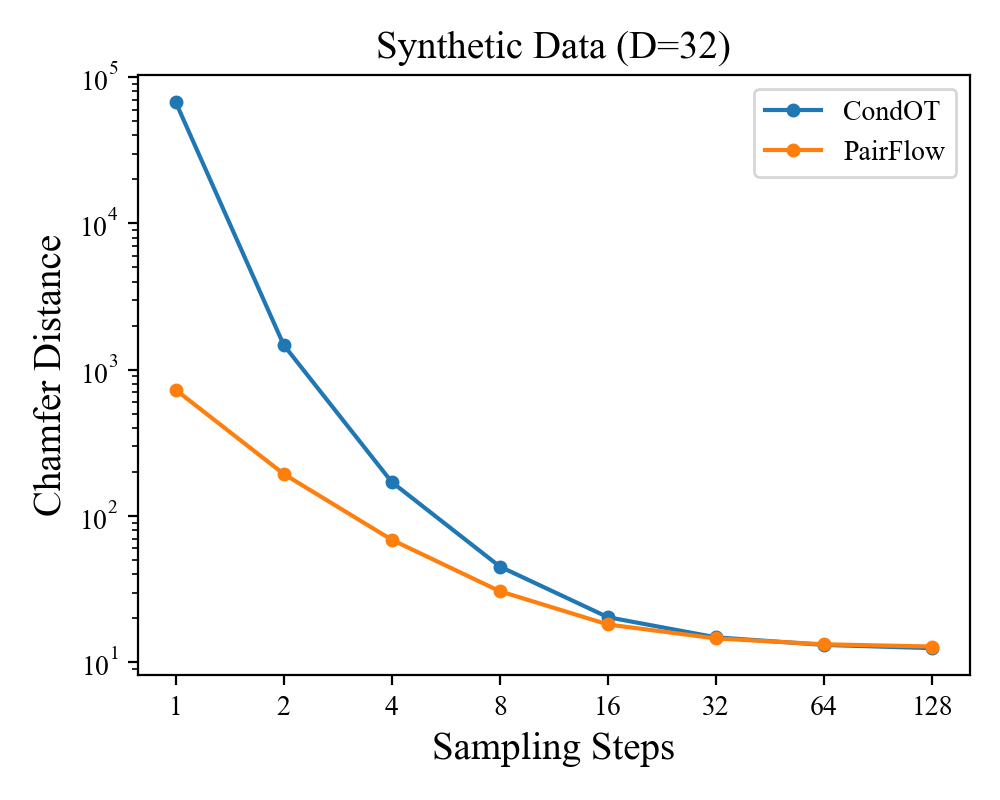}
    \end{minipage}
    \hfill 
    \begin{minipage}[t]{0.46\textwidth} 
        \centering
        \includegraphics[width=\textwidth]{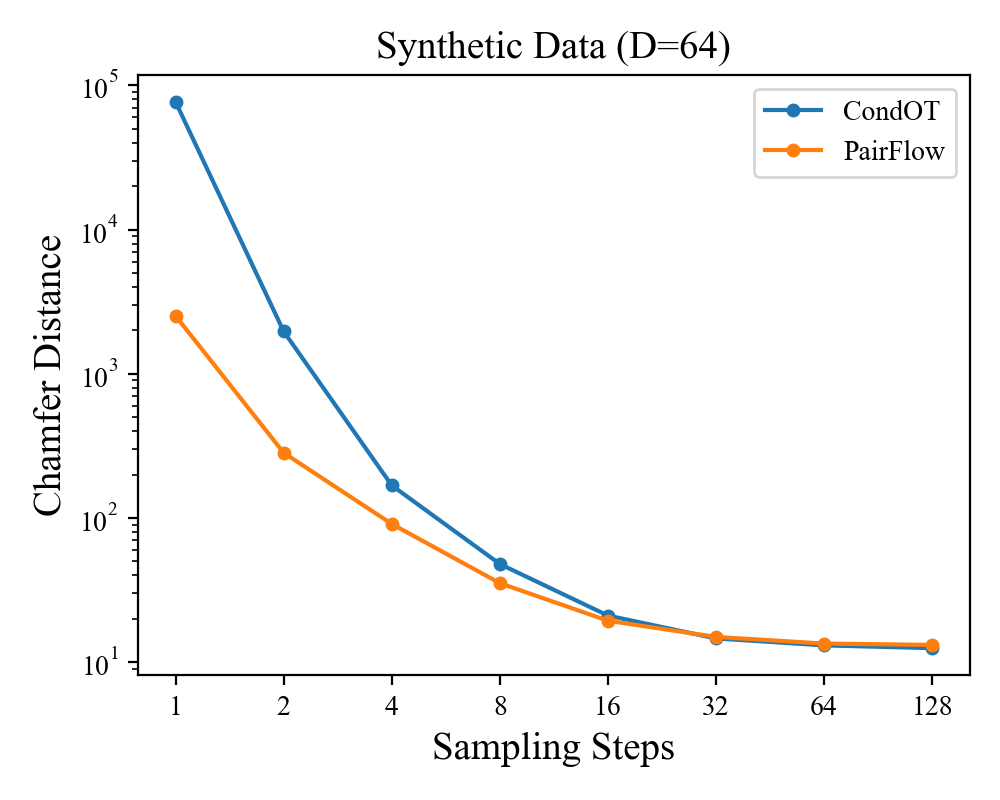}
    \end{minipage}
    \\
    \centering
    \begin{minipage}[t]{0.46\textwidth} 
        \centering
        \includegraphics[width=\textwidth]{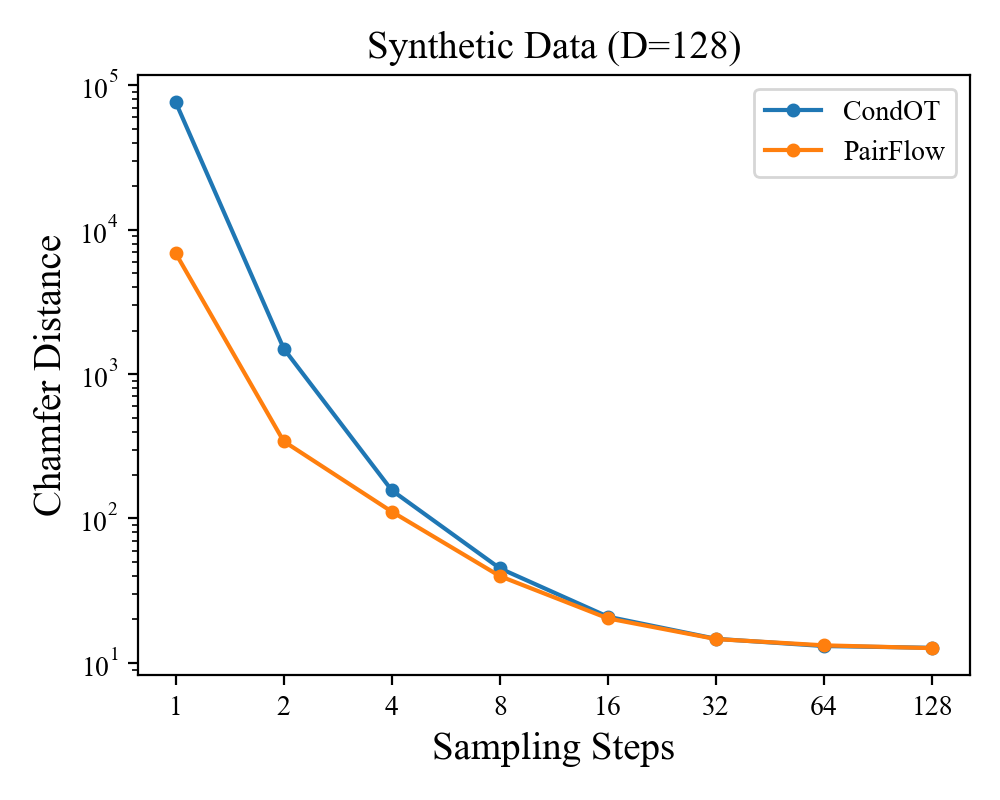}
    \end{minipage}
    \hfill 
    \begin{minipage}[t]{0.46\textwidth} 
        \centering
        \includegraphics[width=\textwidth]{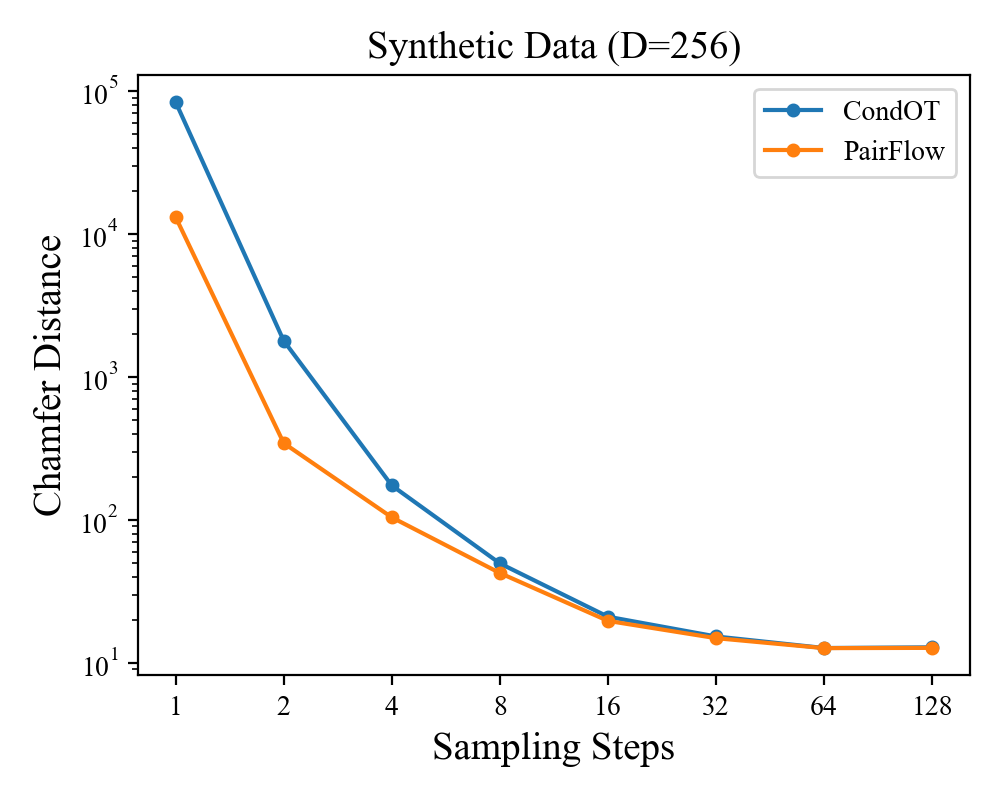}
    \end{minipage}
    \vspace{-1.0\baselineskip} 
    \caption{Step-wise performance analysis on synthetic N-fold two-moons. Each plot shows Chamfer distance (log scale) vs. sampling steps for different dimensions $d$. Closed-form pairing (\ourname{}) consistently outperforms standard CondOT—especially at few sampling steps—while the margin shrinks as d increases, indicating diminishing gains in high dimensions.}
    \label{fig:synthetic_graph}
\end{figure}

\Figref{fig:synthetic_graph} shows the quantitative results. At low dimensions, closed-form pairing yields substantial improvements over the independently paired baseline. However, as the data dimension increases, we observe that the magnitude of the improvement decreases. This trend suggests a practical limitation of closed-form pairing for high-dimensional continuous data.

\section{Additional Qualitative Results}

\begin{figure}[h]
\centering
\setlength{\tabcolsep}{0.4pt}
\renewcommand{\arraystretch}{0.3}
\centering
\begin{tabular}{!{\vrule width 1.2pt} @{\hskip 2.0pt} c @{\hskip 2.0pt} c @{\hskip 2.0pt} cccccccc}
\multirow{3}{*}{\raisebox{-1.0\height}{\rotatebox{90}{\textbf{64 steps}}}} &
\raisebox{0.4\height}{\rotatebox{90}{\scriptsize MDLM}} &
\includegraphics[width=0.11\linewidth]{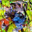} &
\includegraphics[width=0.11\linewidth]{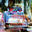} &
\includegraphics[width=0.11\linewidth]{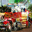} &
\includegraphics[width=0.11\linewidth]{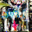} &
\includegraphics[width=0.11\linewidth]{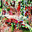} &
\includegraphics[width=0.11\linewidth]{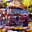} &
\includegraphics[width=0.11\linewidth]{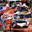} &
\includegraphics[width=0.11\linewidth]{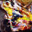} \\
& \raisebox{0.5\height}{\rotatebox{90}{\scriptsize UDLM}} &
\includegraphics[width=0.11\linewidth]{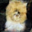} &
\includegraphics[width=0.11\linewidth]{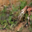} &
\includegraphics[width=0.11\linewidth]{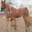} &
\includegraphics[width=0.11\linewidth]{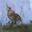} &
\includegraphics[width=0.11\linewidth]{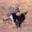} &
\includegraphics[width=0.11\linewidth]{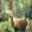} &
\includegraphics[width=0.11\linewidth]{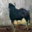} &
\includegraphics[width=0.11\linewidth]{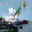} \\
& \raisebox{0.1\height}{\rotatebox{90}{\scriptsize \textbf{\ourname{}}}} &
\includegraphics[width=0.11\linewidth]{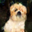} &
\includegraphics[width=0.11\linewidth]{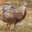} &
\includegraphics[width=0.11\linewidth]{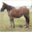} &
\includegraphics[width=0.11\linewidth]{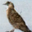} &
\includegraphics[width=0.11\linewidth]{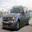} &
\includegraphics[width=0.11\linewidth]{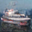} &
\includegraphics[width=0.11\linewidth]{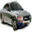} &
\includegraphics[width=0.11\linewidth]{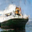} \\ [-0.1em]
\specialrule{1.2pt}{2.5pt}{2.3pt}
\multirow{3}{*}{\raisebox{-0.9\height}{\rotatebox{90}{\textbf{256 steps}}}} &
\raisebox{0.4\height}{\rotatebox{90}{\scriptsize MDLM}} &
\includegraphics[width=0.11\linewidth]{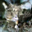} &
\includegraphics[width=0.11\linewidth]{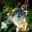} &
\includegraphics[width=0.11\linewidth]{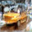} &
\includegraphics[width=0.11\linewidth]{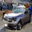} &
\includegraphics[width=0.11\linewidth]{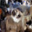} &
\includegraphics[width=0.11\linewidth]{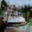} &
\includegraphics[width=0.11\linewidth]{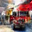} &
\includegraphics[width=0.11\linewidth]{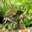} \\
& \raisebox{0.5\height}{\rotatebox{90}{\scriptsize UDLM}} &
\includegraphics[width=0.11\linewidth]{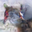} &
\includegraphics[width=0.11\linewidth]{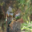} &
\includegraphics[width=0.11\linewidth]{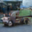} &
\includegraphics[width=0.11\linewidth]{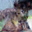} &
\includegraphics[width=0.11\linewidth]{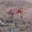} &
\includegraphics[width=0.11\linewidth]{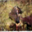} &
\includegraphics[width=0.11\linewidth]{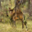} &
\includegraphics[width=0.11\linewidth]{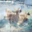} \\
& \raisebox{0.1\height}{\rotatebox{90}{\scriptsize \textbf{\ourname{}}}} &
\includegraphics[width=0.11\linewidth]{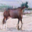} &
\includegraphics[width=0.11\linewidth]{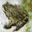} &
\includegraphics[width=0.11\linewidth]{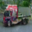} &
\includegraphics[width=0.11\linewidth]{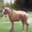} &
\includegraphics[width=0.11\linewidth]{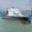} &
\includegraphics[width=0.11\linewidth]{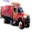} &
\includegraphics[width=0.11\linewidth]{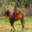} &
\includegraphics[width=0.11\linewidth]{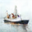} \\
 
\end{tabular}

\vspace{-0.7em}
\caption{Additional qualitative results of 64-step and 256-step generation on CIFAR-10 ($32 \times 32$).}
\label{fig:appendix_qualitative_cifar_64}
\end{figure}

\begin{figure}[h]
\centering
\setlength{\tabcolsep}{0.4pt}
\renewcommand{\arraystretch}{0.3}
\centering
\begin{tabular}{c @{\hskip 2.0pt} cccccccc}
\raisebox{0.35\height}{\rotatebox{90}{\scriptsize MDLM}} &
\includegraphics[width=0.11\linewidth]{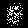} &
\includegraphics[width=0.11\linewidth]{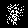} &
\includegraphics[width=0.11\linewidth]{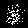} &
\includegraphics[width=0.11\linewidth]{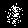} &
\includegraphics[width=0.11\linewidth]{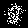} &
\includegraphics[width=0.11\linewidth]{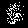} &
\includegraphics[width=0.11\linewidth]{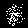} &
\includegraphics[width=0.11\linewidth]{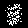} \\
\raisebox{0.35\height}{\rotatebox{90}{\scriptsize UDLM}} &
\includegraphics[width=0.11\linewidth]{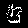} &
\includegraphics[width=0.11\linewidth]{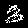} &
\includegraphics[width=0.11\linewidth]{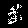} &
\includegraphics[width=0.11\linewidth]{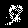} &
\includegraphics[width=0.11\linewidth]{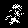} &
\includegraphics[width=0.11\linewidth]{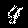} &
\includegraphics[width=0.11\linewidth]{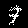} &
\includegraphics[width=0.11\linewidth]{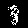} \\
\raisebox{0.01\height}{\rotatebox{90}{\scriptsize \textbf{\ourname{}}}} &
\includegraphics[width=0.11\linewidth]{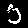} &
\includegraphics[width=0.11\linewidth]{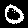} &
\includegraphics[width=0.11\linewidth]{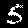} &
\includegraphics[width=0.11\linewidth]{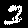} &
\includegraphics[width=0.11\linewidth]{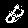} &
\includegraphics[width=0.11\linewidth]{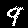} &
\includegraphics[width=0.11\linewidth]{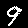} &
\includegraphics[width=0.11\linewidth]{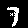} \\
\raisebox{0.35\height}{\rotatebox{90}{\scriptsize \makecell{UDLM \\ + DCD}}} &
\includegraphics[width=0.11\linewidth]{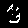} &
\includegraphics[width=0.11\linewidth]{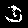} &
\includegraphics[width=0.11\linewidth]{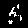} &
\includegraphics[width=0.11\linewidth]{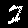} &
\includegraphics[width=0.11\linewidth]{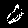} &
\includegraphics[width=0.11\linewidth]{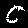} &
\includegraphics[width=0.11\linewidth]{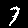} &
\includegraphics[width=0.11\linewidth]{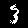} \\
\raisebox{0.01\height}{\rotatebox{90}{\scriptsize \makecell{\textbf{\ourname{}} \\ + DCD}}} &
\includegraphics[width=0.11\linewidth]{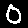} &
\includegraphics[width=0.11\linewidth]{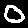} &
\includegraphics[width=0.11\linewidth]{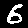} &
\includegraphics[width=0.11\linewidth]{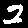} &
\includegraphics[width=0.11\linewidth]{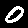} &
\includegraphics[width=0.11\linewidth]{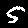} &
\includegraphics[width=0.11\linewidth]{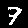} &
\includegraphics[width=0.11\linewidth]{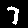} \\
\raisebox{0.35\height}{\rotatebox{90}{\scriptsize \makecell{UDLM \\ + ReDi}}} &
\includegraphics[width=0.11\linewidth]{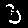} &
\includegraphics[width=0.11\linewidth]{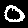} &
\includegraphics[width=0.11\linewidth]{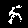} &
\includegraphics[width=0.11\linewidth]{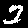} &
\includegraphics[width=0.11\linewidth]{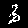} &
\includegraphics[width=0.11\linewidth]{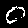} &
\includegraphics[width=0.11\linewidth]{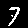} &
\includegraphics[width=0.11\linewidth]{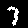} \\
\raisebox{0.01\height}{\rotatebox{90}{\scriptsize \makecell{\textbf{\ourname{}} \\ + ReDi}}} &
\includegraphics[width=0.11\linewidth]{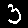} &
\includegraphics[width=0.11\linewidth]{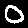} &
\includegraphics[width=0.11\linewidth]{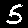} &
\includegraphics[width=0.11\linewidth]{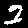} &
\includegraphics[width=0.11\linewidth]{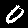} &
\includegraphics[width=0.11\linewidth]{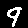} &
\includegraphics[width=0.11\linewidth]{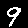} &
\includegraphics[width=0.11\linewidth]{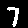} \\
\end{tabular}

\caption{Additional qualitative results of 1-step generation on MNIST-Binary ($28 \times 28$).}
\label{fig:appendix_qualitative_mnist_binary_1}
\end{figure}

\begin{figure}[h]
\centering
\setlength{\tabcolsep}{0.4pt}
\renewcommand{\arraystretch}{0.3}
\centering
\begin{tabular}{c @{\hskip 2.0pt} cccccccc}
\raisebox{0.35\height}{\rotatebox{90}{\scriptsize MDLM}} &
\includegraphics[width=0.11\linewidth]{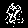} &
\includegraphics[width=0.11\linewidth]{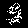} &
\includegraphics[width=0.11\linewidth]{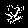} &
\includegraphics[width=0.11\linewidth]{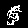} &
\includegraphics[width=0.11\linewidth]{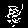} &
\includegraphics[width=0.11\linewidth]{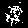} &
\includegraphics[width=0.11\linewidth]{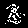} &
\includegraphics[width=0.11\linewidth]{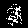} \\
\raisebox{0.35\height}{\rotatebox{90}{\scriptsize UDLM}} &
\includegraphics[width=0.11\linewidth]{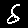} &
\includegraphics[width=0.11\linewidth]{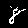} &
\includegraphics[width=0.11\linewidth]{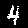} &
\includegraphics[width=0.11\linewidth]{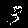} &
\includegraphics[width=0.11\linewidth]{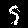} &
\includegraphics[width=0.11\linewidth]{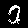} &
\includegraphics[width=0.11\linewidth]{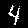} &
\includegraphics[width=0.11\linewidth]{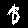} \\
\raisebox{0.01\height}{\rotatebox{90}{\scriptsize \textbf{\ourname{}}}} &
\includegraphics[width=0.11\linewidth]{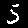} &
\includegraphics[width=0.11\linewidth]{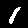} &
\includegraphics[width=0.11\linewidth]{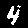} &
\includegraphics[width=0.11\linewidth]{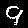} &
\includegraphics[width=0.11\linewidth]{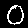} &
\includegraphics[width=0.11\linewidth]{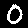} &
\includegraphics[width=0.11\linewidth]{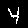} &
\includegraphics[width=0.11\linewidth]{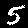} \\
\raisebox{0.35\height}{\rotatebox{90}{\scriptsize \makecell{UDLM \\ + DCD}}} &
\includegraphics[width=0.11\linewidth]{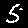} &
\includegraphics[width=0.11\linewidth]{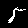} &
\includegraphics[width=0.11\linewidth]{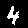} &
\includegraphics[width=0.11\linewidth]{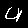} &
\includegraphics[width=0.11\linewidth]{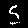} &
\includegraphics[width=0.11\linewidth]{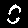} &
\includegraphics[width=0.11\linewidth]{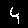} &
\includegraphics[width=0.11\linewidth]{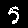} \\
\raisebox{0.01\height}{\rotatebox{90}{\scriptsize \makecell{\textbf{\ourname{}} \\ + DCD}}} &
\includegraphics[width=0.11\linewidth]{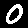} &
\includegraphics[width=0.11\linewidth]{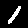} &
\includegraphics[width=0.11\linewidth]{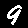} &
\includegraphics[width=0.11\linewidth]{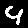} &
\includegraphics[width=0.11\linewidth]{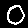} &
\includegraphics[width=0.11\linewidth]{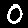} &
\includegraphics[width=0.11\linewidth]{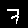} &
\includegraphics[width=0.11\linewidth]{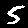} \\
\raisebox{0.35\height}{\rotatebox{90}{\scriptsize \makecell{UDLM \\ + ReDi}}} &
\includegraphics[width=0.11\linewidth]{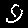} &
\includegraphics[width=0.11\linewidth]{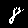} &
\includegraphics[width=0.11\linewidth]{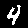} &
\includegraphics[width=0.11\linewidth]{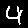} &
\includegraphics[width=0.11\linewidth]{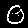} &
\includegraphics[width=0.11\linewidth]{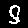} &
\includegraphics[width=0.11\linewidth]{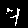} &
\includegraphics[width=0.11\linewidth]{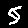} \\
\raisebox{0.01\height}{\rotatebox{90}{\scriptsize \makecell{\textbf{\ourname{}} \\ + ReDi}}} &
\includegraphics[width=0.11\linewidth]{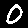} &
\includegraphics[width=0.11\linewidth]{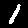} &
\includegraphics[width=0.11\linewidth]{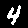} &
\includegraphics[width=0.11\linewidth]{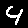} &
\includegraphics[width=0.11\linewidth]{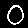} &
\includegraphics[width=0.11\linewidth]{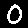} &
\includegraphics[width=0.11\linewidth]{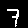} &
\includegraphics[width=0.11\linewidth]{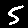} \\
\end{tabular}

\caption{Additional qualitative results of 2-step generation on MNIST-Binary ($28 \times 28$).}
\label{fig:appendix_qualitative_mnist_binary_2}
\end{figure}

In addition to~\Figref{fig:image_discrete_plots} in the main paper, we further visualize the 1-step and 2-step generation results for MNIST-Binary~\citep{lecun2002gradient} in~\Figref{fig:appendix_qualitative_mnist_binary_1} and~\Figref{fig:appendix_qualitative_mnist_binary_2}. As discussed in~\Secref{subsec:result_image},~\methodname{} outperforms the base models~\citep{Schiff:2025UDLM,Sahoo:2024MDLM} and achieves comparable quality to the base models combined with acceleration methods~\citep{Sahoo:2025Duo,Yoo:2025ReDi}. Additional visualizations for CIFAR-10~\citep{krizhevsky2009learning} with 64- and 256-step generations are shown in~\Figref{fig:appendix_qualitative_cifar_64}, demonstrating that our method outperforms the other base models.


\section{Re-Flow Iteration results}
\label{sec:reflow_iteration_results}

This section details the results of applying the iterative re-flow procedure~\citep{Liu:2023RF, Yoo:2025ReDi} on the QM9 dataset~\citep{ramakrishnan2014quantum}. We generated 1,024 samples across various timesteps and re-flow iterations, assessing validity, uniqueness, and novelty. The configuration for subsequent re-flow iterations follows the same protocol as our main experiment. As shown in~\Tabref{tab:qm9_rect_validity},~\Tabref{tab:qm9_rect_uniqueness}, and~\Tabref{tab:qm9_rect_novelty}, all metrics are averaged over 10 independent runs with standard deviations provided. We observed that iterative re-flow consistently enhances few-step generation capabilities. Notably, \methodname{} demonstrates superior performance over the baseline; it outperforms UDLM not only at equivalent iteration levels but also surpasses UDLM with multiple re-flow iterations even when \methodname{} uses one or no additional iterations.

\begin{table}[H]
\small
\centering
\caption{validity scores ($\uparrow$) on QM9~\citep{ramakrishnan2014quantum} for UDLM and \methodname{} across varying rectification steps and NFEs (1 to 64). The best and second-best values per column are highlighted in \textbf{bold} and \underline{underlined}, respectively.}
\label{tab:qm9_rect_validity}
\setlength{\tabcolsep}{3.2pt}
\begin{tabularx}{\textwidth}{l|YYYYYYY}
\toprule
Method & 1 & 2 & 4 & 8 & 16 & 32 & 64 \\
\midrule
UDLM & 17.5{\scriptsize$\pm$3.2} & 125.5{\scriptsize$\pm$11.8} & 497.6{\scriptsize$\pm$8.3} & 826.6{\scriptsize$\pm$10.3} & 953.5{\scriptsize$\pm$6.1} & \textbf{991.9{\scriptsize$\pm$4.2}} & \underline{1000.1{\scriptsize$\pm$3.5}} \\
+ Re-Flow & 59.7{\scriptsize$\pm$8.8} & 232.4{\scriptsize$\pm$9.2} & 588.4{\scriptsize$\pm$15.8} & 849.6{\scriptsize$\pm$14.2} & 940.5{\scriptsize$\pm$8.5} & 967.5{\scriptsize$\pm$5.2} & 978.8{\scriptsize$\pm$5.2} \\
+ Re-Flow & 160.9{\scriptsize$\pm$11.8} & 368.0{\scriptsize$\pm$11.8} & 673.3{\scriptsize$\pm$11.2} & 878.5{\scriptsize$\pm$11.6} & 945.4{\scriptsize$\pm$8.1} & 967.9{\scriptsize$\pm$7.4} & 975.8{\scriptsize$\pm$8.0} \\
+ Re-Flow & 280.2{\scriptsize$\pm$6.7} & 470.7{\scriptsize$\pm$18.6} & 742.2{\scriptsize$\pm$18.7} & 897.9{\scriptsize$\pm$9.4} & 945.3{\scriptsize$\pm$5.2} & 965.0{\scriptsize$\pm$5.9} & 972.6{\scriptsize$\pm$7.1} \\
\midrule
\methodname{} & 223.4{\scriptsize$\pm$12.7} & 416.0{\scriptsize$\pm$12.4} & 734.9{\scriptsize$\pm$7.2} & 921.5{\scriptsize$\pm$11.0} & \textbf{977.1{\scriptsize$\pm$3.9}} & \underline{990.9{\scriptsize$\pm$5.9}} & \textbf{1000.2{\scriptsize$\pm$4.5}} \\
+ Re-Flow & 361.0{\scriptsize$\pm$115.2} & 512.6{\scriptsize$\pm$44.2} & 775.7{\scriptsize$\pm$10.0} & 929.1{\scriptsize$\pm$11.6} & \underline{976.2{\scriptsize$\pm$4.5}} & 985.6{\scriptsize$\pm$6.7} & 993.1{\scriptsize$\pm$7.1} \\
+ Re-Flow & \underline{443.4{\scriptsize$\pm$13.4}} & \underline{598.6{\scriptsize$\pm$18.4}} & \underline{823.0{\scriptsize$\pm$16.0}} & \underline{935.2{\scriptsize$\pm$7.7}} & 969.0{\scriptsize$\pm$5.7} & 984.1{\scriptsize$\pm$6.0} & 989.0{\scriptsize$\pm$3.8} \\
+ Re-Flow & \textbf{529.2{\scriptsize$\pm$10.6}} & \textbf{688.0{\scriptsize$\pm$11.3}} & \textbf{863.6{\scriptsize$\pm$9.5}} & \textbf{945.9{\scriptsize$\pm$4.5}} & 975.7{\scriptsize$\pm$6.1} & 982.6{\scriptsize$\pm$7.1} & 990.2{\scriptsize$\pm$7.2} \\
\bottomrule
\end{tabularx}
\end{table}

\begin{table}[H]
\small
\centering
\caption{Uniqueness scores ($\uparrow$) on QM9~\citep{ramakrishnan2014quantum} for UDLM and \methodname{} across varying rectification steps and NFEs (1 to 64). The best and second-best values per column are highlighted in \textbf{bold} and \underline{underlined}, respectively.}
\label{tab:qm9_rect_uniqueness}
\setlength{\tabcolsep}{3.2pt}
\begin{tabularx}{\textwidth}{l|YYYYYYY}
\toprule
Method & 1 & 2 & 4 & 8 & 16 & 32 & 64 \\
\midrule
UDLM & 17.5{\scriptsize$\pm$3.2} & 125.4{\scriptsize$\pm$11.7} & 495.0{\scriptsize$\pm$8.2} & 819.5{\scriptsize$\pm$11.4} & 943.0{\scriptsize$\pm$5.7} & \underline{979.1{\scriptsize$\pm$5.0}} & \underline{990.0{\scriptsize$\pm$4.7}} \\
+ Re-Flow & 59.7{\scriptsize$\pm$8.8} & 231.6{\scriptsize$\pm$9.5} & 581.0{\scriptsize$\pm$15.1} & 834.7{\scriptsize$\pm$11.4} & 917.5{\scriptsize$\pm$9.6} & 944.8{\scriptsize$\pm$6.2} & 956.1{\scriptsize$\pm$5.2} \\
+ Re-Flow & 159.4{\scriptsize$\pm$10.9} & 363.3{\scriptsize$\pm$12.2} & 657.2{\scriptsize$\pm$9.3} & 846.3{\scriptsize$\pm$11.8} & 910.8{\scriptsize$\pm$7.3} & 930.5{\scriptsize$\pm$9.3} & 938.8{\scriptsize$\pm$11.7} \\
+ Re-Flow & 275.6{\scriptsize$\pm$7.2} & 456.3{\scriptsize$\pm$17.7} & 712.1{\scriptsize$\pm$15.7} & 856.8{\scriptsize$\pm$10.3} & 899.0{\scriptsize$\pm$4.9} & 907.0{\scriptsize$\pm$10.9} & 918.9{\scriptsize$\pm$5.7} \\
\midrule
\methodname{} & 223.0{\scriptsize$\pm$12.3} & 414.7{\scriptsize$\pm$12.0} & 731.4{\scriptsize$\pm$6.9} & \textbf{917.4{\scriptsize$\pm$11.8}} & \textbf{971.6{\scriptsize$\pm$4.3}} & \textbf{986.2{\scriptsize$\pm$5.3}} & \textbf{994.8{\scriptsize$\pm$5.2}} \\
+ Re-Flow & 359.5{\scriptsize$\pm$113.1} & 507.7{\scriptsize$\pm$43.0} & 765.3{\scriptsize$\pm$8.8} & \underline{913.5{\scriptsize$\pm$10.2}} & \underline{959.7{\scriptsize$\pm$5.8}} & 968.8{\scriptsize$\pm$8.4} & 973.0{\scriptsize$\pm$9.6} \\
+ Re-Flow & \underline{437.9{\scriptsize$\pm$13.8}} & \underline{586.7{\scriptsize$\pm$17.9}} & \underline{801.0{\scriptsize$\pm$15.4}} & 906.5{\scriptsize$\pm$10.1} & 939.1{\scriptsize$\pm$6.2} & 955.9{\scriptsize$\pm$7.2} & 960.2{\scriptsize$\pm$6.5} \\
+ Re-Flow & \textbf{516.9{\scriptsize$\pm$10.8}} & \textbf{662.8{\scriptsize$\pm$13.4}} & \textbf{828.7{\scriptsize$\pm$8.4}} & 903.2{\scriptsize$\pm$4.7} & 932.5{\scriptsize$\pm$9.5} & 942.8{\scriptsize$\pm$4.8} & 949.6{\scriptsize$\pm$8.4} \\
\bottomrule
\end{tabularx}
\end{table}

\begin{table}[H]
\small
\centering
\caption{Novelty scores ($\uparrow$) on QM9~\citep{ramakrishnan2014quantum} for UDLM and \methodname{} across varying rectification steps and NFEs (1 to 64). The best and second-best values per column are highlighted in \textbf{bold} and \underline{underlined}, respectively.}
\label{tab:qm9_rect_novelty}
\setlength{\tabcolsep}{3.2pt}
\begin{tabularx}{\textwidth}{l|YYYYYYY}
\toprule
Method & 1 & 2 & 4 & 8 & 16 & 32 & 64 \\
\midrule
UDLM & 13.8{\scriptsize$\pm$2.9} & 52.0{\scriptsize$\pm$8.2} & 120.0{\scriptsize$\pm$3.8} & \textbf{152.4{\scriptsize$\pm$9.1}} & \textbf{144.2{\scriptsize$\pm$12.4}} & \textbf{147.2{\scriptsize$\pm$9.7}} & \textbf{145.1{\scriptsize$\pm$9.0}} \\
+ Re-Flow & 31.4{\scriptsize$\pm$8.0} & 73.3{\scriptsize$\pm$7.1} & 110.4{\scriptsize$\pm$12.2} & \underline{126.8{\scriptsize$\pm$8.9}} & 116.3{\scriptsize$\pm$8.4} & 120.2{\scriptsize$\pm$9.0} & 117.6{\scriptsize$\pm$7.1} \\
+ Re-Flow & 61.1{\scriptsize$\pm$6.5} & 103.0{\scriptsize$\pm$8.6} & \textbf{129.0{\scriptsize$\pm$9.7}} & 124.6{\scriptsize$\pm$12.5} & \underline{128.3{\scriptsize$\pm$5.6}} & \underline{128.4{\scriptsize$\pm$10.6}} & \underline{122.8{\scriptsize$\pm$9.4}} \\
+ Re-Flow & 91.2{\scriptsize$\pm$9.2} & \textbf{116.4{\scriptsize$\pm$8.1}} & \underline{127.2{\scriptsize$\pm$8.3}} & 124.2{\scriptsize$\pm$9.3} & 121.9{\scriptsize$\pm$6.0} & 117.0{\scriptsize$\pm$4.9} & 121.8{\scriptsize$\pm$8.0} \\
\midrule
\methodname{} & 68.8{\scriptsize$\pm$7.8} & 85.6{\scriptsize$\pm$10.0} & 109.2{\scriptsize$\pm$7.8} & 96.8{\scriptsize$\pm$9.9} & 106.5{\scriptsize$\pm$12.5} & 108.9{\scriptsize$\pm$9.4} & 110.0{\scriptsize$\pm$9.9} \\
+ Re-Flow & 84.2{\scriptsize$\pm$11.3} & 92.0{\scriptsize$\pm$6.9} & 101.1{\scriptsize$\pm$9.6} & 98.6{\scriptsize$\pm$8.4} & 98.8{\scriptsize$\pm$13.9} & 95.8{\scriptsize$\pm$8.3} & 98.9{\scriptsize$\pm$7.5} \\
+ Re-Flow & \underline{100.8{\scriptsize$\pm$7.7}} & \underline{109.5{\scriptsize$\pm$8.7}} & 101.0{\scriptsize$\pm$6.8} & 101.1{\scriptsize$\pm$7.8} & 94.5{\scriptsize$\pm$6.6} & 95.6{\scriptsize$\pm$9.6} & 99.0{\scriptsize$\pm$8.8} \\
+ Re-Flow & \textbf{114.6{\scriptsize$\pm$9.4}} & 108.0{\scriptsize$\pm$5.9} & 106.5{\scriptsize$\pm$8.9} & 96.2{\scriptsize$\pm$9.8} & 95.6{\scriptsize$\pm$9.0} & 94.8{\scriptsize$\pm$7.1} & 95.7{\scriptsize$\pm$8.2} \\
\bottomrule
\end{tabularx}
\end{table}

\section{subset pairing results}

\label{sec:subset_pairing_results}

In this section, we present comprehensive experimental results from applying our subset-partition pairing technique to the ZINC-250k molecular dataset~\citep{irwin2012zinc}. Following the same protocol as our main experiments, we generated 1,024 samples across varying timesteps and partition counts to evaluate validity, uniqueness, and novelty. The only deviation from the standard method is the pairing strategy; here, we calculate the closed-form backward velocity exclusively within each subset to reduce the computational cost of~\Eqref{eqn:closed_form_noise_predictor}.
All reported metrics, summarized in~\Tabref{tab:zinc250k_subset_validity},~\Tabref{tab:zinc250k_subset_uniqueness}, and~\Tabref{tab:zinc250k_subset_novelty}, are averaged over 10 independent runs with corresponding standard deviations. Additionally, we report the pairing-time cost for each subset-partition configuration. The results demonstrate that our subset-pairing algorithm effectively reduces the computational time for pairing, while maintaining performance comparable to the full-set baseline.

\begin{table}[H]
\small
\centering
\caption{Validity scores ($\uparrow$) on the Zinc-250k dataset~\citep{irwin2012zinc} evaluated across different subset partitions and NFEs (1 to 64). The best and second-best values per column are highlighted in \textbf{bold} and \underline{underlined}, respectively. $T_{\methodname{}}$ denotes the runtime (in minutes) of each configuration, measured in wall-clock time using an RTX A6000 GPU.}
\label{tab:zinc250k_subset_validity}
\setlength{\tabcolsep}{3.2pt}
\begin{tabularx}{\textwidth}{l|c|YYYYYYY}
\toprule
Method & $T_{\text{\methodname{}}}$ & 1 & 2 & 4 & 8 & 16 & 32 & 64 \\
\midrule
UDLM & 0 & 0.3{\scriptsize$\pm$0.5} & 65.2{\scriptsize$\pm$8.2} & 435.7{\scriptsize$\pm$14.4} & 775.1{\scriptsize$\pm$19.5} & \textbf{887.3{\scriptsize$\pm$12.7}} & \textbf{921.5{\scriptsize$\pm$8.5}} & \textbf{937.3{\scriptsize$\pm$3.9}} \\
Random & 0 & 0.6{\scriptsize$\pm$0.9} & 68.3{\scriptsize$\pm$10.7} & 351.2{\scriptsize$\pm$15.8} & 569.4{\scriptsize$\pm$16.6} & 611.0{\scriptsize$\pm$16.3} & 602.4{\scriptsize$\pm$13.3} & 571.0{\scriptsize$\pm$13.2} \\
\midrule 
\methodname{} (Full) & 13m & 9.9{\scriptsize$\pm$2.3} & \textbf{146.3{\scriptsize$\pm$10.4}} & \textbf{533.9{\scriptsize$\pm$13.9}} & \underline{799.4{\scriptsize$\pm$9.2}} & 873.2{\scriptsize$\pm$14.1} & 901.0{\scriptsize$\pm$14.2} & 907.8{\scriptsize$\pm$7.7} \\
\methodname{} (2-Sub) & 6m & 10.6{\scriptsize$\pm$2.8} & \underline{145.7{\scriptsize$\pm$13.5}} & \underline{530.6{\scriptsize$\pm$20.3}} & \textbf{802.5{\scriptsize$\pm$6.9}} & \underline{882.6{\scriptsize$\pm$7.1}} & \underline{902.4{\scriptsize$\pm$13.4}} & \underline{911.5{\scriptsize$\pm$9.2}} \\
\methodname{} (4-Sub) & 2.9m & \underline{12.1{\scriptsize$\pm$3.3}} & 142.5{\scriptsize$\pm$5.2} & 509.7{\scriptsize$\pm$11.0} & 780.9{\scriptsize$\pm$14.4} & 858.9{\scriptsize$\pm$7.9} & 886.7{\scriptsize$\pm$8.7} & 899.0{\scriptsize$\pm$10.5} \\
\methodname{} (8-Sub) & 1.5m & \textbf{12.3{\scriptsize$\pm$2.5}} & 141.1{\scriptsize$\pm$6.7} & 510.9{\scriptsize$\pm$17.2} & 766.8{\scriptsize$\pm$12.6} & 857.5{\scriptsize$\pm$8.4} & 886.9{\scriptsize$\pm$9.6} & 896.5{\scriptsize$\pm$6.3} \\
\bottomrule
\end{tabularx}
\end{table}

\begin{table}[H]
\small
\centering
\caption{Uniqueness scores ($\uparrow$) on Zinc-250k~\citep{irwin2012zinc} evaluated across different subset partitions and NFEs (1 to 64). The best and second-best values per column are highlighted in \textbf{bold} and \underline{underlined}, respectively. $T_{\methodname{}}$ denotes the runtime (in minutes) of each configuration.}
\label{tab:zinc250k_subset_uniqueness}
\setlength{\tabcolsep}{3.2pt}
\begin{tabularx}{\textwidth}{l|c|YYYYYYY}
\toprule
Method & $T_{\methodname{}}$ & 1 & 2 & 4 & 8 & 16 & 32 & 64 \\
\midrule
UDLM & 0 & 0.3{\scriptsize$\pm$0.5} & 65.2{\scriptsize$\pm$8.2} & 435.7{\scriptsize$\pm$14.4} & 775.1{\scriptsize$\pm$19.5} & \textbf{887.3{\scriptsize$\pm$12.7}} & \textbf{921.5{\scriptsize$\pm$8.5}} & \textbf{937.2{\scriptsize$\pm$3.8}} \\
Random & 0 & 0.6{\scriptsize$\pm$0.9} & 68.3{\scriptsize$\pm$10.7} & 351.2{\scriptsize$\pm$15.8} & 569.4{\scriptsize$\pm$16.6} & 611.0{\scriptsize$\pm$16.3} & 602.4{\scriptsize$\pm$13.3} & 571.0{\scriptsize$\pm$13.2} \\
\midrule
\methodname{} (Full) & 13m & 9.9{\scriptsize$\pm$2.3} & \textbf{146.3{\scriptsize$\pm$10.4}} & \textbf{533.9{\scriptsize$\pm$13.9}} & \underline{799.4{\scriptsize$\pm$9.2}} & 873.2{\scriptsize$\pm$14.1} & 901.0{\scriptsize$\pm$14.2} & 907.8{\scriptsize$\pm$7.7} \\
\methodname{} (2-Sub) & 6m & 10.6{\scriptsize$\pm$2.8} & \underline{145.7{\scriptsize$\pm$13.5}} & \underline{530.6{\scriptsize$\pm$20.3}} & \textbf{802.5{\scriptsize$\pm$6.9}} & \underline{882.6{\scriptsize$\pm$7.1}} & \underline{902.4{\scriptsize$\pm$13.4}} & \underline{911.5{\scriptsize$\pm$9.2}} \\
\methodname{} (4-Sub) & 2.9m & \underline{12.1{\scriptsize$\pm$3.3}} & 142.5{\scriptsize$\pm$5.2} & 509.7{\scriptsize$\pm$11.0} & 780.9{\scriptsize$\pm$14.4} & 858.9{\scriptsize$\pm$7.9} & 886.7{\scriptsize$\pm$8.7} & 899.0{\scriptsize$\pm$10.5} \\
\methodname{} (8-Sub) & 1.5m & \textbf{12.3{\scriptsize$\pm$2.5}} & 141.1{\scriptsize$\pm$6.7} & 510.9{\scriptsize$\pm$17.2} & 766.8{\scriptsize$\pm$12.6} & 857.5{\scriptsize$\pm$8.4} & 886.9{\scriptsize$\pm$9.6} & 896.5{\scriptsize$\pm$6.3} \\
\bottomrule
\end{tabularx}
\end{table}
\begin{table}[H]
\small
\centering
\caption{Novelty scores ($\uparrow$) on Zinc-250k~\citep{irwin2012zinc} evaluated across different subset partitions and NFEs (1 to 64). The best and second-best values per column are highlighted in \textbf{bold} and \underline{underlined}, respectively. $T_{\methodname{}}$ denotes the runtime (in minutes) of each configuration.}
\label{tab:zinc250k_subset_novelty}
\setlength{\tabcolsep}{3.2pt}
\begin{tabularx}{\textwidth}{l|c|YYYYYYY}
\toprule
Method & $T_{\methodname{}}$ & 1 & 2 & 4 & 8 & 16 & 32 & 64 \\
\midrule
UDLM & 0 & 0.3{\scriptsize$\pm$0.5} & 65.2{\scriptsize$\pm$8.2} & 435.7{\scriptsize$\pm$14.4} & 775.1{\scriptsize$\pm$19.5} & \textbf{887.3{\scriptsize$\pm$12.7}} & \textbf{921.3{\scriptsize$\pm$8.8}} & \textbf{936.9{\scriptsize$\pm$4.1}} \\
Random & 0 & 0.6{\scriptsize$\pm$0.9} & 68.3{\scriptsize$\pm$10.7} & 351.2{\scriptsize$\pm$15.8} & 569.4{\scriptsize$\pm$16.6} & 611.0{\scriptsize$\pm$16.3} & 602.4{\scriptsize$\pm$13.3} & 571.0{\scriptsize$\pm$13.2} \\
\midrule
\methodname{} (Full) & 13m & 9.9{\scriptsize$\pm$2.3} & \textbf{146.3{\scriptsize$\pm$10.4}} & \textbf{533.9{\scriptsize$\pm$13.9}} & \underline{799.4{\scriptsize$\pm$9.2}} & 873.2{\scriptsize$\pm$14.1} & 901.0{\scriptsize$\pm$14.2} & 907.8{\scriptsize$\pm$7.7} \\
\methodname{} (2-Sub) & 6m & 10.6{\scriptsize$\pm$2.8} & \underline{145.7{\scriptsize$\pm$13.5}} & \underline{530.6{\scriptsize$\pm$20.3}} & \textbf{802.5{\scriptsize$\pm$6.9}} & \underline{882.6{\scriptsize$\pm$7.1}} & \underline{902.4{\scriptsize$\pm$13.4}} & \underline{911.5{\scriptsize$\pm$9.2}} \\
\methodname{} (4-Sub) & 2.9m & \underline{12.1{\scriptsize$\pm$3.3}} & 142.5{\scriptsize$\pm$5.2} & 509.7{\scriptsize$\pm$11.0} & 780.9{\scriptsize$\pm$14.4} & 858.8{\scriptsize$\pm$8.0} & 886.7{\scriptsize$\pm$8.7} & 899.0{\scriptsize$\pm$10.5} \\
\methodname{} (8-Sub) & 1.5m & \textbf{12.3{\scriptsize$\pm$2.5}} & 141.1{\scriptsize$\pm$6.7} & 510.9{\scriptsize$\pm$17.2} & 766.8{\scriptsize$\pm$12.6} & 857.5{\scriptsize$\pm$8.4} & 886.9{\scriptsize$\pm$9.6} & 896.5{\scriptsize$\pm$6.3} \\
\bottomrule
\end{tabularx}
\end{table}

\section{Application for more Complex Systems}
\label{sec:higher_dimension_results}

In this section, we evaluate our method on a higher-dimensional dataset. Specifically, we use the FFHQ~\citep{karras2019styleffhq} dataset, downsampled to $64\times64$. Following the same protocol as in our main experiments, we generate 5{,}000 samples across varying timesteps and report the FID computed against the training set. The results of this experiment are provided in~\Tabref{tab:ffhq_fid}. All training hyperparameters are kept identical to those used in the CIFAR-10 experiments described in Section~\ref{sec:results}.

\begin{table}[H]
\small
\centering
\caption{Comparison of FID scores ($\downarrow$) on FFHQ~\citep{karras2019styleffhq} downsampled to $64\times 64$ resolution across extended NFE steps (1 to 1024). Best values per column are highlighted in bold.}
\label{tab:ffhq_fid}
\resizebox{\textwidth}{!}{
\begin{tabular}{l|ccccccccccc}
\toprule
Method & 1 & 2 & 4 & 8 & 16 & 32 & 64 & 128 & 256 & 512 & 1024 \\
\midrule
UDLM & 403.04 & 399.26 & 363.97 & 273.31 & 153.71 & 97.87 & 74.85 & 63.93 & 59.28 & 55.99 & 55.30 \\
\methodname{} & \textbf{394.14} & \textbf{368.36} & \textbf{329.13} & \textbf{243.88} & \textbf{140.05} & \textbf{90.85} & \textbf{69.67} & \textbf{59.86} & \textbf{56.52} & \textbf{54.19} & \textbf{53.18} \\
\bottomrule
\end{tabular}
}
\end{table}

We adopt the LM1B~\citep{chelba2013lm1b} dataset to evaluate our method under a substantially larger vocabulary size and training corpus. The text corpus is segmented into sequences of varying lengths ($N = 16, 32, 64, 128$), while keeping the total number of training samples fixed ($|X_1| \approx 3.5\text{M}$). To assess generation quality, we compute generative perplexity using GPT-2 Large and entropy on 1,024 generated samples for each NFE setting. The results are summarized in~\Tabref{tab:lm1b_gen_ppl} and~\Tabref{tab:lm1b_entropy}. For training, we follow the network hyperparameter configuration of~\citep{Schiff:2025UDLM}, modifying only the number of training iterations for each sequence dimensionality.

\begin{table}[H]
\small
\centering
\caption{Generative Perplexity ($\downarrow$) on LM1B~\citep{chelba2013lm1b} measured with GPT2-large across varying lengths ($N$) and their corresponding training iterations (Iter.) over NFE steps 4 to 1024. Best values are highlighted in bold.}
\label{tab:lm1b_gen_ppl}
\resizebox{\textwidth}{!}{
\begin{tabular}{c|c|l|ccccccccc}
\toprule
$N$ & Iter. & Method & 4 & 8 & 16 & 32 & 64 & 128 & 256 & 512 & 1024 \\
\midrule
\multirow{2}{*}{16} & \multirow{2}{*}{200k}
& UDLM & 299.18 & 225.92 & 207.17 & 195.82 & 200.77 & \textbf{197.04} & 199.12 & \textbf{195.37} & 198.22 \\
& & \methodname{} & \textbf{242.22} & \textbf{208.04} & \textbf{200.99} & \textbf{190.36} & \textbf{191.74} & 199.45 & \textbf{188.84} & 196.91 & \textbf{198.12} \\
\midrule
\multirow{2}{*}{32} & \multirow{2}{*}{200k}
& UDLM & 263.93 & 192.78 & 167.85 & 167.49 & 155.68 & 150.52 & 152.40 & 151.74 & 154.02 \\
& & \methodname{} & \textbf{218.48} & \textbf{172.27} & \textbf{156.35} & \textbf{150.53} & \textbf{143.83} & \textbf{145.77} & \textbf{142.54} & \textbf{141.04} & \textbf{147.57} \\
\midrule
\multirow{2}{*}{64} & \multirow{2}{*}{400k}
& UDLM & 214.07 & 150.59 & 130.49 & 120.19 & 117.90 & 116.23 & 112.24 & 113.77 & 115.11 \\
& & \methodname{} & \textbf{174.78} & \textbf{138.94} & \textbf{123.06} & \textbf{115.71} & \textbf{114.73} & \textbf{112.92} & \textbf{111.29} & \textbf{107.06} & \textbf{110.83} \\
\midrule
\multirow{2}{*}{128} & \multirow{2}{*}{600k}
& UDLM & 169.61 & 123.48 & 105.13 & 98.94 & 97.89 & 94.92 & 93.75 & 94.12 & 93.59 \\
& & \methodname{} & \textbf{167.90} & \textbf{121.09} & \textbf{102.16} & \textbf{96.61} & \textbf{93.93} & \textbf{91.51} & \textbf{90.21} & \textbf{89.09} & \textbf{89.07} \\
\bottomrule
\end{tabular}
}
\end{table}

\begin{table}[H]
\small
\centering
\caption{Comparison of Entropy ($\uparrow$) on LM1B~\citep{chelba2013lm1b} across varying lengths ($N$) and training iterations (Iter.) over NFE steps 4 to 1024. Best values are highlighted in bold.}
\label{tab:lm1b_entropy}
\resizebox{\textwidth}{!}{
\begin{tabular}{c|c|l|ccccccccc}
\toprule
$N$ & Iter. & Method & 4 & 8 & 16 & 32 & 64 & 128 & 256 & 512 & 1024 \\
\midrule
\multirow{2}{*}{16} & \multirow{2}{*}{200k}
& UDLM & 2.46 & \textbf{2.49} & 2.50 & 2.50 & 2.50 & 2.51 & 2.50 & 2.51 & 2.50 \\
& & \methodname{} & \textbf{2.48} & \textbf{2.49} & \textbf{2.51} & \textbf{2.52} & \textbf{2.52} & \textbf{2.53} & \textbf{2.52} & \textbf{2.53} & \textbf{2.52} \\
\midrule
\multirow{2}{*}{32} & \multirow{2}{*}{200k}
& UDLM & 3.05 & 3.09 & 3.12 & 3.13 & 3.13 & 3.13 & 3.13 & 3.13 & 3.13 \\
& & \methodname{} & \textbf{3.06} & \textbf{3.12} & \textbf{3.13} & \textbf{3.14} & \textbf{3.15} & \textbf{3.15} & \textbf{3.15} & \textbf{3.15} & \textbf{3.16} \\
\midrule
\multirow{2}{*}{64} & \multirow{2}{*}{400k}
& UDLM & \textbf{3.57} & 3.63 & 3.67 & 3.68 & 3.69 & 3.70 & 3.70 & 3.69 & 3.70 \\
& & \methodname{} & \textbf{3.57} & \textbf{3.65} & \textbf{3.69} & \textbf{3.70} & \textbf{3.71} & \textbf{3.71} & \textbf{3.71} & \textbf{3.72} & \textbf{3.71} \\
\midrule
\multirow{2}{*}{128} & \multirow{2}{*}{600k}
& UDLM & 3.98 & 4.09 & 4.14 & 4.16 & 4.17 & 4.17 & 4.17 & 4.18 & 4.18 \\
& & \methodname{} & \textbf{4.00} & \textbf{4.11} & \textbf{4.16} & \textbf{4.18} & \textbf{4.19} & \textbf{4.20} & \textbf{4.20} & \textbf{4.19} & \textbf{4.20} \\
\bottomrule
\end{tabular}
}
\end{table}

\section{Analysis for the Overfitting in Image Domains}
\label{sec:overfit_analysis}

In this section, we evaluate our method using FID computed on the test sets of two image domains: CIFAR-10~\citep{krizhevsky2009learning} and MNIST-Binary~\citep{lecun2002gradient}. For CIFAR-10, we additionally report FID scores measured with DINOv2~\citep{oquab2024dinov2}. The overall results are summarized in~\Tabref{tab:cifar_fid_test} and~\Tabref{tab:mnist_binary_fid_test}. Across all evaluation metrics, the performance trend is consistent with our main findings—\methodname{} delivers improved generation quality over the baseline, with especially strong gains in the few-step generation regime.

\begin{table}[H]
\small
\centering
\caption{FID and FID-Dino scores ($\downarrow$) on test dataset for CIFAR-10~\citep{krizhevsky2009learning} comparison across extended NFE steps (1 to 1024). Best values are bolded.}
\label{tab:cifar_fid_test}
\resizebox{\textwidth}{!}{
\begin{tabular}{l|ccccccccccc}
\toprule
NFE & 1 & 2 & 4 & 8 & 16 & 32 & 64 & 128 & 256 & 512 & 1024 \\
\midrule
Method & \multicolumn{11}{c}{FID ($\downarrow$)} \\
\midrule
UDLM & 306.45 & 296.77 & 266.64 & 178.11 & 114.04 & 80.40 & 62.70 & 53.83 & 50.96 & 47.61 & 47.16 \\
\methodname{} & \textbf{235.65} & \textbf{247.18} & \textbf{209.94} & \textbf{137.16} & \textbf{94.24} & \textbf{67.53} & \textbf{51.48} & \textbf{43.79} & \textbf{42.44} & \textbf{40.85} & \textbf{39.92} \\
\midrule
Method & \multicolumn{11}{c}{FID-DINOv2 ($\downarrow$)} \\
\midrule
UDLM & 2448.46 & 2410.65 & 1975.58 & 1344.44 & 959.39 & 755.80 & 646.47 & 598.53 & 598.54 & 553.17 & 560.75 \\
\methodname{} & \textbf{2059.26} & \textbf{1988.21} & \textbf{1626.08} & \textbf{1127.30} & \textbf{828.09} & \textbf{623.89} & \textbf{530.70} & \textbf{486.59} & \textbf{484.27} & \textbf{470.33} & \textbf{470.05} \\
\bottomrule
\end{tabular}
}
\end{table}


\begin{table}[H]
\small
\centering
\caption{FID scores ($\downarrow$) on the MNIST-Binary~\citep{lecun2002gradient} test set across various NFE steps (1 to 64). Best values are bolded.}
\label{tab:mnist_binary_fid_test}
\resizebox{\textwidth}{!}{
\begin{tabularx}{\textwidth}{l|YYYYYYY}
\toprule
Method & 1 & 2 & 4 & 8 & 16 & 32 & 64 \\
\midrule
UDLM & 129.05 & 42.17 & 11.42 & \textbf{6.18} & \textbf{5.13} & \textbf{5.37} & 5.50 \\
\methodname{} & \textbf{42.87} & \textbf{17.37} & \textbf{9.62} & 6.36 & 5.80 & 5.51 & \textbf{5.24} \\
\midrule
UDLM+DCD & 57.85 & 19.23 & \textbf{9.82} & \textbf{8.41} & 7.87 & \textbf{7.12} & \textbf{7.38} \\
\methodname{}+DCD & \textbf{19.56} & \textbf{13.06} & 10.90 & 8.55 & \textbf{7.40} & 7.22 & 7.85 \\
\midrule
UDLM+ReDi & 19.08 & 10.79 & \textbf{8.77} & \textbf{7.01} & \textbf{6.89} & \textbf{6.57} & \textbf{6.61} \\
\methodname{}+ReDi & \textbf{13.73} & \textbf{9.59} & 8.98 & 7.24 & 7.22 & 6.98 & 7.12 \\
\bottomrule
\end{tabularx}
}
\end{table}

We further assess potential memorization by measuring nearest-neighbor distances with respect to the training set. For MNIST-Binary~\citep{lecun2002gradient}, we compute pixel-wise $\ell_2$ distances, whereas for CIFAR-10~\citep{krizhevsky2009learning}, we evaluate both $\ell_2$ distance and cosine similarity between features extracted using DINOv2~\citep{oquab2024dinov2}. As summarized in~\Tabref{tab:cifar_nn_distance} and~\Tabref{tab:mnist_binary_nn_distance}, across all evaluation settings, the nearest-neighbor distances of \methodname{} are comparable to or slightly larger than those of the baseline. These results support the conclusion that our method does not suffer from severe overfitting or excessive memorization of the training data.

\begin{table}[H]
\small
\centering
\caption{Comparison of $\ell_2$ and Dino~\citep{oquab2024dinov2} Cosine nearest neighbor distance on the CIFAR-10~\citep{krizhevsky2009learning} training set across extended NFE steps (1 to 1024). Best values are bolded.}
\label{tab:cifar_nn_distance}
\resizebox{\textwidth}{!}{
\begin{tabular}{l|ccccccccccc}
\toprule
NFE & 1 & 2 & 4 & 8 & 16 & 32 & 64 & 128 & 256 & 512 & 1024 \\
\midrule
Metric & \multicolumn{11}{c}{$\ell_2$ ($\uparrow$)} \\
\midrule
UDLM & 7.97 & \textbf{9.13} & \textbf{10.06} & \textbf{10.06} & \textbf{9.40} & \textbf{8.94} & \textbf{8.63} & 8.41 & 8.29 & 8.29 & 8.22 \\
\methodname{} & \textbf{8.03} & 8.76 & 9.42 & 9.56 & 9.18 & 8.75 & \textbf{8.63} & \textbf{8.55} & \textbf{8.51} & \textbf{8.52} & \textbf{8.52} \\
\midrule
Metric & \multicolumn{11}{c}{Cosine(DINOv2) ($\downarrow$)} \\
\midrule
UDLM & \textbf{0.242} & \textbf{0.227} & \textbf{0.231} & 0.241 & 0.237 & 0.238 & 0.235 & 0.237 & 0.237 & \textbf{0.235} & 0.237 \\
\methodname{} & 0.245 & 0.228 & 0.235 & \textbf{0.239} & \textbf{0.236} & \textbf{0.232} & \textbf{0.232} & \textbf{0.230} & \textbf{0.232} & \textbf{0.235} & \textbf{0.233} \\
\bottomrule
\end{tabular}
}
\end{table}


\begin{table}[H]
\small
\centering
\caption{Comparison of $\ell_2$ nearest neighbor distance on the MNIST-Binary~\citep{lecun2002gradient} training set across extended NFE steps (1 to 64). Best values are bolded.}
\label{tab:mnist_binary_nn_distance}
\resizebox{\textwidth}{!}{
\begin{tabularx}{\textwidth}{l|YYYYYYY}
\toprule
Method & 1 & 2 & 4 & 8 & 16 & 32 & 64 \\
\midrule
UDLM & \textbf{8.18} & \textbf{7.06} & 6.55 & \textbf{6.42} & \textbf{6.27} & 6.29 & 6.25 \\
\methodname{} & 7.36 & 6.95 & \textbf{6.63} & \textbf{6.42} & 6.26 & \textbf{6.34} & \textbf{6.30} \\
\midrule
UDLM+DCD & 7.24 & 6.78 & 6.57 & 6.50 & 6.31 & 6.40 & 6.39 \\
\methodname{}+DCD & \textbf{7.64} & \textbf{7.34} & \textbf{7.11} & \textbf{6.91} & \textbf{6.74} & \textbf{6.76} & \textbf{6.79} \\
\midrule
UDLM+ReDi & \textbf{7.14} & \textbf{6.81} & \textbf{6.49} & \textbf{6.24} & \textbf{6.10} & \textbf{6.14} & \textbf{6.08} \\
\methodname{}+ReDi & 6.84 & 6.57 & 6.35 & 6.09 & 5.95 & 5.97 & 5.96 \\
\bottomrule
\end{tabularx}
}
\end{table}

\section{Limitations and Future Work}
\label{sec:limitation_future_work}

We hope this work initiates broader discussion on reducing training compute while still enabling fast generation in generative models. Such efficiency can have a significant impact, from reducing energy consumption in training large-scale generative models to contributing to the democratization of foundation model development.

A natural follow-up question to our work is whether the same idea can be applied to continuous Flow Matching (FM). We have evaluated this extension on continuous FM models, with results provided in App.~\ref{sec:continuous_flow_matching}. Our experiments with synthetic data show that the method is effective for relatively low-dimensional data, while its advantage a bit diminishes for higher-dimensional data. We will further investigate the effect of our method on continuous data, where we hypothesize that a substantially larger number of source–target pairs will be required. Nonetheless, we emphasize that even in this initial exploration of accelerating flow models through well-aligned pairing,~\ourname{} is particularly well-suited for low-dimensional discrete data, which includes many forms of scientific data such as molecular and protein structures.


\end{document}